\documentclass{article}
\PassOptionsToPackage{numbers,square}{natbib}
\usepackage[preprint]{neurips_2024}

\usepackage{graphicx}
\usepackage{amsmath}
\usepackage{amssymb}
\usepackage[colorlinks=true,linkcolor=blue,urlcolor=blue,citecolor=blue]{hyperref}
\usepackage{booktabs}
\usepackage{xcolor}
\usepackage{listings}

\lstnewenvironment{promptblock}{%
\lstset{
  basicstyle=\ttfamily\small,
  breaklines=true,
  breakatwhitespace=true,
  columns=fullflexible,
  frame=single,
  backgroundcolor=\color{gray!10},
  keepspaces=true,
  showstringspaces=false,
}
}{}

\newcommand{\promptinline}[1]{%
  \colorbox{gray!10}{\ttfamily\small #1}%
}

\title{When Models Manipulate Manifolds: The Geometry of a Counting Task}
  \author{
    Wes Gurnee\rlap{,}$^{*}$ Emmanuel Ameisen\rlap{,}$^{*}$ Isaac Kauvar, Julius Tarng, \\
    \bf Adam Pearce, Chris Olah, Joshua Batson$^{*\dagger}$  \\[1em]
    Anthropic \\ \hspace{2em} \small{$^{*}$Equal contribution. $^{\dagger}$Corresponding author: \texttt{joshb@anthropic.com}}
  }
\begin{document}

\maketitle

\begin{abstract}
    Language models can perceive visual properties of text despite receiving only sequences of tokens—we mechanistically investigate how Claude 3.5 Haiku accomplishes one such task: linebreaking in fixed-width text. We find that character counts are represented on low-dimensional curved manifolds discretized by sparse feature families, analogous to biological place cells. Accurate predictions emerge from a sequence of geometric transformations: token lengths are accumulated into character count manifolds, attention heads ``twist" these manifolds to estimate distance to the line boundary, and the decision to break the line is enabled by arranging estimates orthogonally to create a linear decision boundary. We validate our findings through causal interventions and discover ``visual illusions"—character sequences that hijack the counting mechanism. Our work demonstrates the rich sensory processing of early layers, the intricacy of attention algorithms, and the importance of combining feature-based and geometric views of interpretability.\footnote{This is an archival version of a paper originally published on October 21, 2025 at \href{https://transformer-circuits.pub/2025/linebreaks/index.html}{Transformer Circuits}, which we recommend for its improved interactivity and styling.}

\end{abstract}

\section{Introduction}\label{sec:introduction}

Intelligent systems need perception to understand, predict, and navigate their environment. These sensory capabilities reflect what's useful for survival in a specific environment: bats use echolocation, migratory birds sense magnetic fields, Arctic reindeer shift their UV vision seasonally. But when your world is made of text, what do you see?  Language models encounter many text-based tasks that benefit from visual or spatial reasoning: parsing ASCII art, interpreting tables, or handling text wrapping constraints. Yet their only “sensory” input is a sequence of integers representing tokens. They must learn perceptual abilities from scratch, developing specialized mechanisms in the process.

In this work, we investigate the mechanisms that enable Claude 3.5 Haiku to perform a natural perceptual task which is common in pretraining corpora and involves tracking position in a document. We find learned representations of position that are in some ways quite similar to the biological neurons found in mammals who perform analogous tasks (“place cells” and “boundary cells” in mice), but in other ways unique to the constraints of the residual stream in language models. We study these representations and find dual interpretations: we can understand them as a family of discrete features or as a one-dimensional “feature manifold”/“multidimensional feature” \cite{olah2023manifold,olah2024multidimensional,gorton2024curve,engels2025not}.\footnote{All features have a magnitude dimension; so a discrete feature is a one-dimensional ray, and a one-dimensional feature manifold is the set of all scalings of that manifold, contracting to the origin. See \href{https://transformer-circuits.pub/2024/july-update/index.html\#linear-representations}{What is a Linear Representation? What is a Multidimensional Feature?}} In the first interpretation, position is determined by which features activate and how strongly; in the latter interpretation, it's determined by angular movement on the feature manifold. Similarly, computation has two dual interpretations, as discrete circuits or geometric transformations.

\begin{figure}[t]
    \centering
    \includegraphics[width=0.9\textwidth,height=0.7\textheight,keepaspectratio]{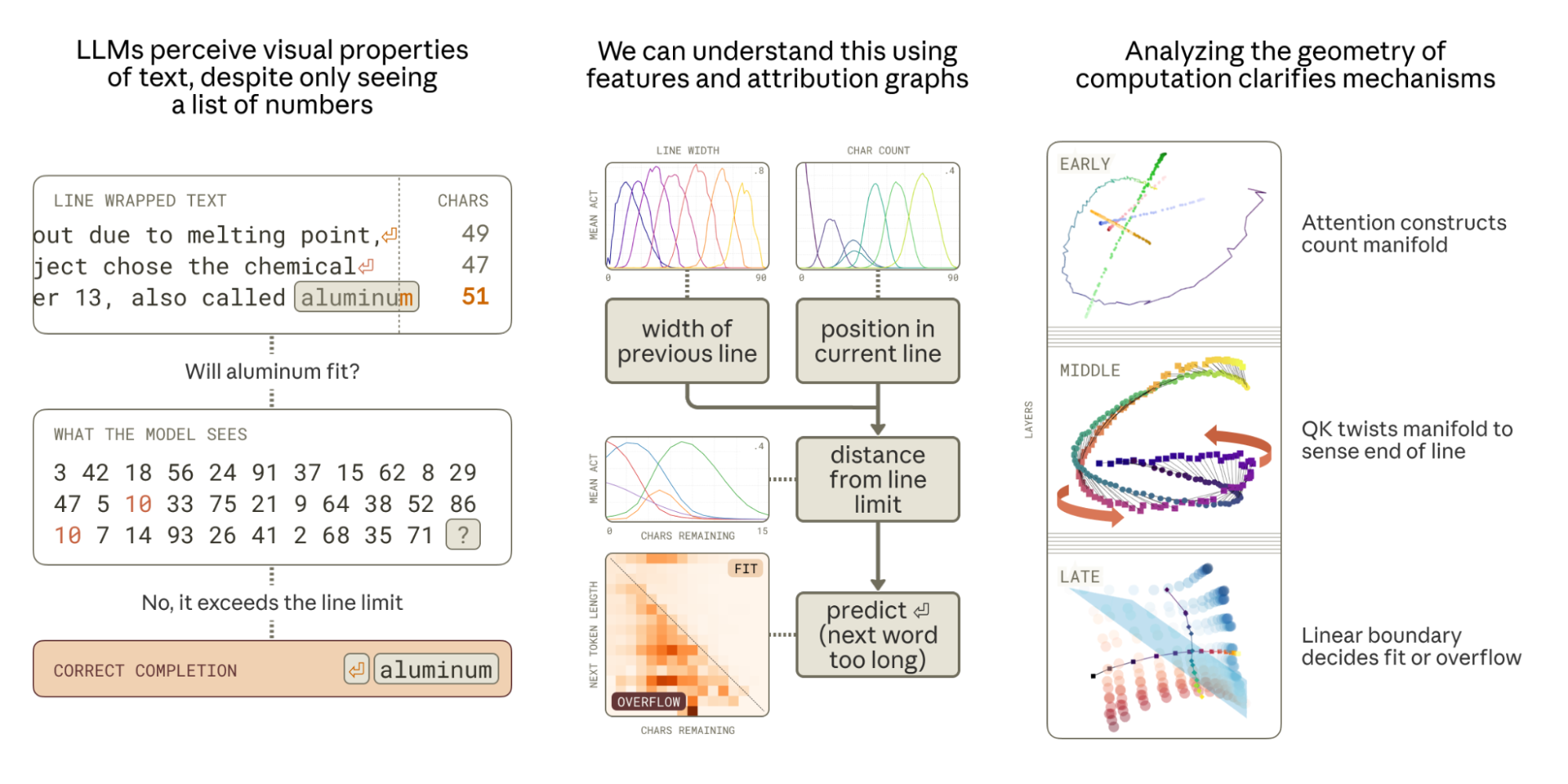}
    \label{fig:gdoc_1}
    \caption{Left: a description of the linebreaking task. Center: example feature families we study. Right: a highlight of the geometry of representations and computations.}
\end{figure}

The task we study is linebreaking in fixed-width text. When training on source code, chat logs, email archives, scanned articles, or judicial rulings that have line width constraints, how does the model learn to predict when to break a line?\footnote{Michaud\textit{ et al.} looked for “quanta” of model skills by clustering gradients \cite{michaud2023the}. Their Figure 1 shows that predicting newlines in fixed-width text formed one of the top 400 clusters for the smallest model in the 70m Pythia model.} Human visual perception lets us do this almost completely subconsciously -- when writing a birthday card, you can see when you are out of room on a line and need to begin the next -- but language models see just a list of integers. In order to correctly predict the next \textit{token}, in addition to selecting the next \textit{word}, the model must somehow count the characters in the current line, subtract that from the line width constraint of the document, and compare the number of characters remaining to the length of the next word. As a concrete example, consider the below pair of prompts with an implicit 50-character line wrapping constraint.\footnote{The wrapping constraint is implicit. Each newline gives a lower bound (the previous word did fit) and an upper bound (the next word did not). We do not nail down the extent to which the model performs optimal inference with respect to those constraints, rather focusing on how it approximately uses the length of each preceding line to determine whether to break the next. There are also many edge cases for handling tokenization and punctuation. A model could even attempt to infer whether the source document used a non-monospace font and then use the pixel count rather than the character count as a predictive signal!} When the next word fits, the model says it; when it does not, the model breaks the line:

\begin{figure}[!htp]
    \centering
    \includegraphics[width=\textwidth,height=0.7\textheight,keepaspectratio]{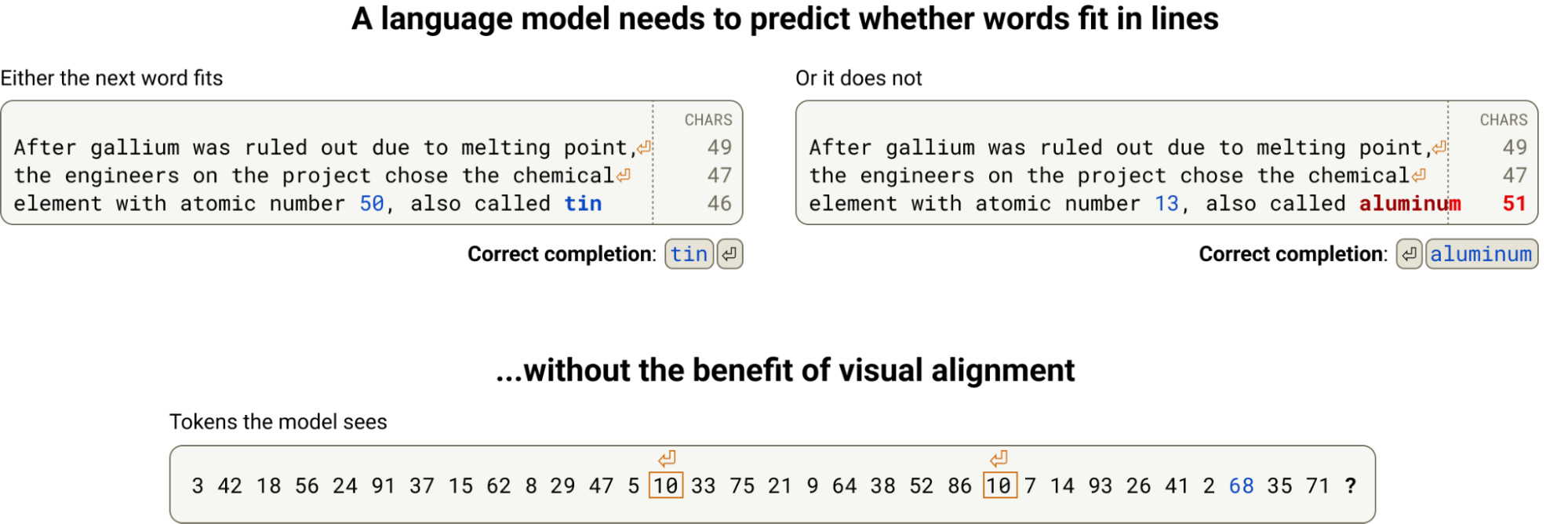}
    \label{fig:gdoc_2}
    \caption{A pair of prompts highlighting the difficulty of the linebreaking task.}
\end{figure}

To orient ourselves to the stages of the computation, we first studied the model using discrete dictionary features. In this frame, we can understand computation as an “attribution graph” \cite{ameisen2025circuit} where a cascade of features excite or inhibit each other.\footnote{We actually first tried to use patching and probing without looking at the graph as a kind of methodological test of the utility of features, but did not make much progress. In hindsight, we were training probes for quantities different than the ones the model represents cleanly, e.g., a fusion of the current token position and the line width.}

\begin{figure}[!htp]
    \centering
    \includegraphics[width=0.9\textwidth,height=0.7\textheight,keepaspectratio]{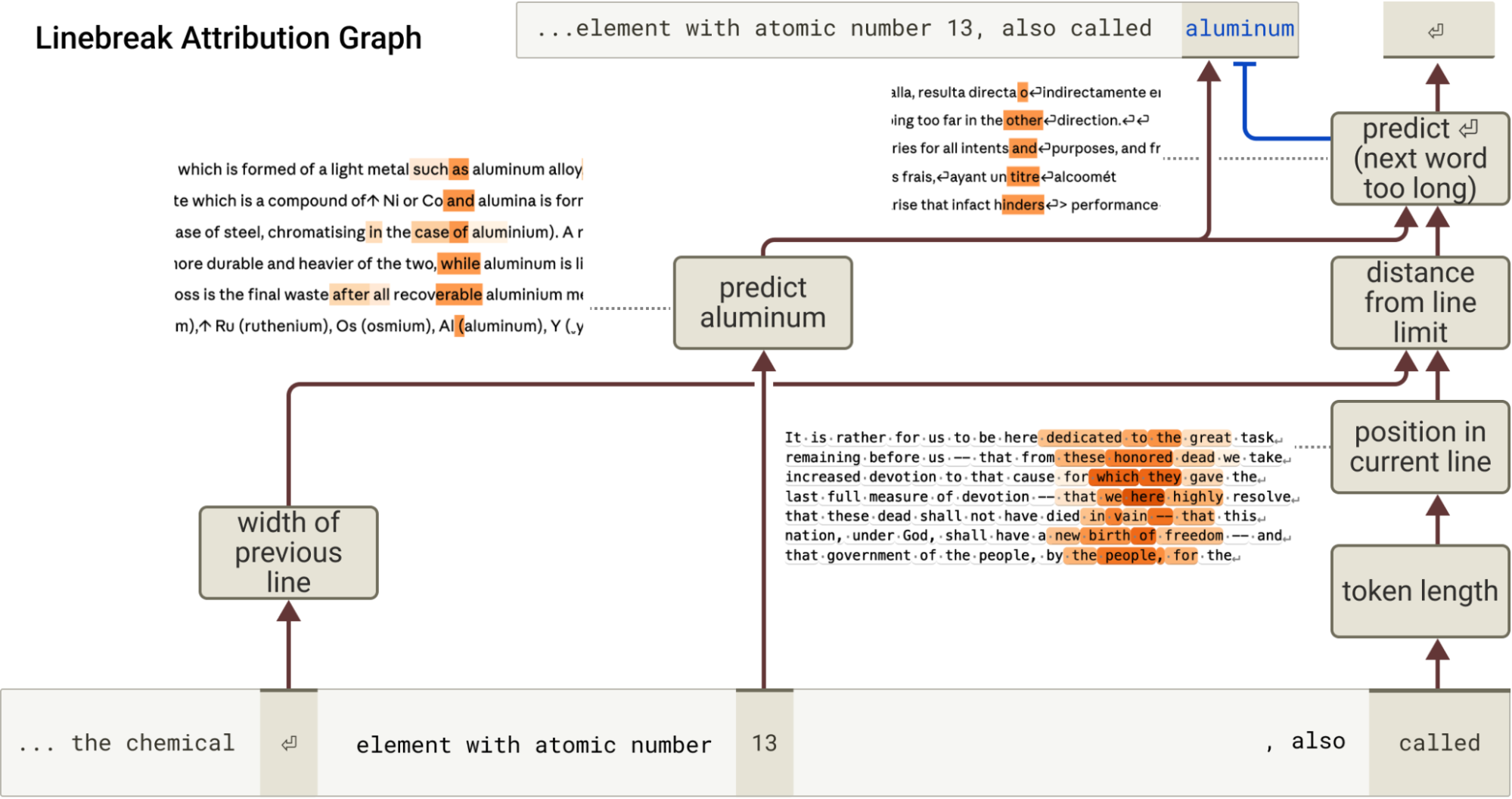}
    \label{fig:gdoc_3}
    \caption{Attribution Graph for Claude 3.5 Haiku’s prediction of a newline in the aluminum prompt. We see features relating to “width of the previous line” and “position in the current line” which together activate features for “distance from line limit”. Combined with features for the planned next word, these features activate “predict newline” features.}
\end{figure}

The attribution graph shows how the model performs this task by combining features that represent different concepts it needs to track:

\begin{enumerate}
    \item Features for the current position in the line (the character count) as well as features for the total line width (the constraint) are computed by accumulating features for individual token lengths.
    \item The model then combines these two representations --- current position and line width --- to estimate the distance from the end of the line, leading to “characters remaining” features.
    \item Finally, the model uses this estimate of characters remaining along with features for the planned next word to determine if the next word will fit on the line or not.
\end{enumerate}

The attribution graph provides a kind of execution trace of the algorithm, showing on this prompt which variables are computed and from what. After finding large feature families involved in representing these quantities across a diverse dataset, we suspected a simpler lens might be provided in terms of lower-dimensional feature manifolds interacting geometrically. We found geometric perspectives on the following questions:

\begin{figure}[!htp]
    \centering
    \includegraphics[width=0.9\textwidth,height=0.7\textheight,keepaspectratio]{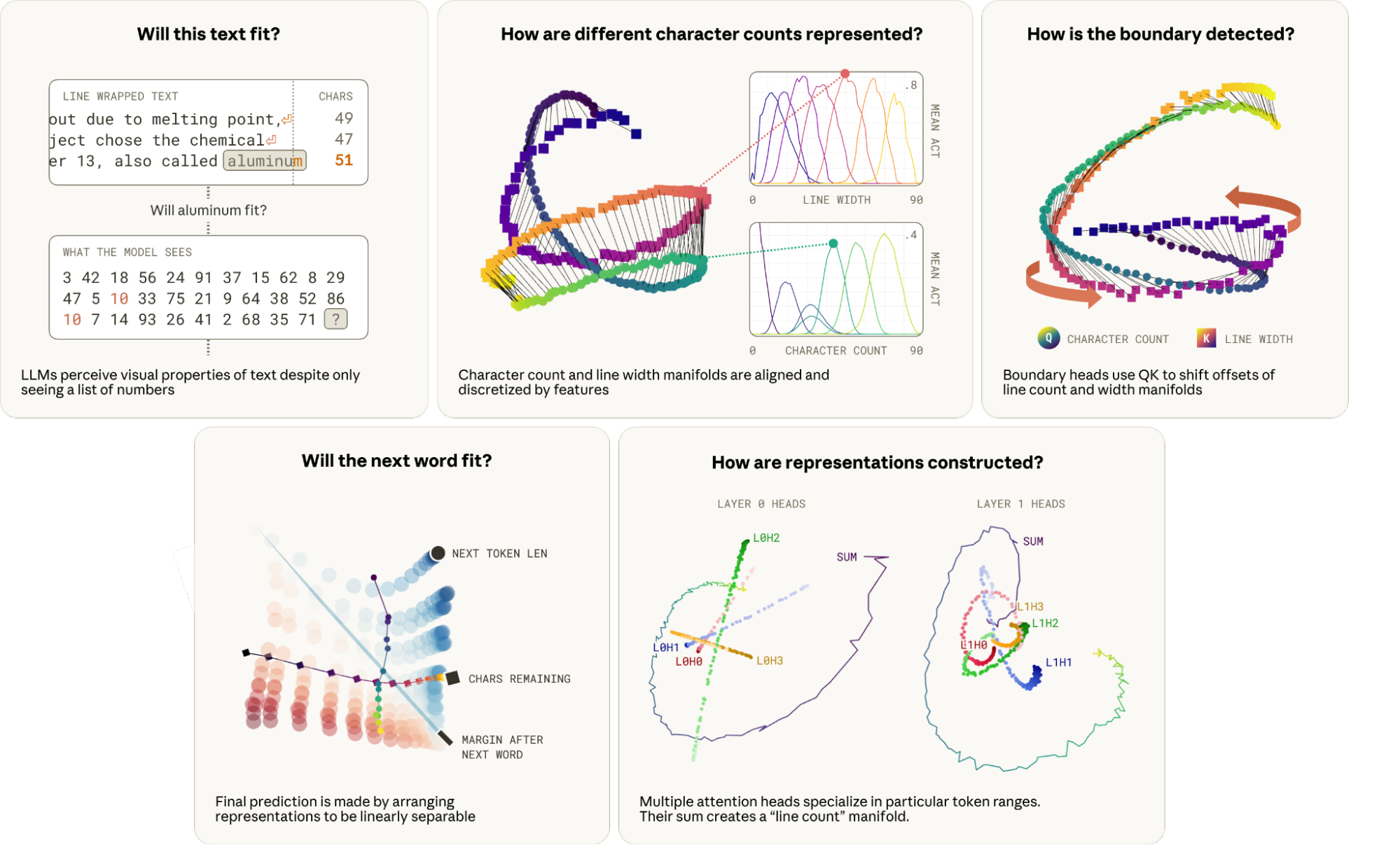}
    \label{fig:gdoc_4}
    \caption{Key steps in the linebreaking behavior can be described in terms of the construction and manipulation of manifolds.}
\end{figure}

\textbf{How does the model represent different counts?} The number of characters in a token, the number of characters in the current line, the overall line width constraint, and the number of characters remaining in the current line are each represented on 1-dimensional feature manifolds embedded with high curvature in low-dimensional subspaces of the residual stream. These manifolds have a dual interpretation in terms of discrete features, which tile the manifold in a canonical way, providing approximate local coordinates. Manifolds with similar geometry arise for a variety of ordinal concepts, and a ringing pattern we see in the embedded geometry in all these cases is optimal with respect to a simple physical model (§\hyperref[sec:char-count]{Representing Character Count}).\footnote{Ringing, in the manifold perspective, corresponds to interference in the feature superposition perspective.}

\textbf{How }\textbf{does the model detect the boundary?} To detect an approaching line boundary, the model must compare two quantities: the current character count and the line width. We find attention heads whose QK matrix rotates one counting manifold to align it with the other at a specific offset, creating a large inner product when the difference of the counts falls within a target range. Multiple heads with different offsets work together to precisely estimate the characters remaining (§\hyperref[sec:boundary]{Sensing the Line Boundary}).

\textbf{How does the model know if the next word fits? }The final decision --- whether to predict a newline --- requires combining the estimate of characters remaining with the length of the predicted next word. We discover that the model positions these counts on near-orthogonal subspaces, creating a geometric structure where the correct linebreak prediction is linearly separable (§\hyperref[sec:prediction]{Predicting the Newline}).

\textbf{How does the model construct these curved geometries? }The curvature in the character count representation manifold is produced by many attention heads working together, each contributing a piece of the overall curvature. This distributed algorithm is necessary because individual components cannot generate sufficient output variance to create the full representation (§\hyperref[sec:count-algo]{A Distributed Character Counting Algorithm}).

We validate these interpretations through targeted interventions, ablations, and “visual illusions” --- character sequences that hijack specific attention mechanisms to disrupt spatial perception (§\hyperref[sec:illusion]{Visual Illusions}).

Zooming out, we take several broader lessons from this mechanistic case study:

\textbf{When }\textbf{Models Manipulate Manifolds}. For representing a scalar quantity (e.g., integer counts from $1$ to $N$), it is inefficient to use $N$ orthogonal dimensions, and not expressive enough to use just one\footnote{Orthogonal dimensions would also not be robust to estimation noise.}. Instead models learn to represent these quantities on a feature manifold with intrinsic dimension 1 (the count) embedded in a subspace with extrinsic dimension $1 < d \ll N$ (e.g., \cite{gorton2024curve,engels2025not,modell2025origins}), in which the curve “ripples”. Such rippled manifolds optimally trade off capacity constraints (roughly, dimensionality) with maintaining the distinguishability of different scalar values (curvature). Our work demonstrates the intricate ways in which these manifolds can be manipulated to perform \textit{computation} and show how this can require distributing computation across multiple model components.

\textbf{Duality of}\textbf{ Features and Geometry}. Dictionary features provide an unsupervised entry point for discovering mechanisms, and attribution graphs surface the important features for any particular prediction. Sometimes, discrete features (and their interactions) can be equivalently described using continuous feature manifolds (and their transformations). In cases where it is possible to explicitly parameterize the manifold (as with the various integer counts we study), we can directly study the geometry, making some operations clearer (e.g., boundary detection). But this approach is expensive in researcher time and potentially limited in scope: it's straightforward when studying \textit{known} continuous variables but becomes difficult to execute correctly for more complex, difficult-to-parametrize concepts.

\textbf{Complexity Tax}. While unsupervised discovery is a victory in and of itself, dictionary features fragment the model into a multitude of small pieces and interactions -- a kind of complexity tax on the interpretation. In cases where a manifold parametrization exists, we can think of the geometric description as reducing this tax. In other cases, we will need additional tools to reduce the interpretation burden, like hierarchical representations \cite{costa2025flat} or macroscopic structure in the global weights \cite{olah2023interpretability}. We would be excited to see methods that extend the dictionary learning paradigm to unsupervised discovery of other kinds of geometric structures (e.g., those found in prior work \cite{hewitt2019structural,reif2019visualizing,chang2022geometry,wattenberg2024relational,park2024geometry,li2025geometry,wollschlager2025geometry,hindupur2025projecting,modell2025origins}).

\textbf{Natural Tasks}. The crispness of the representations and circuits we found was quite striking, and may be due to how well the model does the task. Linebreaking is an extremely natural behavior for a pretrained language model, and even tiny models are capable of it given enough context. Studying tasks which are natural for pretrained language models, instead of those of more theoretical interest to human investigators, may offer promising targets for finding general mechanisms.

\subsection{Preliminaries}

To enable systematic analysis, we created a synthetic dataset using a text corpus of diverse prose where we (1) stripped out all newlines and (2) reinserted newlines every $k$ characters to the nearest word boundary $\leq k$ for $k=15,20,\ldots,150$. As an example, here is the opening sentence of the Gettysburg Address, wrapped to $k=40$ characters, with the newlines shown explicitly.

\begin{promptblock}
Four score and seven years ago our[newline]
fathers brought forth on this continent,[newline]
a new nation, conceived in Liberty, and[newline]
dedicated to the proposition that all[newline]
men are created equal.
\end{promptblock}

Claude 3.5 Haiku is able to adapt to the line length for every value of $k$, predicting newlines at the correct positions with high probability by the third line (see \hyperref[sec:appendix-task-performance]{Appendix}).

All features in the main text of this paper are from a 10 million feature Weakly Causal Crosscoder (WCC) dictionary \cite{lindsey2024crosscoders} trained on Claude 3.5 Haiku. Feature activation values are normalized to their max throughout.

\section{Representing Character Count}\label{sec:char-count}

We define the \textit{line character count} (or \textit{character count}) at a given token in a prompt to be the total number of characters since the last newline, including the characters of the current token.

A natural thing to check is if the model linearly represents the character count as a quantitative variable: that is, can we predict character count with high accuracy via linear regression on the residual stream? Yes: a linear probe fit on the residual stream after layer 1 has an $R^2$ of 0.985. This success does not mean, however, that the model actually represents the character count along a single line.

Instead, we find a multidimensional representation of the character count that we will analyze from four perspectives:

\begin{enumerate}
    \item Sparse crosscoder features.\footnote{Each feature has an encoder, which acts as a linear + (Jump)ReLU probe on the residual stream, and a decoder. Ten features $f_1,\ldots,f_{10}$ are associated with line character count. The model's estimate of the character count, given a residual stream vector $x$, is summarized by the set of activities of each of the 10 features $\{f_i(x)\}$.}
    \item A low-dimensional subspace.\footnote{The model's estimate of the character count is summarized by the projection $\pi(x)$ of $x$ onto that subspace. Two datapoints have similar character counts if their projections are close in that subspace.}
    \item A continuous 1-dimensional manifold contained in that low-dimensional subspace.\footnote{The model's estimate of the character count is summarized by the nearest point on the manifold to the projection of $x$ into the subspace, and its confidence in that estimate by the magnitude of $\pi(x)$.}
    \item A set of 150 logistic probes (corresponding to values of line character count from 1 to 150).\footnote{The model's estimate of the character count is summarized by the probability distribution given by the softmax of the probe activities, softmax$(Px)$.}
\end{enumerate}

Each of these perspectives provides a complementary view of the same underlying object. The feature perspective is valuable for getting oriented, the subspace is perfect for causal intervention, the manifold is helpful for understanding how the representation is constructed and then manipulated to detect boundaries, and the logistic probes are useful for analyzing the OV and QK matrices of the individual attention heads involved.

\subsection{Character Count Features}

We begin with the features. In layers one and two, we found features that seemed to activate based on a token’s character position within a line. For example, in the attribution graph for the aluminum prompt, there were two features active on the final word “called” that seemed to fire when the line character count was between 35--55 and 45--65, respectively. To find more such features, we computed the mean activation of each feature binned by line character count. There were ten features with smooth profiles and large between-character-count variance, shown below:

\begin{figure}[!htp]
    \centering
    \includegraphics[width=0.9\textwidth,height=0.7\textheight,keepaspectratio]{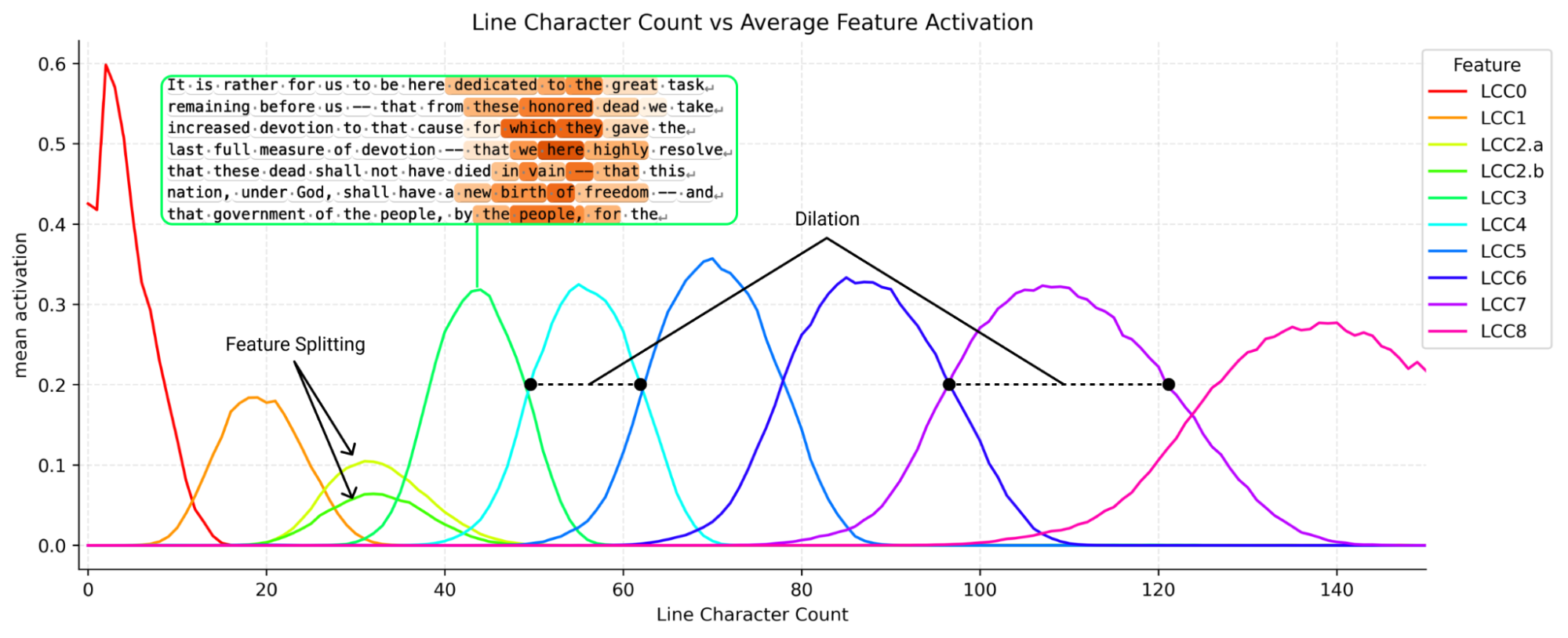}
    \label{fig:gdoc_5}
    \caption{A family of features representing the current character count in a line of text. The tuning curve of the features’ activity increases at larger line character counts.}
\end{figure}

We find these features especially interesting as they are quite analogous to curve-detector features in vision models \cite{cammarata2021curve,gorton2024missing} and place cells in biological brains \cite{Moser2008PlaceCG}. In all three of these cases, a continuous variable is represented by a collection of discrete elements that activate for particular value ranges. Moreover, we also observe \textit{dilation} of the receptive fields (i.e., subsequent features activate over increasingly large character ranges) which is a common characteristic of biological perception of numbers (e.g., \cite{dehaene2003neural,piazza2004tuning}).

In the \hyperref[sec:appendix-feature-splitting]{Appendix}, we show these features are universal across dictionaries of different sizes, but that some feature splitting occurs with respect to the line width constraint.

\subsection{The Model Represents Character Count on a Continuous Manifold}

We observe that character count feature activations rise and fall at an offset, with two features being active at a time for most counts. This pattern suggests that the features are reconstructing a curved continuous manifold, locally parametrized by the activity of the two most active features. Given that their joint activation profiles follow a sinusoidal pattern, we expect reconstructions to lie on a curve between adjacent feature decoders.

To visualize this, we first compute the average layer 2 residual stream for each value of line character count on our synthetic dataset. We compute the PCA of these 150 vectors, and find that the top 6 components capture 95\% of the variance; we project data to that 6 dimensional subspace which we call the “character count subspace” (top 3 PCs on the left below, next 3 PCs on the right). We observe the data form a twisting curve, resembling a helix from the perspective of PCs 1--3 and a more complex twist from the perspective of PCs 4--6.

We also reconstruct the residual stream for each datapoint using only the 10 character count features identified above, and compute the average \textit{reconstructed} residual stream. We project the resulting curve, along with the feature decoders, into the same subspace. We find that the average line character count vectors are quite closely approximated by the feature reconstruction, though with mild kinks near the feature vectors themselves, reminiscent of a spline approximation of a smooth curve. While the 10 feature vectors discretize the curve, interpolating between the 2--3 neighboring features which are active at a time allows for a high-quality reconstruction of 150 data points.

\begin{figure}[t]
    \centering
    \includegraphics[width=0.9\textwidth,height=0.7\textheight,keepaspectratio]{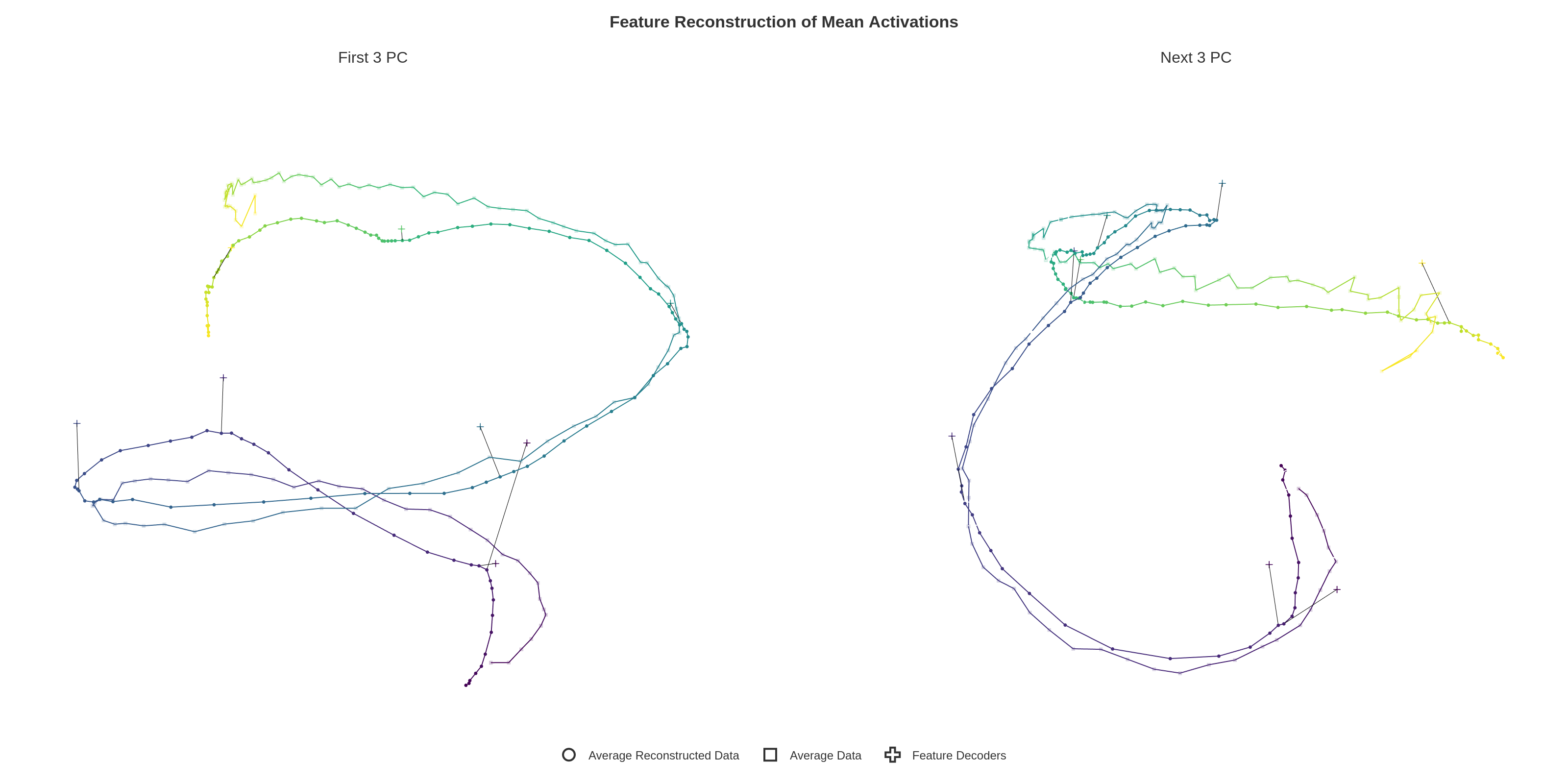}
    \label{fig:feature_reconstruction}
    \caption{Character count is represented on a manifold in a 6 dimensional subspace (jagged line). This manifold can be approximately locally parametrized by the features we identified (crosses).}
\end{figure}

\subsection{Validation: The Character Count Subspace is Causal}

To validate our interpretation of the character count subspace, we perform a coarse-grained ablation and a fine-grained intervention.

\textbf{Ablation Experiment.}  For our ablation experiment, we zero ablate (from a single early layer) a $k$-dimensional subspace corresponding to the top $k$ principal components of the per--character count mean activations and compare to a baseline of ablating a random $k$-dimensional subspace. Below we measure the loss effect, broken down by newlines and nonnewlines.\footnote{Note, in general one should not assume that a subspace spanned by features (or a PCA) is dedicated to those features because it could be in superposition with many other features. However, because in this case the character count subspace is densely active (and therefore less amenable to being in superposition), this experimental design is more justified.}

\begin{figure}[!htp]
    \centering
    \includegraphics[width=0.9\textwidth,height=0.7\textheight,keepaspectratio]{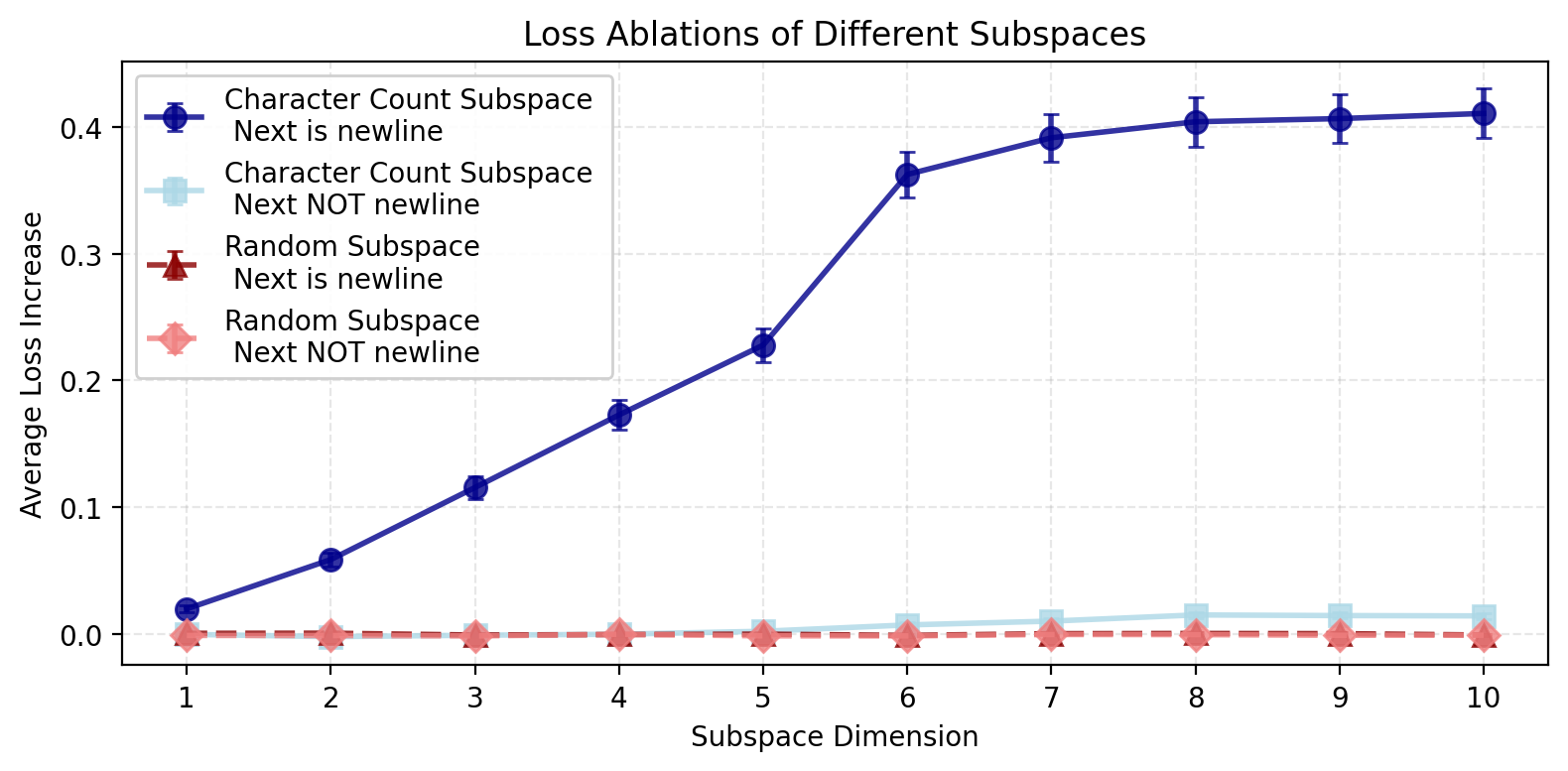}
    \label{fig:gdoc_7}
    \caption{Ablating the character count subspace has a large effect only when the next token is a newline.}
\end{figure}

\textbf{Intervention Experiment.  }As a more surgical intervention, we perform an experiment to modify the perceived character count at the end of the aluminum prompt (originally 42 characters). Specifically, we sweep over character counts $c$, and substitute the mean activation across all tokens in our dataset with count $c$. That is, $a_{\text{patched}} = a_{\text{original}} - \mu_{\text{original}} + \mu_{c}$ for activation $a$ and average activation matrix $\mu$. We perform this intervention for three adjacent early layers and the last two tokens for both the entire mean vector and within the 6 dimensional PCA space of the mean vectors.\footnote{The attribution graph has several positional features and edges on both the last token (“called”) as well as the second-to-last token (“also”). We change the “also” count representation to be 6 characters prior to that for the final token, to maintain consistency.}

\begin{figure}[!htp]
    \centering
    \includegraphics[width=0.9\textwidth,height=0.7\textheight,keepaspectratio]{figures/fig_008_gdoc.png}
    \label{fig:gdoc_8}
    \caption{Intervening on a rank 6 subspace is sufficient to change the model’s linebreaking behavior.}
\end{figure}

\subsection{The Probe Perspective}

We also train supervised logistic regression probes to predict character count.\footnote{as a 150-way multiclass classification problem} Probes trained after layer 1 achieve a root mean squared error of 5, indicating some intrinsic noise in the character count representation --- which is consistent with our features having relatively wide receptive fields. Performing PCA on the 150 probe weight vectors, we find that 6 components capture 82\% of the variance.

When we look at the average responses of each probe to tokens with different line character counts, we see a striking pattern. In addition to a diagonal band (probes, like the sparse features, have increasingly wide receptive fields), we see two faint off-diagonal bands on each side! The response curve of each probe is not monotonically decreasing away from its max, but rebounds. This “ringing” turns out to be a natural consequence of embedding a “rippled” manifold into low dimensions.

\begin{figure}[t]
    \centering
    \includegraphics[width=0.6\textwidth,keepaspectratio]{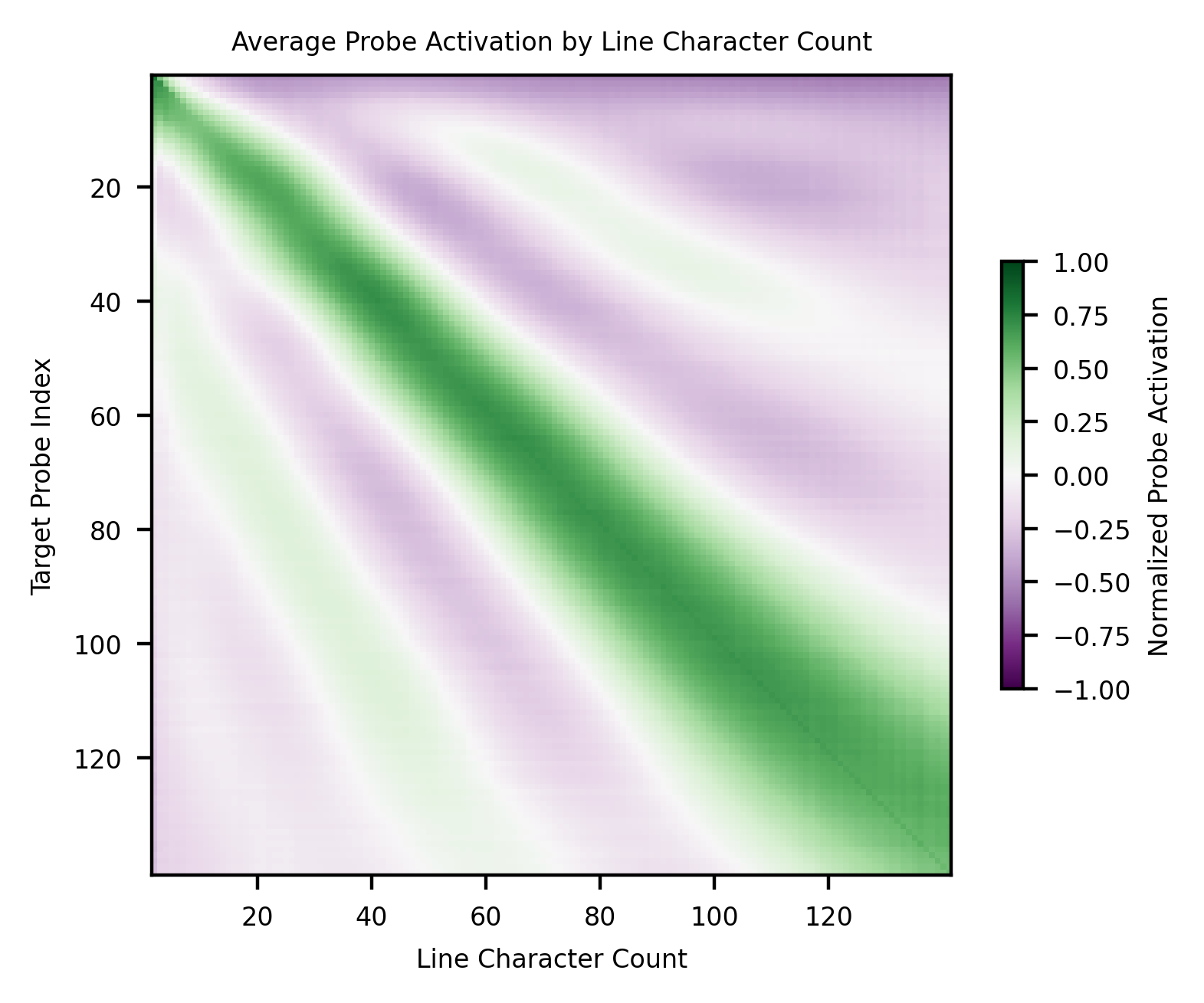}
    \label{fig:gdoc_9}
    \caption{Response curve of Line Character Count probes as a function of Line Character Count show widening receptive fields and a ``ringing'' pattern of off-diagonal stripes.}
\end{figure}

\subsection{Rippled Representations are Optimal}\label{sec:rippled-representations}

We note that the cosine similarities of the mean activation vector (which form the helix-like curve visualized in PCA space above), the linear probe vectors, and feature decoder vectors all exhibit similar ringing patterns to the above figure.\footnote{We use the term “\href{https://en.wikipedia.org/wiki/Ringing_(signal)}{ringing}” in the sense of signal processing, a transient oscillation in response to a sharp peak, such as in the Gibbs Phenomenon).} Note that not only are neighboring features not orthogonal, features further away have negative similarities, and then those even further away have positive ones again.

\begin{figure}[!htp]
    \centering
    \includegraphics[width=0.9\textwidth,height=0.7\textheight,keepaspectratio]{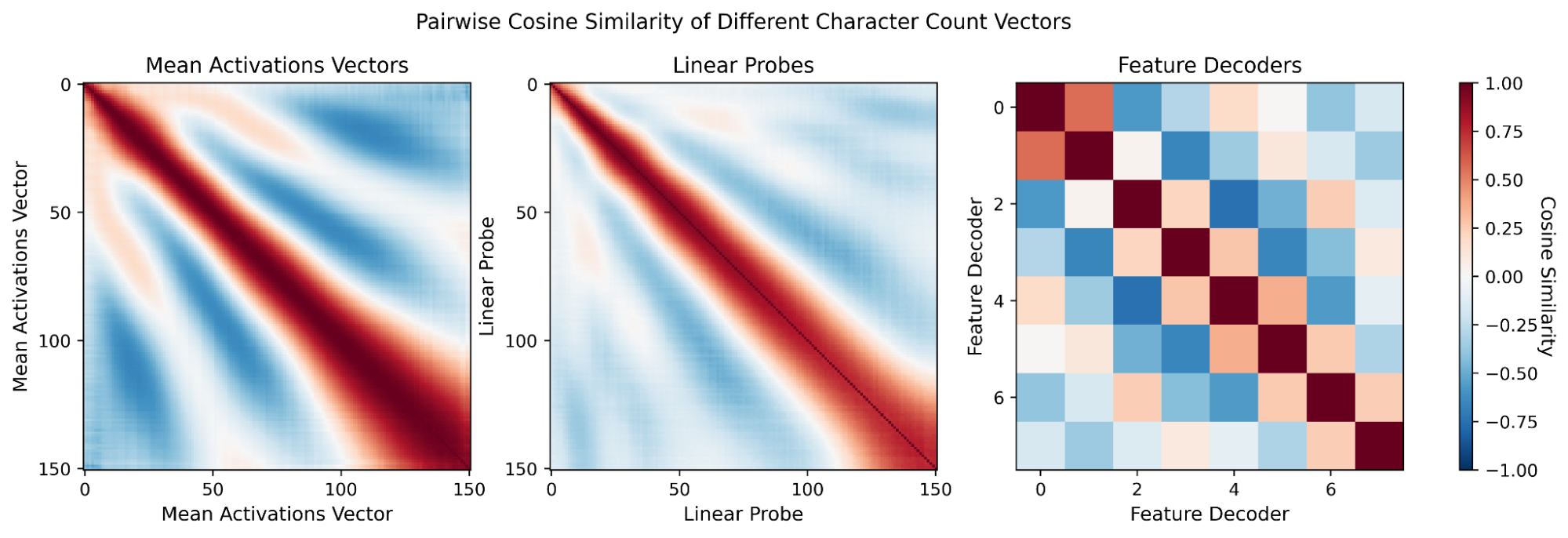}
    \label{fig:gdoc_10}
    \caption{The pairwise cosine similarity of different character count vectors. (Left) mean activation vectors, (center) linear probes, (right) feature decoders.}
\end{figure}

This structure turns out to be a natural consequence of having the desired pattern of similarity, trivially achievable in 150 dimensions, projected down to low dimensions. As a toy model of this, suppose that we wish to have a discretized circle's worth of unit vectors, each similar to its neighbors but orthogonal to those further away. This can be realized by a symmetric set of unit vectors in 150 dimensions with cosine similarity matrix $X$ pictured below (left). Projecting this to its top 5 eigenvectors yields a 5-dimensional embedding of the same vectors with cosine similarity matrix (below right) exhibiting ringing. We also plot the curve these vectors form in the top 3 eigenvectors. We can think of the original 150-dimensional embedding of the circle as being highly curved, and the resulting 5-dimensional embedding as retaining as much of that curvature as possible. This manifests as ripples in the embedding of the circle when viewed in a 3D projection. A relationship of this construction to Fourier features is discussed in the \hyperref[sec:appendix-gibbs]{appendix}.

\begin{figure}[!htp]
    \centering
    \includegraphics[width=0.9\textwidth,height=0.7\textheight,keepaspectratio]{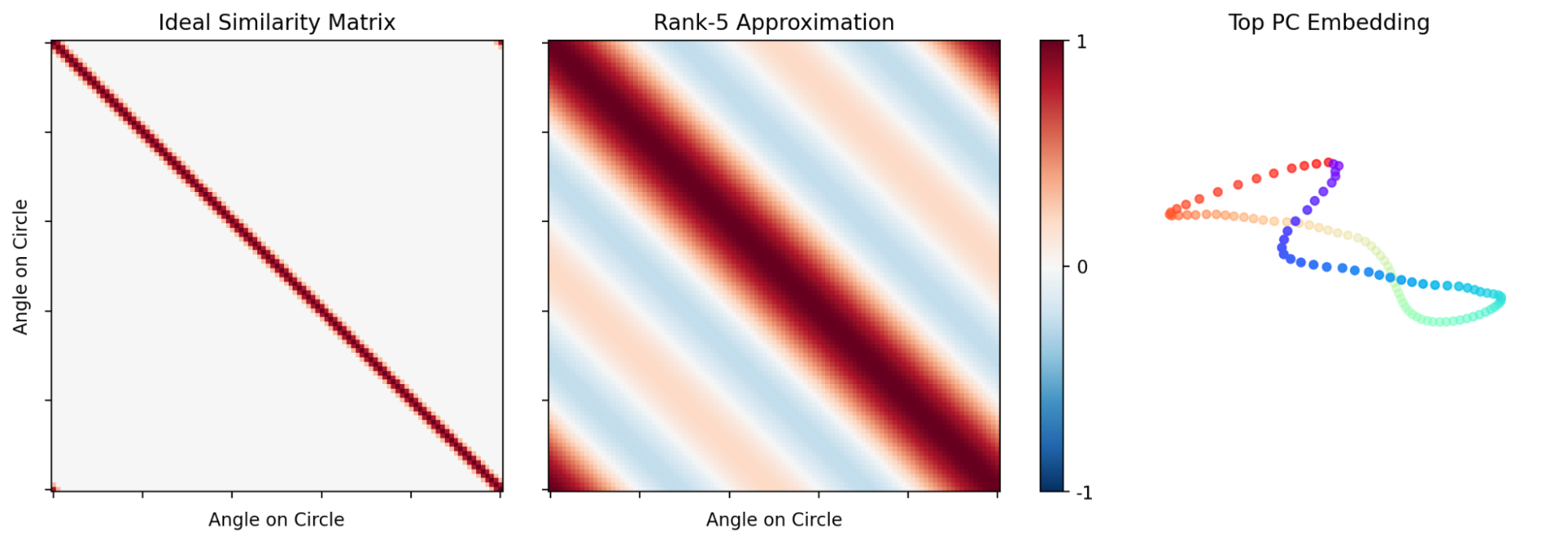}
    \label{fig:gdoc_11}
    \caption{Left panel shows an ideal similarity matrix for vectors representing points along a circle. Middle panel shows the optimal (PCA) approximation possible when embedding the points in 5 dimensions. Right panel shows the resulting projection of the circle to the top 3 dimensions, exhibiting rippling.}
\end{figure}

Alternatively, one can view the ringing from the perspective of sparse feature decoders as a kind of interference weight \cite{olah2025toy}. With no capacity constraints, the model might use orthogonal vectors to represent the quantitative response of each feature, with its own receptive field, to the input data. Forced to put them into lower dimensional superposition, the similarity matrix picks up both a wider diagonal stripe and the upper/lower diagonal ringing stripes.

Finally, we also construct a simple physical model showing that the rippling and ringing arise even when the solution is found dynamically, whenever many vectors are packed into a small number of dimensions. We show the result of a simulation in which 100 points confined to a $6$-dimensional hypersphere are subjected to attractive forces to their 6 closest neighbors on each side (matching the RMSE error of our probes) and repulsive forces to all other points. (To avoid boundary conditions, we use the topology of a circle instead of an interval.) On the right below is a heatmap exhibiting two rings, and on the left is a 3-dimensional projection of the 6-dimensional curve. There exists an interactive version in the web version of this paper \href{https://transformer-circuits.pub/2025/linebreaks/index.html#rippled-representations}{here}, and the reader is encouraged to experiment with reinitializing the points, switching the ambient dimension, and modifying the width of the attractive zone. Decreasing the attractive zone or increasing the embedding dimension both increase curvature (and the amount of ringing), and vice versa.\footnote{The simulation can sometimes find itself in local minima. Increasing the width of the attractive zone before decreasing it again usually solves this issue.} As the number of points on the curve grows and the attractive zone width shrinks (in relative terms), the curvature grows quite extreme, approaching a space-filling curve in the limit.

\begin{figure}[!htp]
    \centering
    \includegraphics[width=0.8\textwidth,height=0.7\textheight,keepaspectratio]{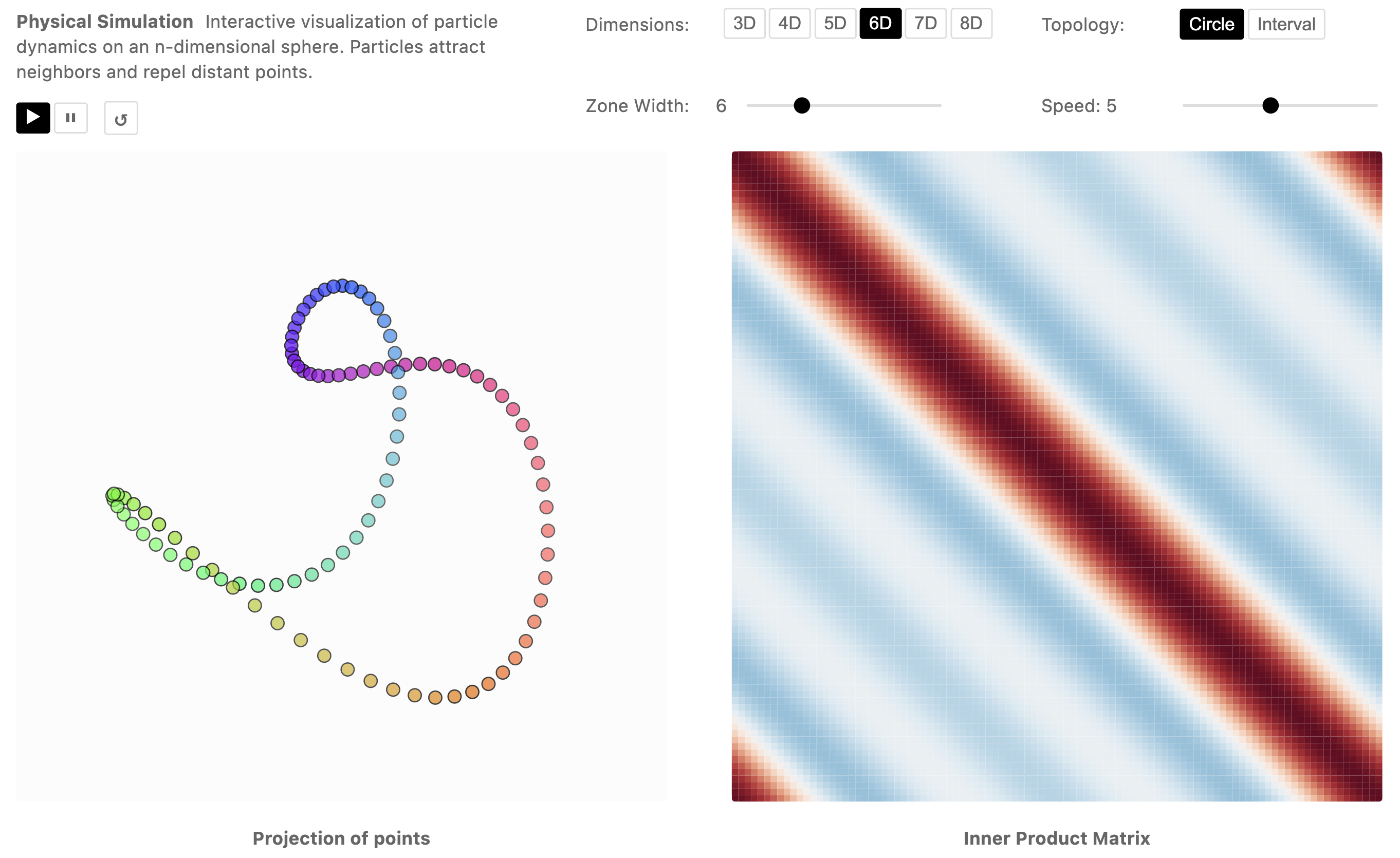}
    \label{fig:dynamical_simulation}
    \caption{Screenshot of our interactive particle simulation.}
\end{figure}

Of particular interest is the result from setting the ambient dimension to 3\footnote{Optimization in dimension 3, unlike in higher dimensions, admits bad local minima, because a generic curve on the surface of a sphere self-intersects. To avoid this, either increase the zone width until you get a great circle, then decrease it, or do the optimization in 4D, then select 3D.}: the result is a curve similar to the seams of a baseball (below left, circle), which matches the topology observed for three intrinsically one-dimensional phenomena observed in \cite{modell2025origins,engels2025not}, of colors by hue, dates of the year, and years of the 20th century (which also exhibit dilation). Similar ripples were predicted to occur by Olah \cite{olah2023manifold} and then observed by Gorton \cite{gorton2024curve} in curve detector features in Inception v1. One of the earliest observations of ringing in a cosine similarity plot and rippled spiral/helix shape in a low-dimensional embedding was of the learned positional embeddings of tokens in GPT2 \cite{yedidia2023gpt2,yedidia2023positional}. We also find similar structure in other representations, which we study in \hyperref[sec:appendix-sensory]{More Sensory and Counting Representations} in the appendix.

\begin{figure}[!htp]
    \centering
    \includegraphics[width=0.9\textwidth,height=0.7\textheight,keepaspectratio]{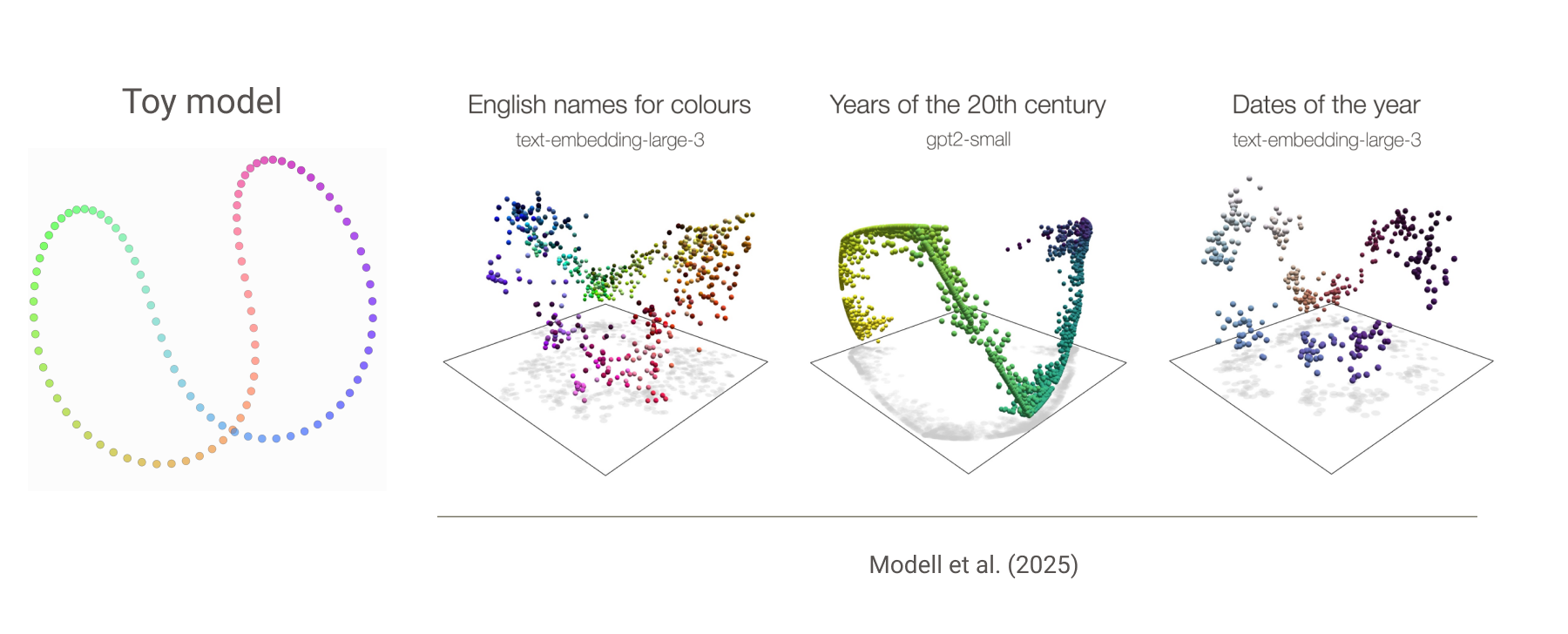}
    \label{fig:gdoc_13}
    \caption{Left curve is a locally optimal high-curvature embedding of the circle onto the 2-sphere. Right figures, reproduced with permission from \textit{Modell et al.} \cite{modell2025origins}, show 3-dimensional PCA projections of data or features related to colours, years, and dates.}
\end{figure}

\section{Sensing the Line Boundary}\label{sec:boundary}

We now study how the character counting representations are used to determine if the current line of text is approaching the\textit{ line boundary}. To detect the line boundary, the model needs to (1) determine the overall \textit{line width} constraint and (2) compare the current character count with the line width to calculate the characters remaining.

\subsection{Twisting with QK}

We find that newline tokens have their own dedicated character \hyperref[sec:appendix-width-features]{counting features} that activate based on the width of the line, counting the number of characters between adjacent newlines.

To better understand how these representations are related, we train 150 probes for each possible value of “Line Width” like we did for “Character Count”. Using the attribution graph, we identify an attention head which activates boundary detection features. We visualize both sets of counting representations directly using the first 3 components of their joint PCA in the residual stream (left) and in the reduced QK space of this \textit{boundary} \textit{head} (right).\footnote{Specifically we multiply the line width probes through $W_K$ and the character count probes through $W_Q$, and plot the points in the 3D PCA basis of their joint embedding.}

\begin{figure}[!htp]
    \centering
    \includegraphics[width=\textwidth,height=0.7\textheight,keepaspectratio]{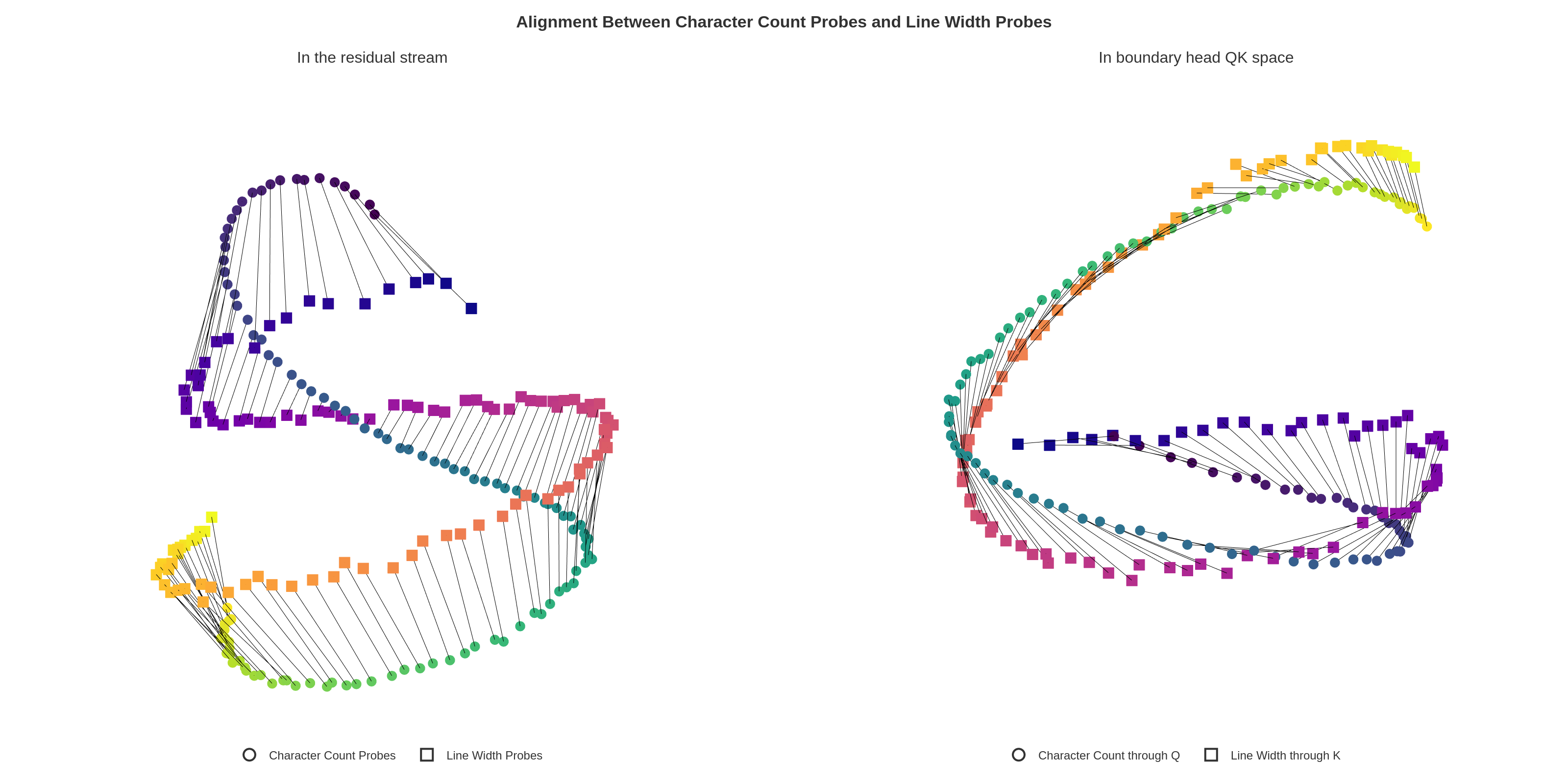}
    \label{fig:resid_vs_head_twist}
    \caption{Boundary heads twist the representation of line width and character count to detect the line boundary. Left: Joint PCA of character count and line width probes. Right: Same after multiplying them through the corresponding QK weights of the boundary head. Range is from 40 (dark) to 150 (light).}
\end{figure}

We find that this attention head “twists” the character count manifold such that character count $i$ is aligned with line width $k=i+\epsilon$. This causes the head to attend to the newline when the character count is just a bit \textit{less} than the line width, thereby indicating that the boundary is approaching. This algorithm is quite general, and enables this head to detect approaching line boundaries for arbitrary line widths!\footnote{This algorithm also generalizes to arbitrary kinds of separators (e.g., double newlines or pipes), as the QK circuit can handle the positional offset independently of the OV circuit copying the separator type.}

\begin{figure}[!htp]
    \centering
    \includegraphics[width=\textwidth,height=0.7\textheight,keepaspectratio]{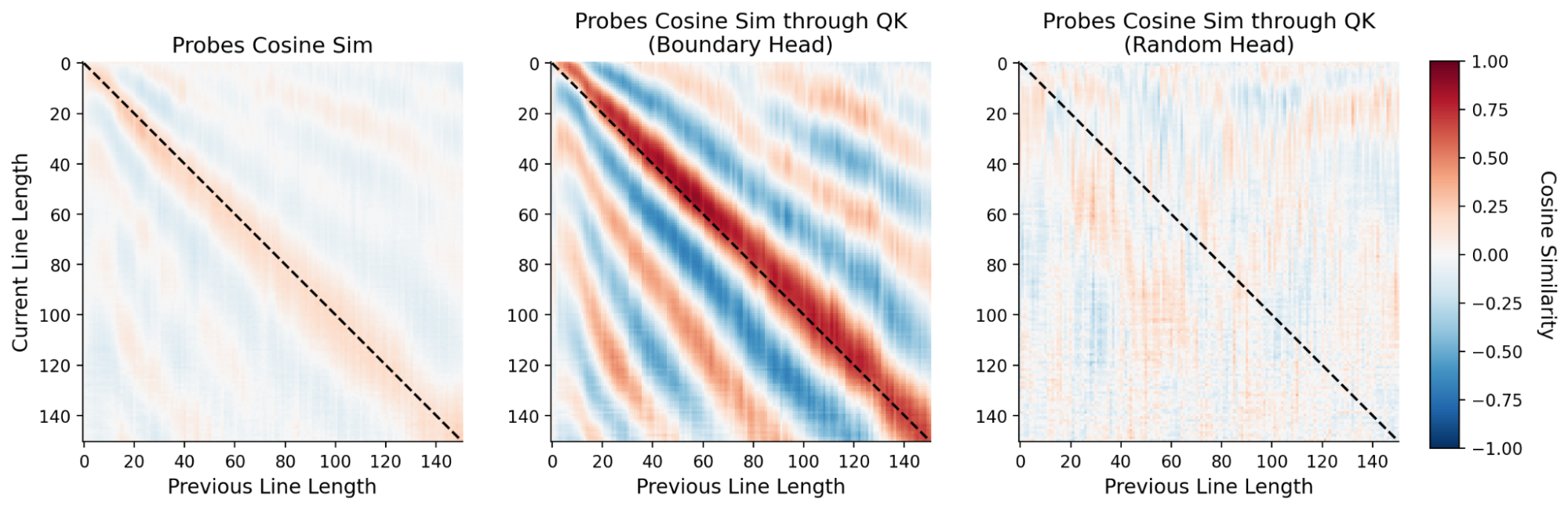}
    \label{fig:gdoc_15}
    \caption{Cosine similarity of previous line width and character count probes through different transforms. (Left) the identity map, (Center) QK of boundary head, (Right) QK of random head in the same layer. Boundary heads align the probes but with a small offset.}
\end{figure}

This plot shows that

\begin{itemize}
    \item In the residual stream, probes for character count $i$ are maximally aligned with probes line width probes $k$ when $i=k$, but are not highly aligned in absolute terms -- the maximum cosine sim is $\sim$0.25.
    \item In the QK space of the boundary head, the probes are maximally aligned on the offdiagonal $i < k$, and are almost perfectly aligned in absolute terms -- the maximum cosine sim is $\approx 1$.
    \item In the QK space of a random head, there is almost no structure between the probes.
\end{itemize}

As a consequence of the ringing in the character count representations, we also observe ringing in the inner products (see \hyperref[sec:rippled-representations]{Rippled Representations are Optimal} above). The model is robust to these off-diagonal interference terms via the softmax applied to attention scores.

\subsection{Leveraging Multiple Boundary Heads}

We find that the model actually uses multiple boundary heads, each twisting the manifolds by a different offset to implement a kind of “stereoscopic” algorithm for computing the number of characters remaining.\footnote{There are also multiple sets of boundary heads at multiple layers that usually come in sets of $\sim$3 with similar relative offsets (so not actually “stereo”).} We attach more visualizations of boundary heads in the \hyperref[sec:appendix-twisting]{Appendix}.

\begin{figure}[!htp]
    \centering
    \includegraphics[width=\textwidth,height=0.7\textheight,keepaspectratio]{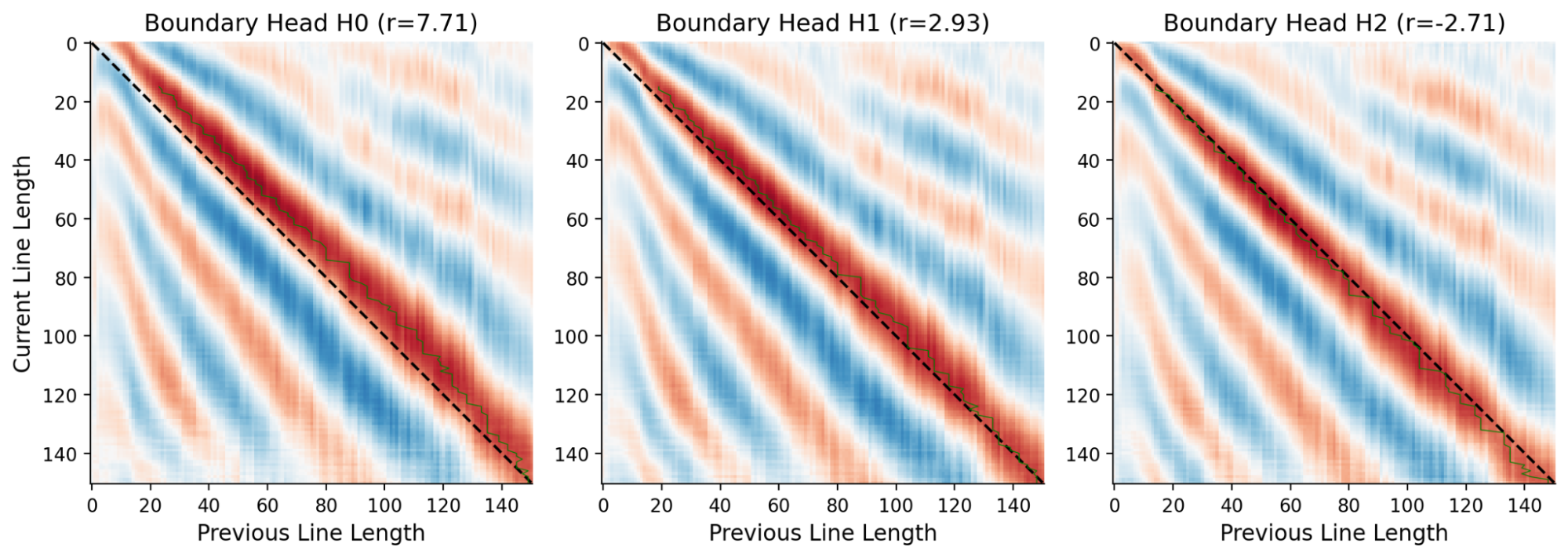}
    \label{fig:gdoc_16}
    \caption{Cosine similarity of line width and character count probes through three different boundary heads in the same layer with different amounts of twisting. Green line indicates the argmax for each row, and is used to calculate the average offset reported in the subtitles.}
\end{figure}

To better understand each boundary head’s output, we train a set of probes for each value of characters remaining in the line (i.e., the line width $k$ minus the character count $i$, restricted to $k - i < 40$). For each boundary head, we show the proportion of attention on the newline, as well as the norm of each head’s output projected onto the probe space as a function of characters remaining.

As predicted by our weights based analysis, we observe that boundary heads have distinct but overlapping response curves that “tile” the possible values of characters remaining.

\begin{figure}[!htp]
    \centering
    \includegraphics[width=0.9\textwidth,height=0.7\textheight,keepaspectratio]{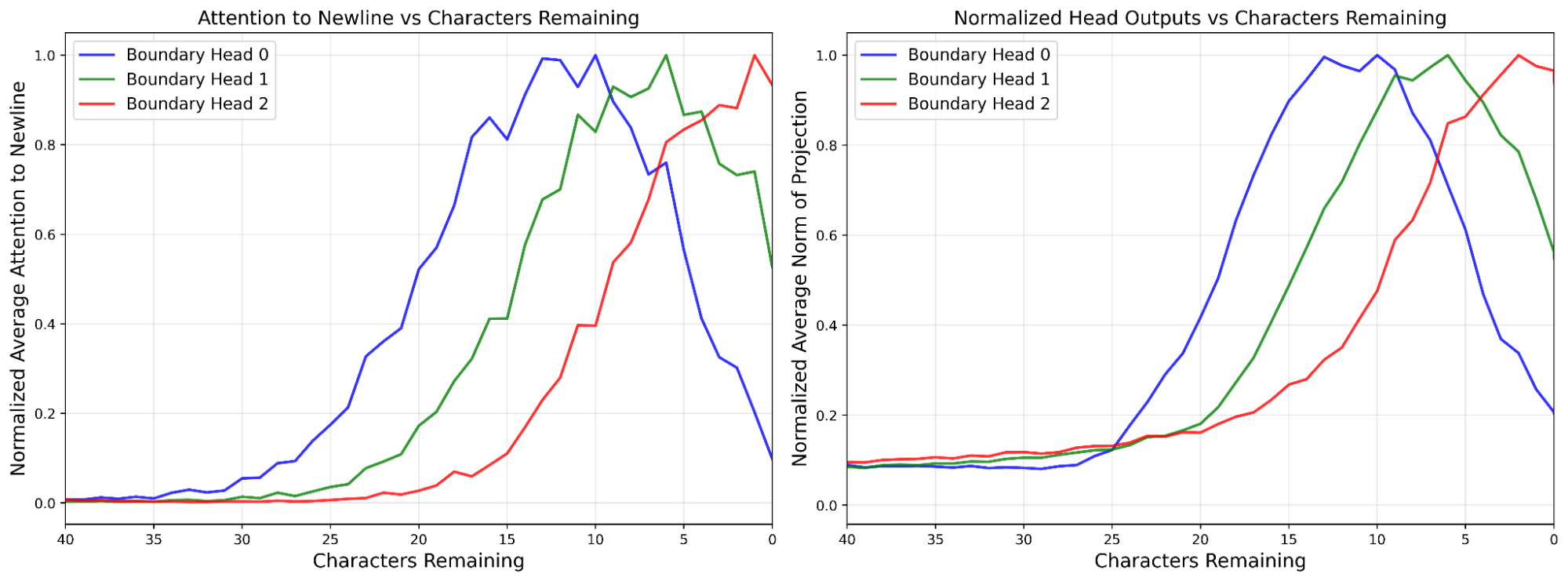}
    \label{fig:gdoc_17}
    \caption{Each boundary head’s response curve peaks at a different distance from the end of the line.}
\end{figure}

It's worth understanding why the model needs multiple boundary heads rather than just one. If the model relied only on boundary head 0, it couldn't distinguish between 5 characters remaining and 17 characters remaining---both would produce similar outputs. By having each head's output vary most significantly in different ranges, their sum achieves high resolution across the entire relevant range of “Characters Remaining” values.

We can see this more clearly by plotting each head's output in the first two principal components of the characters remaining space (which captures 92\% of the variance). Head 0 shows large variance in the [0, 10] and [15, 20] ranges, Head 1 varies most in the [10, 20] range, and Head 2 varies most in the [5, 15] range. While no single head provides high resolution across the entire curve, their sum produces an evenly spaced representation that covers all values effectively.

\begin{figure}[!htp]
    \centering
    \includegraphics[width=0.9\textwidth,height=0.7\textheight,keepaspectratio]{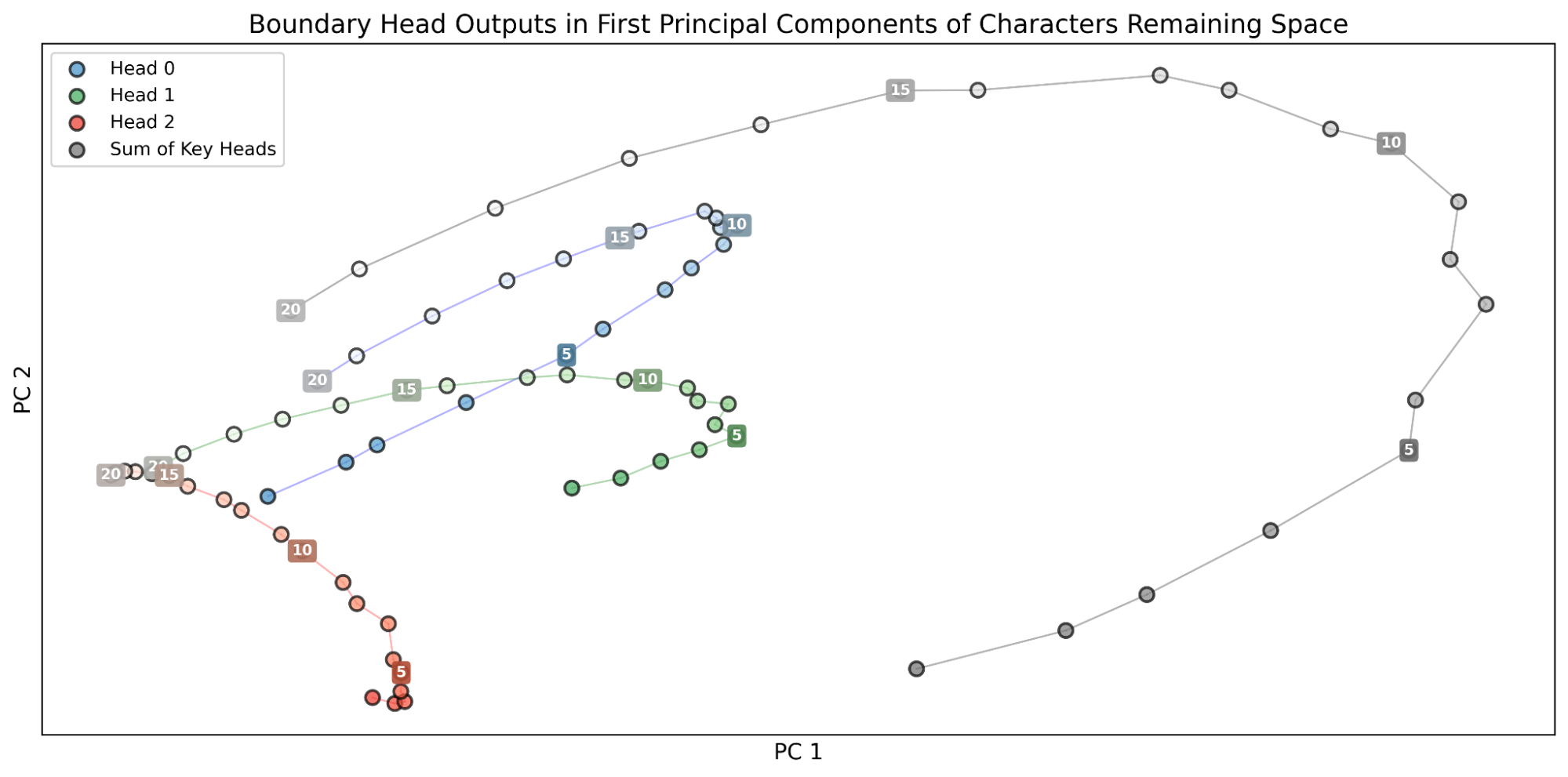}
    \label{fig:gdoc_18}
    \caption{Each head’s output as a function of characters remaining, and their sum in the PCA basis. Individual head outputs are almost one-dimensional, while the sum is a two-dimensional curve.}
\end{figure}

We validate the causal importance of this two-dimensional subspace by performing an ablation and intervention experiment. Specifically, we conduct the same experiments as before: ablate the subspace and measure its effect on loss by token (left) and precisely modulate the characters remaining estimate on the last token in the aluminum prompt by substituting mean activation vectors.

\begin{figure}[!htp]
    \centering
    \includegraphics[width=0.9\textwidth,height=0.7\textheight,keepaspectratio]{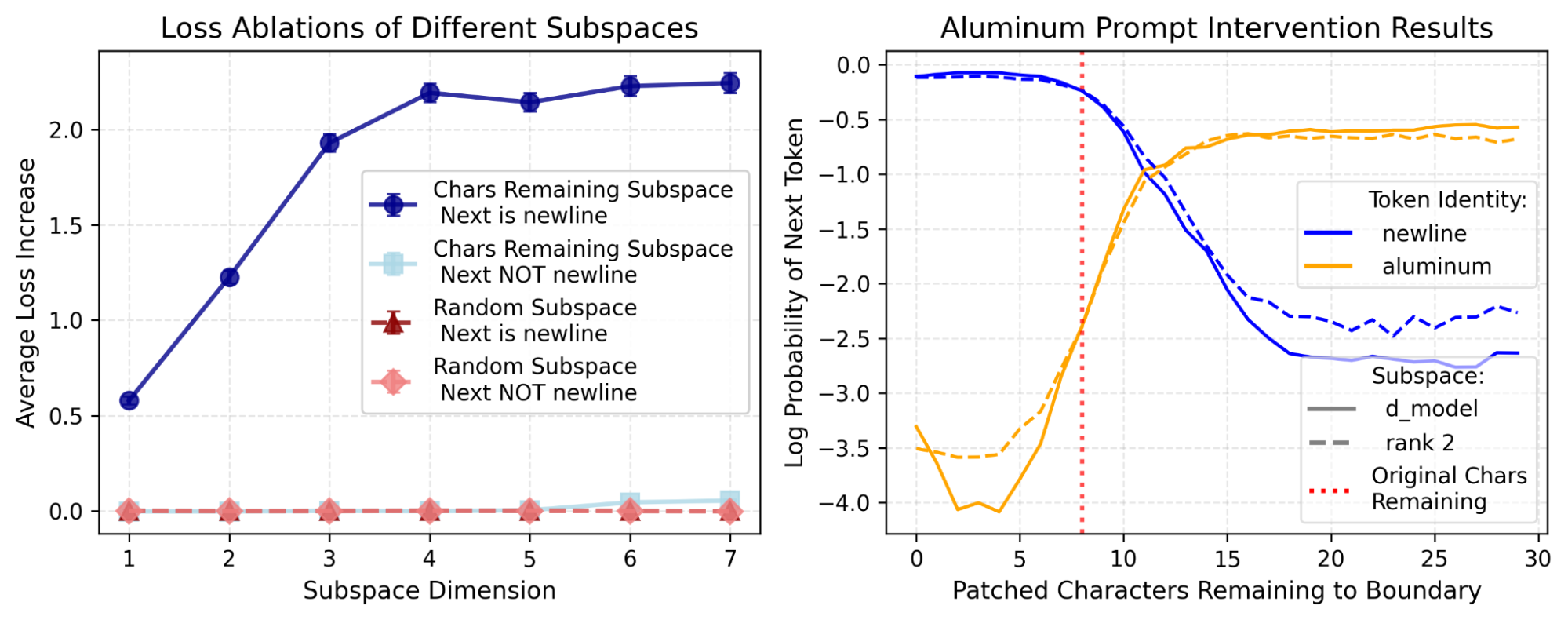}
    \label{fig:gdoc_19}
    \caption{Characters remaining subspace can be causally intervened upon. (Left) Ablating the subspace has a large effect only when the next token is a newline. (Right) We surgically intervene on the characters remaining space to modulate the prediction of the newline by subtracting the true characters remaining mean activation and adding in a patched characters remaining activations. Note that the completion “ aluminum.” requires ten characters to fit.}
\end{figure}

\subsection{The Role of the Extra Dimensions}\label{sec:one-dim}

We are now in a position to understand two distinct but related questions: (1) why these counting representations are multidimensional and (2) why multiple attention heads are required to compute these multidimensional representations.

\textbf{Geometric Computations} -- A multi-dimensional representation enables the model to rotate position encodings using linear transformations---something impossible with one-dimensional representations. For instance, to detect an approaching line boundary, the model can rotate the position manifold to align with line width, then use a dot product to identify when only a few characters remain. With a 1D encoding, linear operations reduce to scaling and translation, so comparing position against line width would just multiply the two values, producing a monotonically increasing result with no natural threshold. Higher dimensions beyond 2D allow the manifold to pack more information through additional curvature.

\textbf{Resolution} -- For character counting, the model must distinguish between adjacent counts for a large range of character positions, as this determines whether the next word fits. In a one-dimensional representation, positions would be arranged along a ray, with each position separated by some constant $\delta$. To reliably distinguish adjacent positions above noise, we need $||v_{42} - v_{41}|| = \delta$ to exceed some threshold. But with 150+ positions to represent, this creates an untenable choice: either use enormous dynamic range ($||v_{150}|| \gg ||v_1||$), which is problematic for transformer computations, or sacrifice resolution between adjacent positions. (Normalization blocks only exacerbate this effect: while points can be spaced far away on a ray if their norms get large enough, there is at most $\pi$ worth of angular distance along the projection of that ray onto the unit hypersphere.) Embedding the curve into higher dimensions solves this: positions maintain similar norms while being well-separated in the ambient space, achieving fine resolution without norm explosion (See \hyperref[sec:rippled-representations]{Rippled Representations are Optimal} above.) For counting the characters remaining, the dynamic range is smaller, and so the model is able to embed the representation in a smaller subspace as a result.

To achieve the curvature for necessary high resolution, multiple attention heads are needed to cooperatively construct the curved geometry of the counting manifold. An individual attention head's output is a linear combination of its inputs (weighted by attention and transformed by the OV circuit), and thus is fundamentally constrained by the curvature already present in those inputs. In the absence of MLP contributions to the counting representation, if the output manifold needs to exhibit substantial curvature, multiple attention heads need to coordinate---each contributing a piece of the overall geometric structure. We will see another example of distributed head computation in the section on the \hyperref[sec:count-algo]{Distributed Character Counting Algorithm}.

\subsection{A Discovery Story}\label{sec:discovery}

How did we originally find this boundary detection mechanism? When we first computed an attribution graph, we saw several edges from the previous newline features and embedding to predict-newline features. QK attributions showed that the top key feature was a “the previous line was 40--60 characters long” feature and the top query feature was “the current character count is 35--50” feature. At any one time there were often multiple counting features active at different strengths, suggesting that these features might be discretizing a manifold.

\begin{figure}[!htp]
    \centering
    \includegraphics[width=0.9\textwidth,height=0.7\textheight,keepaspectratio]{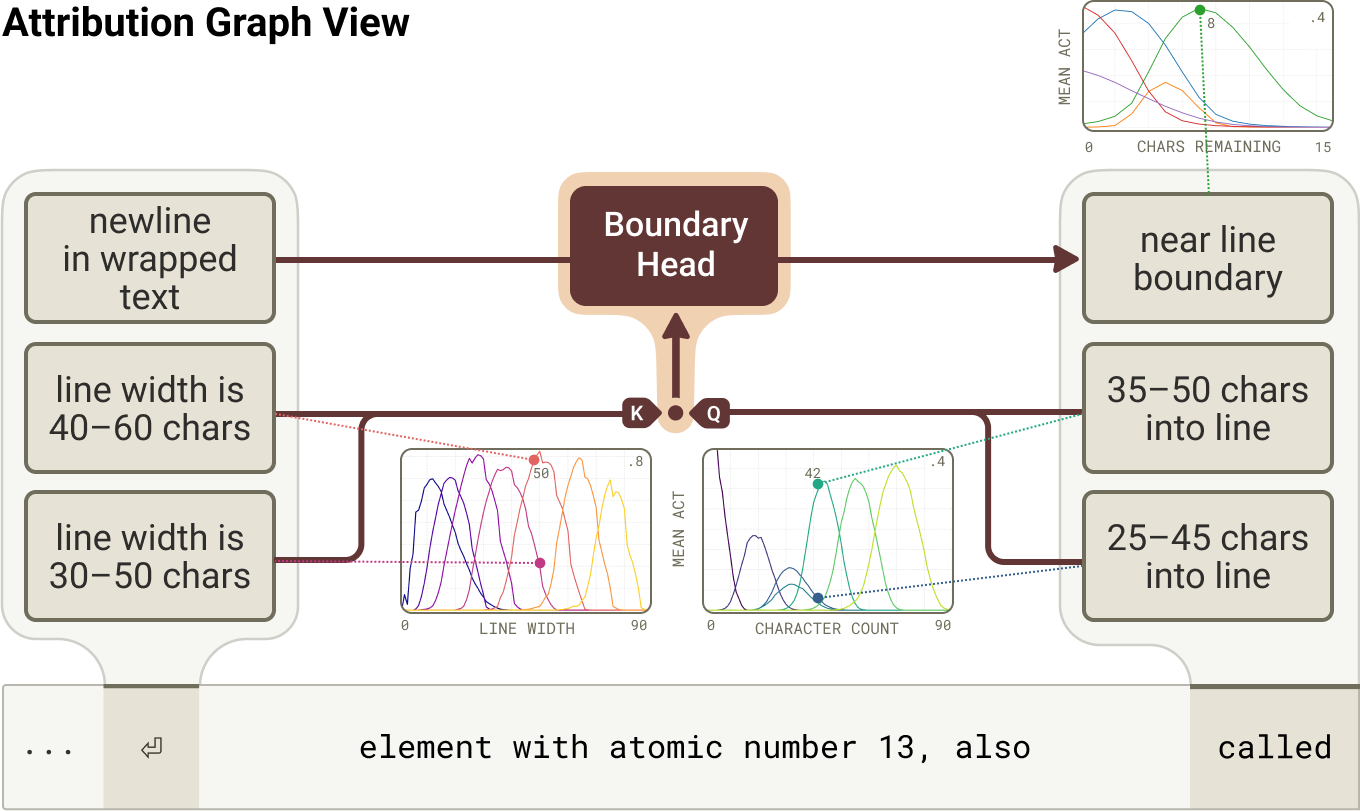}
    \label{fig:gdoc_20}
    \caption{Attribution graph representation of the boundary detection subcircuit.}
\end{figure}

The boundary heads cause a family of boundary detecting features to activate in response to how close the current line is to the global line width. That is, they sense the approaching line boundary or the reverse index of the line count. Investigating these three sets of feature families led us to the count manifolds which they sparsely parametrize, and investigating the relevant attention heads let us find the boundary heads.

Finally, we note that these boundary-sensing representations parallels a well-studied phenomenon in neuroscience: boundary cells \cite{solstad2008representation}, which activate at specific distances from environmental boundaries (e.g., walls). Both the artificial features and biological cells come in families with varied receptive fields and offsets.

\section{Predicting the Newline}\label{sec:prediction}

The final step of the linebreak task is to combine the estimate of the line boundary with the prediction of the next word to determine whether the next word will fit on the line, or if the line should be broken.

In the attribution graph for the aluminum prompt, we see exactly this merging of paths. The most influential feature\footnote{Influence in the sense of influence on the logit node, as defined in Ameisen et al. \cite{ameisen2025circuit}} in the entire graph is a late feature that activates in contexts where the next word would cause the current line to exceed the overall line width. For our prompt, this feature upweights the probability of newline and downweights the probability of “aluminum.” The top two inputs to this break predictor feature are a “say aluminum” feature and “boundary detecting” feature that gets activated by the aforementioned boundary head.

\begin{figure}[!htp]
    \centering
    \includegraphics[width=0.9\textwidth,height=0.7\textheight,keepaspectratio]{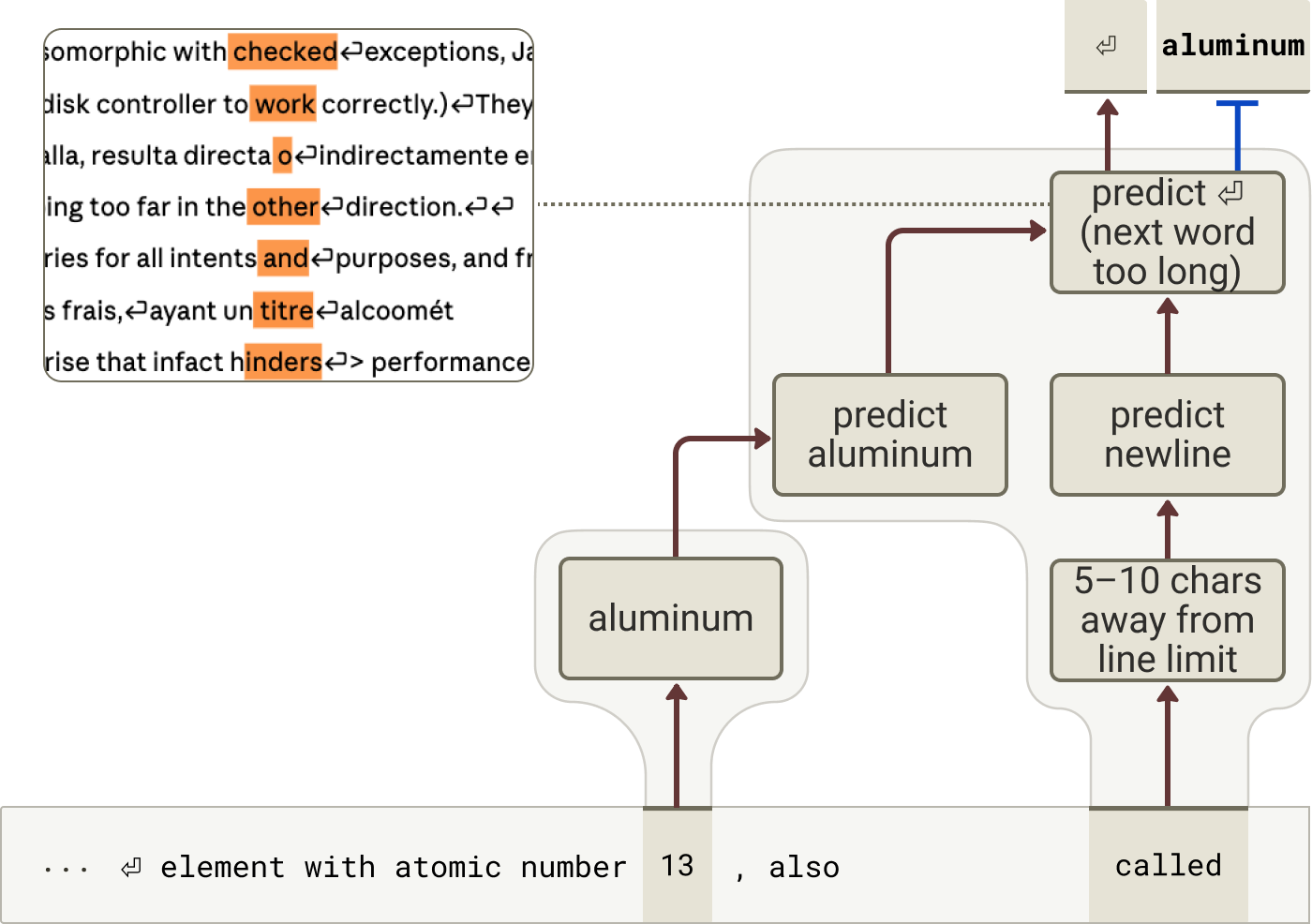}
    \label{fig:gdoc_21}
    \caption{Attribution graph representation of the final linebreaking decision.}
\end{figure}

While the boundary detector activates regardless of the next token length, break\textit{ }predictor\textit{ }features activate only if the next token will exceed the length of the current line (as in the Aluminum prompt), and hence upweight the prediction of a newline.\footnote{These features also sometimes activate on zero-width \textit{modifier} tokens (e.g., a token which indicates the first letter of the following token should be capitalized) that need to be adjacent to the modified token, and the modified token is sufficiently long to go over the line limit (e.g. for “Aluminum” instead of “aluminum”).} We also see break \textit{suppressor} features, which only activate if the next token would just barely fit on the line, and hence downweight the prediction of a newline. Both break predictors and suppressors come in larger feature families, which we display in the \hyperref[sec:appendix-break-predictors]{Appendix}.

\begin{figure}[!htp]
    \centering
    \includegraphics[width=0.9\textwidth,height=0.7\textheight,keepaspectratio]{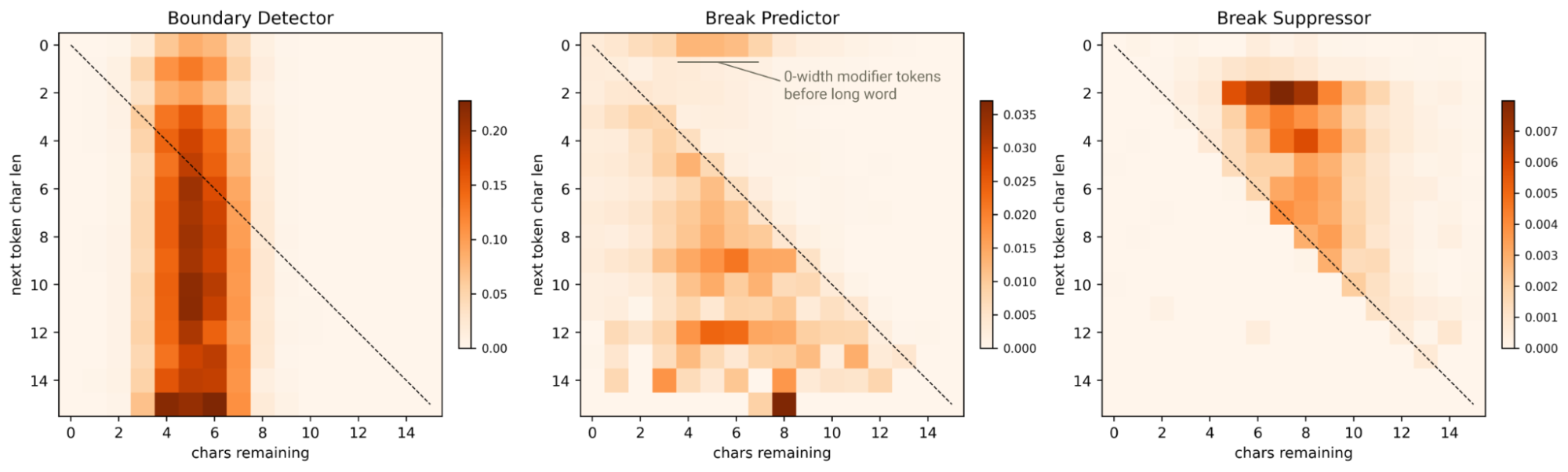}
    \label{fig:gdoc_22}
    \caption{Average activations of three features based on true next token character length and the characters remaining in a line (line width - character count).}
\end{figure}

\subsection{Joint Geometry Enables Easy Computation}

What is the geometry underlying the model’s ability to determine if the next token will fit on the line? Put another way, how is the break predictor feature above constructed from the boundary detector and next-word features?

To study this, we compute the average activations at the end of the model ($\sim$90\% depth) across all tokens for all values of characters remaining $i$ and next token lengths $j$.\footnote{We use the true next non-newline token as the label. This is an approximation because it assumes that the model perfectly predicts the next token. } By performing a PCA on the combination of mean vectors, we see that the two counts are arranged in orthogonal subspaces with only moderate curvature. Note, this lower dimensional geometry may suffice here because the dynamic range of the count is much smaller.

\begin{figure}[!htp]
    \centering
    \includegraphics[width=0.9\textwidth,height=0.7\textheight,keepaspectratio]{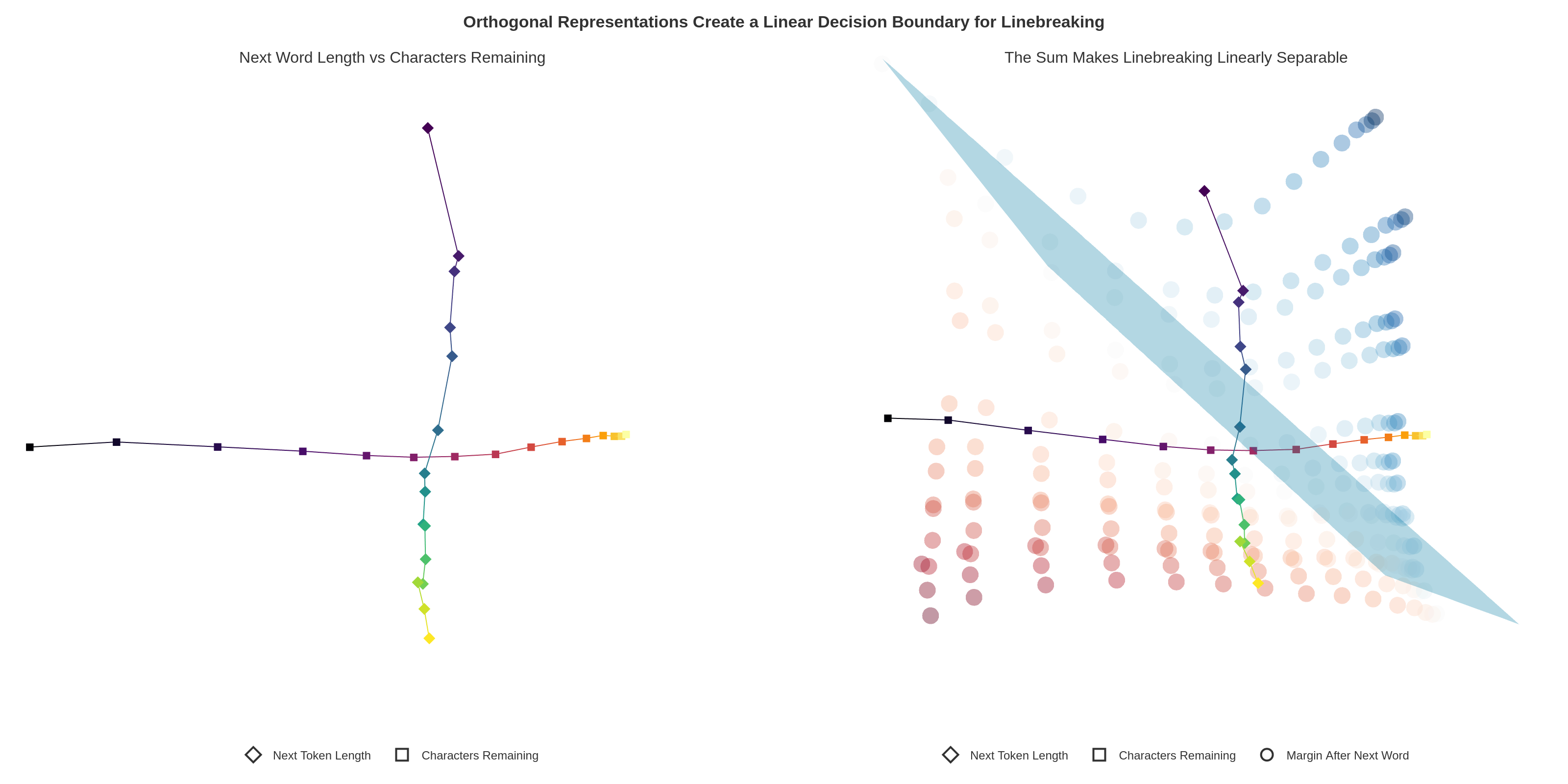}
    \label{fig:linear_classifier}
    \caption{Low dimensional projections of next token character length and characters remaining counting manifolds for 1 (dark) to 15 (light) characters. (Left) The PCA of their union. (Right) The PCA of all their pairwise combinations. The orthogonal representations make the correct newline decision linearly separable.}
\end{figure}

Now consider the pairwise sum of each possible character-remaining vector $i$ and next-token-length vector $j$.\footnote{This sum is principled because both sets of vectors are marginalized data means, so collectively have the mean of the data, which we center to be 0.} Since these counts are arranged orthogonally, the decision to break the line $i-j \geq 0$ corresponds to a simple separating hyperplane. In other words, the prediction to break the line is made trivial by the underlying geometry!

When we use the separating hyperplane from the PCA of these average embeddings on real data, we achieve an AUC of 0.91 on the ground truth of whether the next token should be a newline. This reflects both the error of the three dimensional classifier \textit{and} the error from Haiku’s estimates of the next token.

If the length of the most likely next word is linearly represented, this scheme would allow the model to predict newlines when that word is longer than the length remaining in the line. One could imagine a more general mechanism where the model comprehensively redirects the probability mass from all words that exceed the line limit to the newline. Claude 3.5 Haiku does not seem to leverage such a mechanism: when we compare the predicted distribution of tokens at the end of a line to the distribution on an identical prompt with the newlines stripped, we find them to be quite different.

\section{A Distributed Character Counting Algorithm}\label{sec:count-algo}

Having described how the various character counting representations are \textit{used}, the last big remaining question is: how are they \textit{computed}?

We will show how Haiku uses many attention heads across multiple layers to cooperatively compute an increasingly accurate estimate of the character count. This turned out to be the most complicated mechanism we studied, though there are many similarities with the boundary detection mechanism.

\begin{figure}[!htp]
    \centering
    \includegraphics[width=0.95\textwidth,height=0.7\textheight,keepaspectratio]{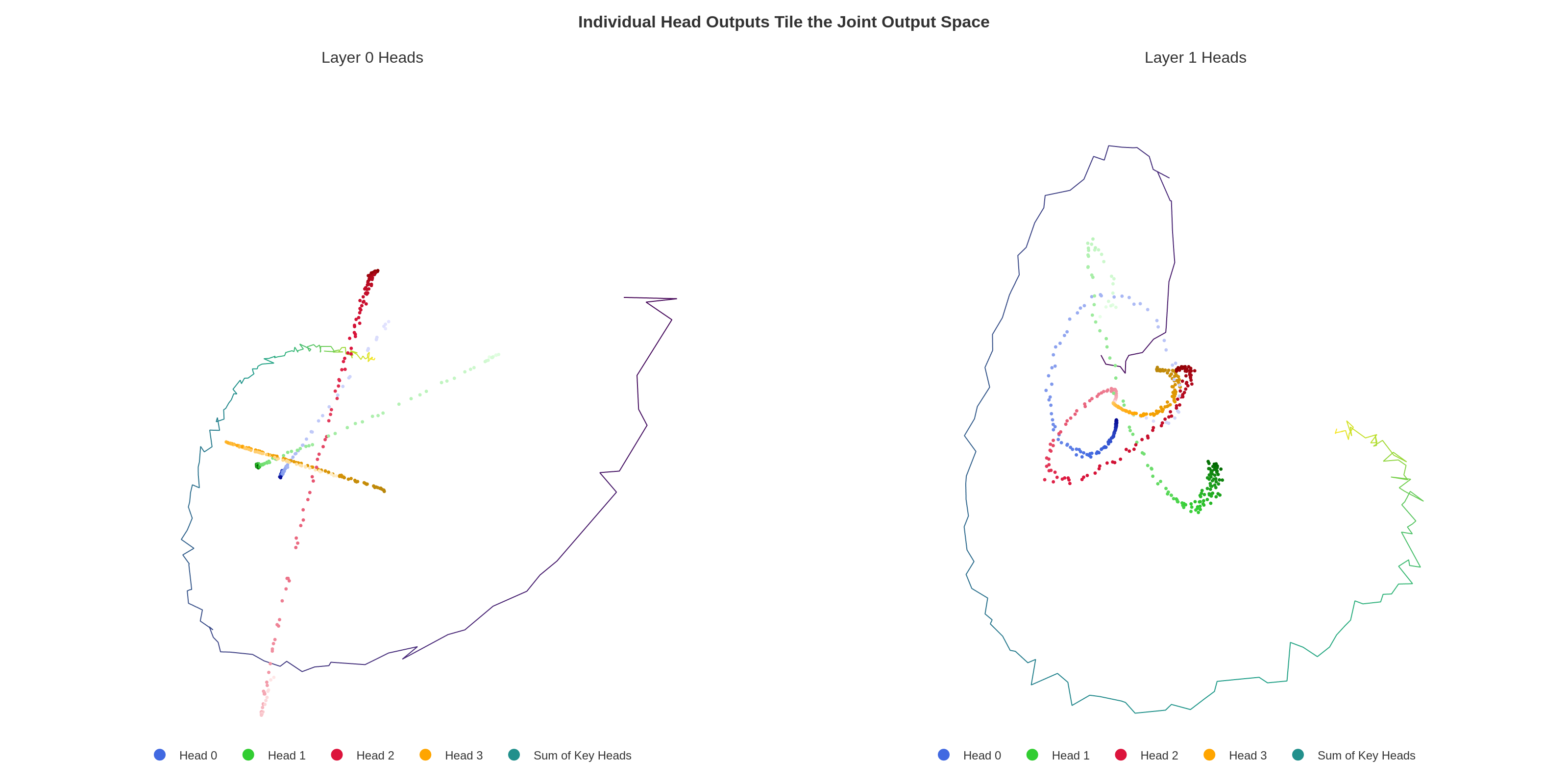}
    \label{fig:head_tiling_per_layer}
    \caption{Comparison of Layer 0 (left) vs Layer 1 (right) average attention outputs in the PCA basis of the character count probes from 1 (dark) to 150 (light) characters. In each layer, the outputs from each head tile the space. In Layer 0, each head output is almost 1-dimensional, while in Layer 1 heads display more curvature (which they got from Layer 0!).}
\end{figure}

To get an intuitive understanding of the behavior of the heads important for counting, we project their outputs into the PCA space of the line character count probes.\footnote{We display the average outputs over many prompts.} Layer 0 heads each write along what appears as a ray when visualized in the first 3 principal components---it is their sum that generates a curved manifold. Layer 1 heads instead output curves which combine to produce an increasingly complex manifold. They appear responsible for sharpening the Layer 0 representation and thus the estimate of the count. We find that the $R^2$ for the character count prediction\footnote{The prediction is the argmax of the head outputs projected on the character count probes.} of the 5 key Layer 0 heads is 0.93, compared to 0.97 using 11 heads in the first two layers.

\subsection{Embedding Geometry}

To understand how the character count is computed, we start at the very beginning: the embedding matrix.

As before, we can train probes or compute the average weights for every distinct token length in the embedding. We visualize the \textit{token} character count probes for character length 1--14 and visualize their top principal components. Using the first 3 principal components, which capture 70\% of the variance, we see that embedding character counts are arranged in a circular pattern (PC1 vs PC2) with an oscillating component (PC3). This pattern is consistent with the ones observed in \hyperref[sec:rippled-representations]{Rippled Representations are Optimal}.

\begin{figure}[!htp]
    \centering
    \includegraphics[width=0.6\textwidth,height=0.7\textheight,keepaspectratio]{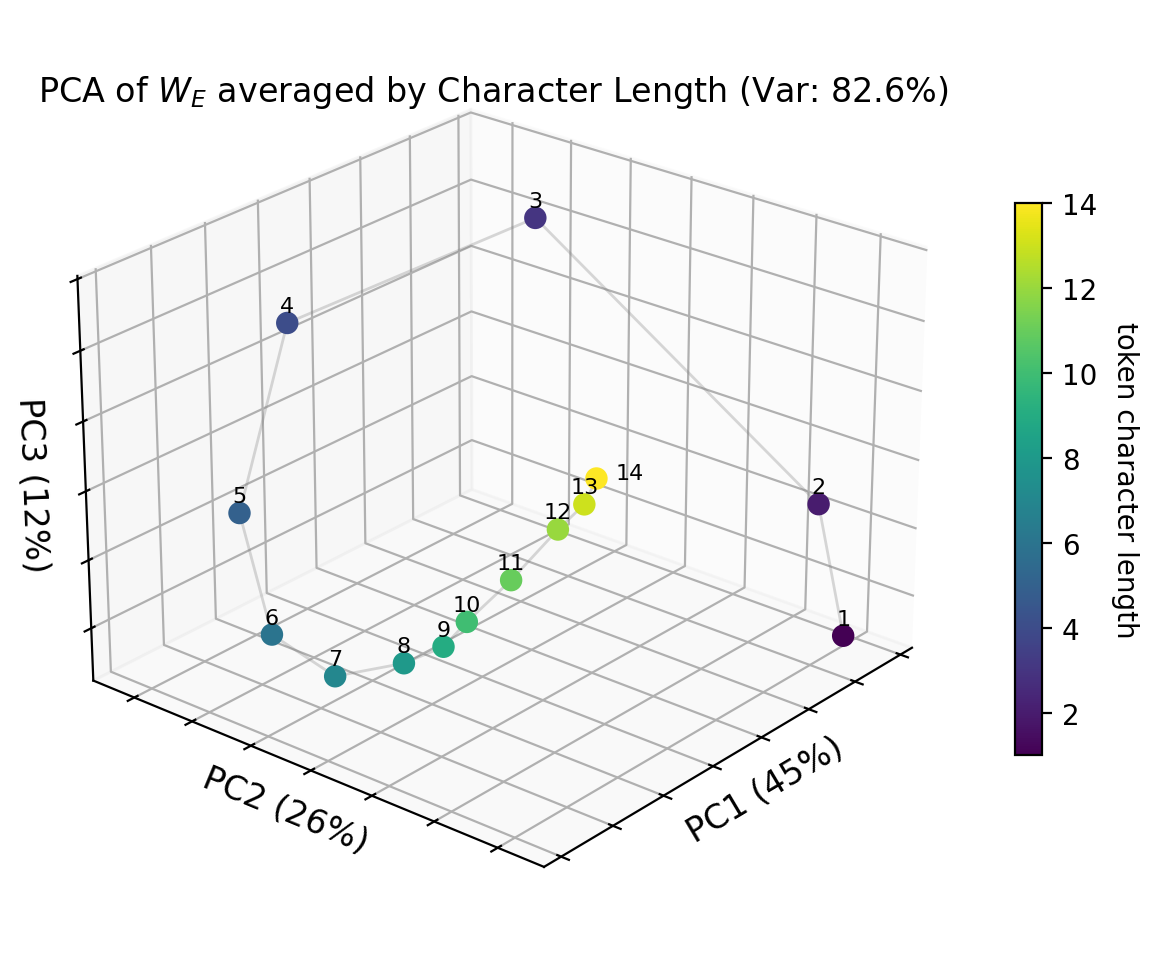}
    \label{fig:gdoc_25}
    \caption{PCA of embedding vectors in $W_E$ averaged by token character length.}
\end{figure}

As with all of the counting manifolds, we also find \hyperref[sec:appendix-token-lengths]{features} that discretize this space into overlapping notions of short, medium, and long words.

\subsection{Attention Head Outputs Sum To Produce the Count}

To understand the counting mechanism, we will work backwards from the summed attention outputs to the embedding. Notably, we:

\begin{itemize}
    \item \textbf{Ignore MLPs} -- The attention head outputs affect the character count representation 4$\times$ more than the MLPs, so we restrict our focus to attention;
    \item \textbf{Focus on }\textbf{First Two Layers} -- Even after layer 0, counting probes have reasonable accuracy and there are coarse positional features. Therefore, we focus on how attention transforms the embeddings into the count and how layer 1 further refines this representation.
\end{itemize}

\begin{figure}[!htp]
    \centering
    \includegraphics[width=0.9\textwidth,height=0.7\textheight,keepaspectratio]{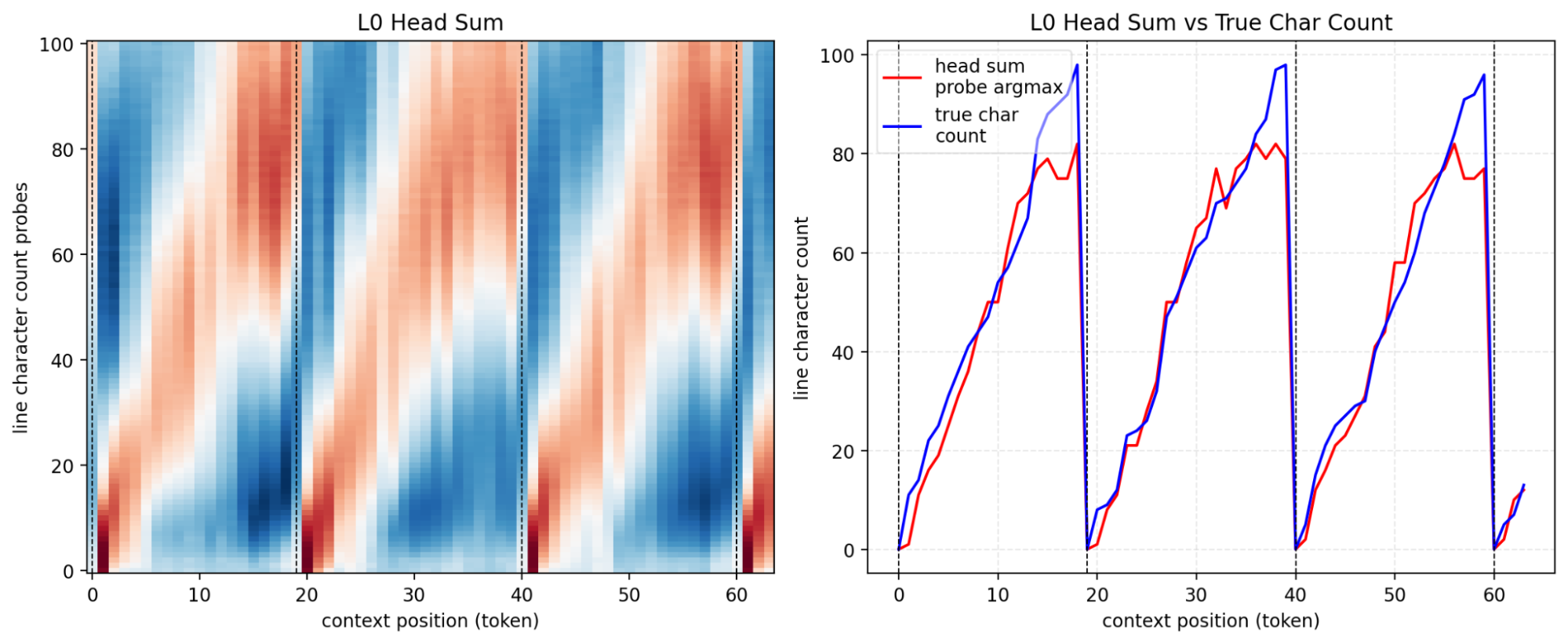}
    \label{fig:gdoc_26}
    \caption{The summed output of 5 important Layer 0 heads on one prompt by token. (Left) The inner product of the summed attention outputs and the character counting probes; (Right) How the argmax of this product compares to the true line count. Context position starts at the first newline, with newlines denoted with dashes.}
\end{figure}

We can decompose the sum above into the contribution from the output of each individual head in layer 0.\footnote{We omit a previous token head for visual presentation.} Under this lens, we see each head performing a relatively low rank computation akin to a classification.

\begin{figure}[!htp]
    \centering
    \includegraphics[width=\textwidth,height=0.7\textheight,keepaspectratio]{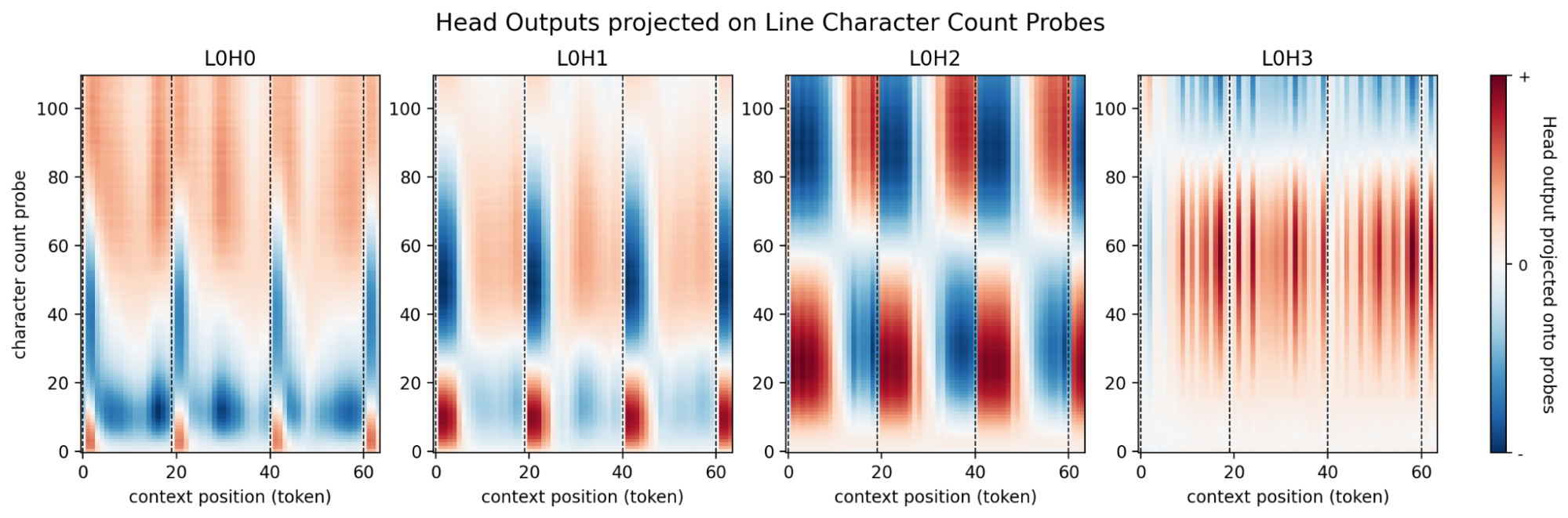}
    \label{fig:gdoc_27}
    \caption{The individual outputs of 4 important Layer 0 heads on one prompt projected onto the character count probes.}
\end{figure}

How do individual heads implement this behavior? We can break down the behavior of an individual head by analyzing its QK circuit (where it attends) and OV circuit (the linear transformation from the embeddings to the output) \cite{nelhage2021mathematical}.

\begin{figure}[!htp]
    \centering
    \includegraphics[width=0.6\textwidth,height=0.7\textheight,keepaspectratio]{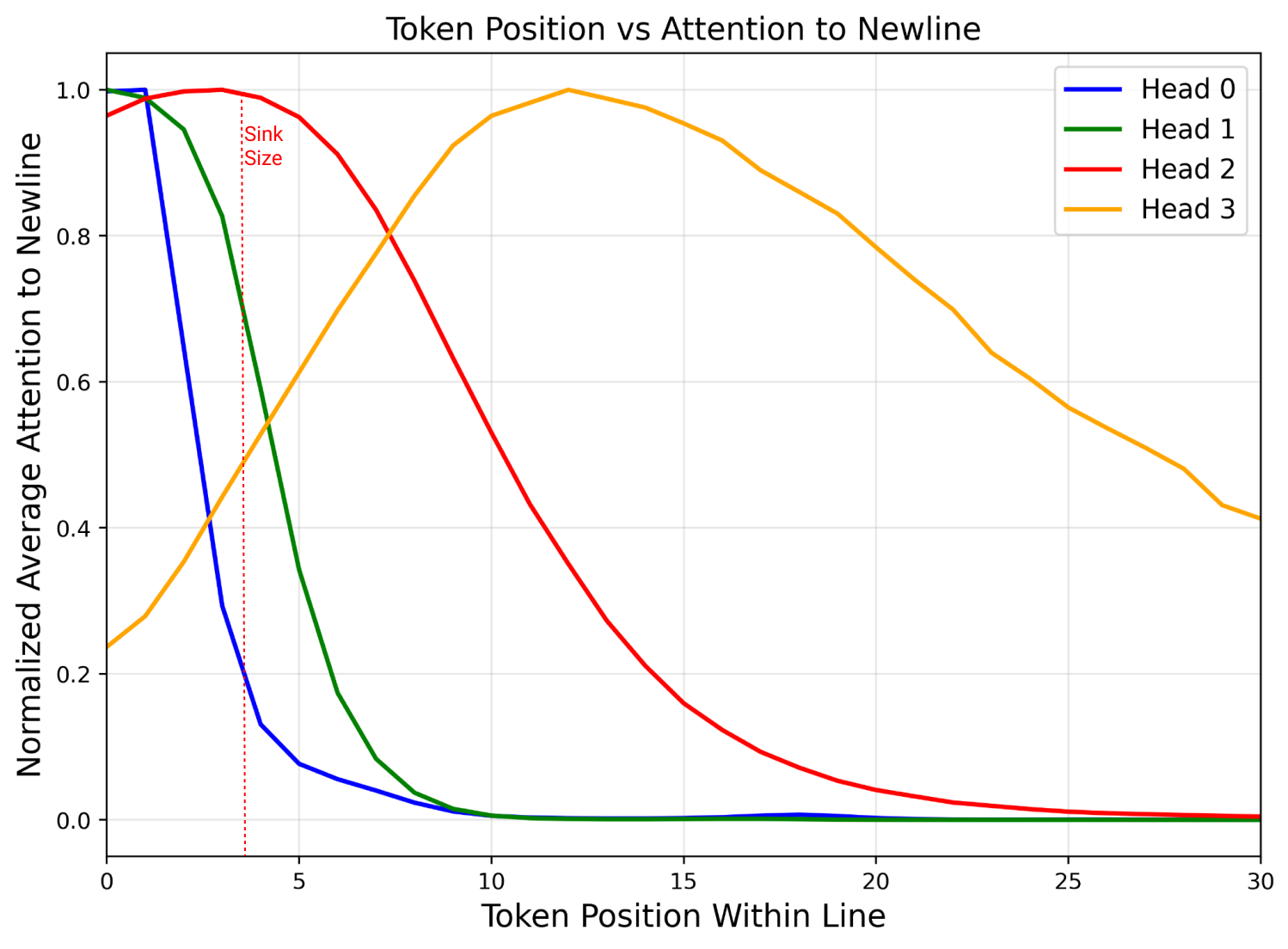}
    \label{fig:gdoc_28}
    \caption{The average attention to the previous newline as a function of the token index in the line. Like boundary heads, these counting heads specialize with different positional offsets.}
\end{figure}
\textbf{QK Circuit}.  Each head $h$ uses the previous newline as an “attention sink,” such that for some number of tokens after the newline ($s_h$), the head just attends to the newline. After $s_h$ tokens, the head begins to smear its attention over its receptive field, which goes up to a maximum of $r_h$ tokens.

\textbf{OV Circuit}. The OV circuit coordinates with the QK circuit to create a heuristic estimate based on the number of tokens in the line multiplied by the average token length ($\mu_c \approx 4$), with an additional length correction term. When attending to the newline, each head upweights the average token length multiplied by the head’s sink size: $s_h\times\mu_c$ characters. If no attention is paid to the newline, then from the perspective of the head, the current token must be at least $s_h+r_h$ tokens into the line and should upweight $(s_h+r_h) \times \mu_c$ character outputs. Finally, the OV circuit applies an additional correction depending on whether the tokens in the receptive field are above or below average in length.

Below, we include a detailed walkthrough of L0H1.

\begin{figure}[!htp]
    \centering
    \includegraphics[width=0.95\textwidth,height=0.7\textheight,keepaspectratio]{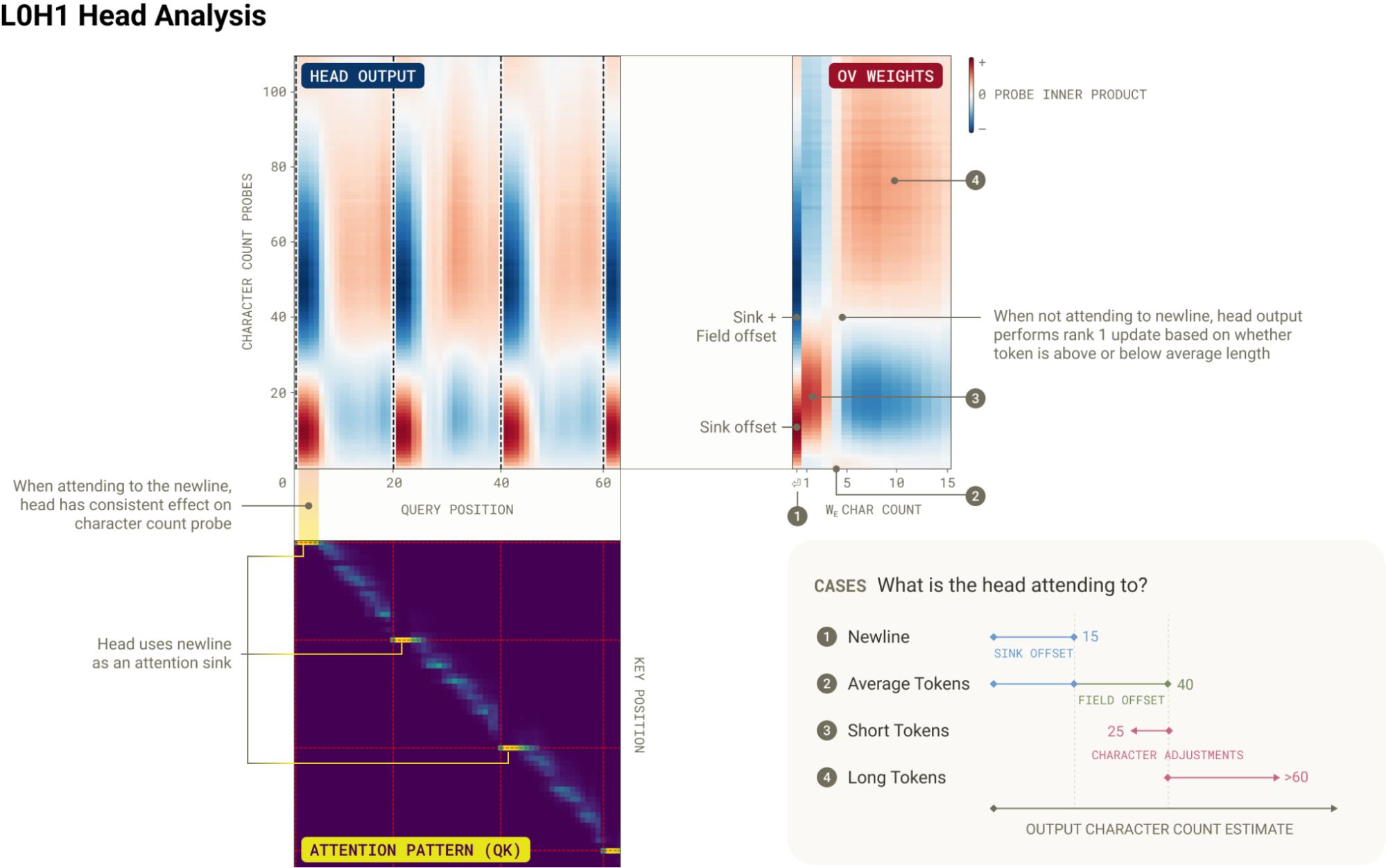}
    \label{fig:gdoc_29}
    \caption{The QK and OV circuit of counting head L0H1. Top left: the head output projected onto the character counting probes for 64 tokens of a single prompt (truncated to the first newline). Bottom left: the attention pattern (transposed of the canonical ordering). Top right: the average embedding vectors projected onto the character count probes via the OV matrix. Bottom right: a summary of the overall computation.}
\end{figure}

For a more detailed analysis of each head, see \hyperref[sec:appendix-mechanics]{The Mechanics of Head Specialization}. Layer 1 attention heads perform a similar operation, but additionally leverage the initial estimate of the character count (see \hyperref[sec:appendix-l1-attn]{Layer 1 Head OVs}).

\subsection{Computing the Line Width}

To compute the line width, the model seems to use a similar distributed counting algorithm to count the characters between adjacent newlines. However, one subtlety that we do not address in this work is how the line width is actually aggregated. It is possible that the model computes a global line width by taking the max over all line lengths in the document or uses an exponentially weighted moving average of the last several line lengths. We do note that the line width uses a partially disjoint set of heads, likely because the “attend to previous newline as a sink” mechanism needs modification when the \textit{current} token is also a newline.

\section{Visual Illusions}\label{sec:illusion}

Humans are susceptible to “visual illusions” in which contextual cues can modulate perception in seemingly unexpected ways. Famous examples include the Müller-Lyer illusion, in which arrows placed on the ends of a line can alter the perceived length of the line \cite{howe2005muller}; the Ponzo and Sander illusions which also modulate perceived line length \cite{yildiz2022review}; and others \cite{schwartz2007space}.

\begin{figure}[!htp]
    \centering
    \includegraphics[width=0.9\textwidth,height=0.7\textheight,keepaspectratio]{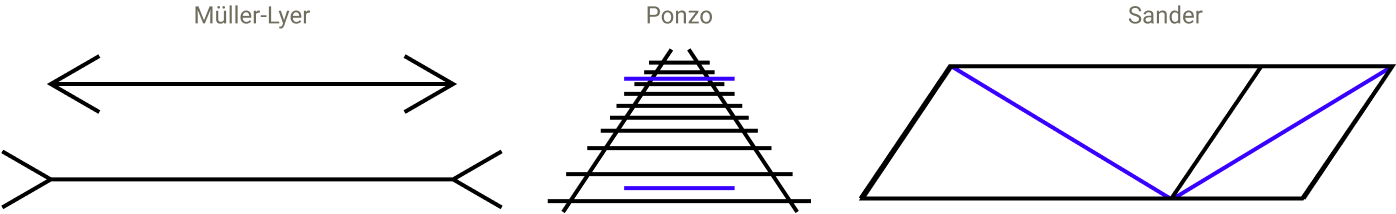}
    \label{fig:gdoc_30}
    \caption{Classic visual illusions in which perception of line-length is modulated.}
\end{figure}

Can we use our understanding of the character counting mechanism to construct a “visual illusion” for language models?

To get started, we took the important attention heads for character counting and investigated what other roles they perform on a wider data distribution. We identified instances in which heads that normally attend from a newline to the previous newline would instead attend from a newline to the two-character string \promptinline{@@}. This string occurs as a delimiter in git diffs, a circumstance in which you might want to start your line count at a location other than the newline:

\begin{promptblock}
[Enter]@@-14,30 +31,24 @@ export interface ClaudeCodeIAppTheme \{[Enter]
\end{promptblock}

But what happens when this sequence appears outside of a git diff context---for instance, if we insert \promptinline{@@} in the aluminum prompt without changing the line length?

\begin{figure}[!htp]
    \centering
    \includegraphics[width=0.9\textwidth,height=0.7\textheight,keepaspectratio]{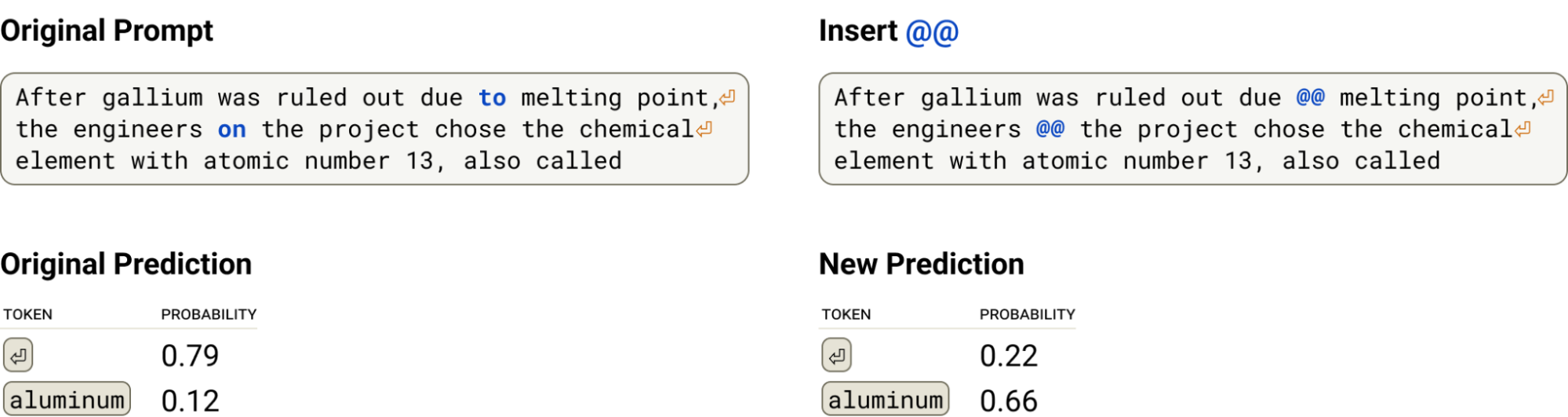}
    \label{fig:gdoc_31}
    \caption{Prediction comparison for original aluminum prompt (left) with visual illusion insert (right).}
\end{figure}

We find that it does modulate the predicted next token, disrupting the newline prediction! As predicted, the relevant heads get distracted: whereas with the original prompt, the heads attend from newline to newline, in the altered prompt, the heads also attend to the \promptinline{@@}.

\begin{figure}[!htp]
    \centering
    \includegraphics[width=0.9\textwidth,height=0.7\textheight,keepaspectratio]{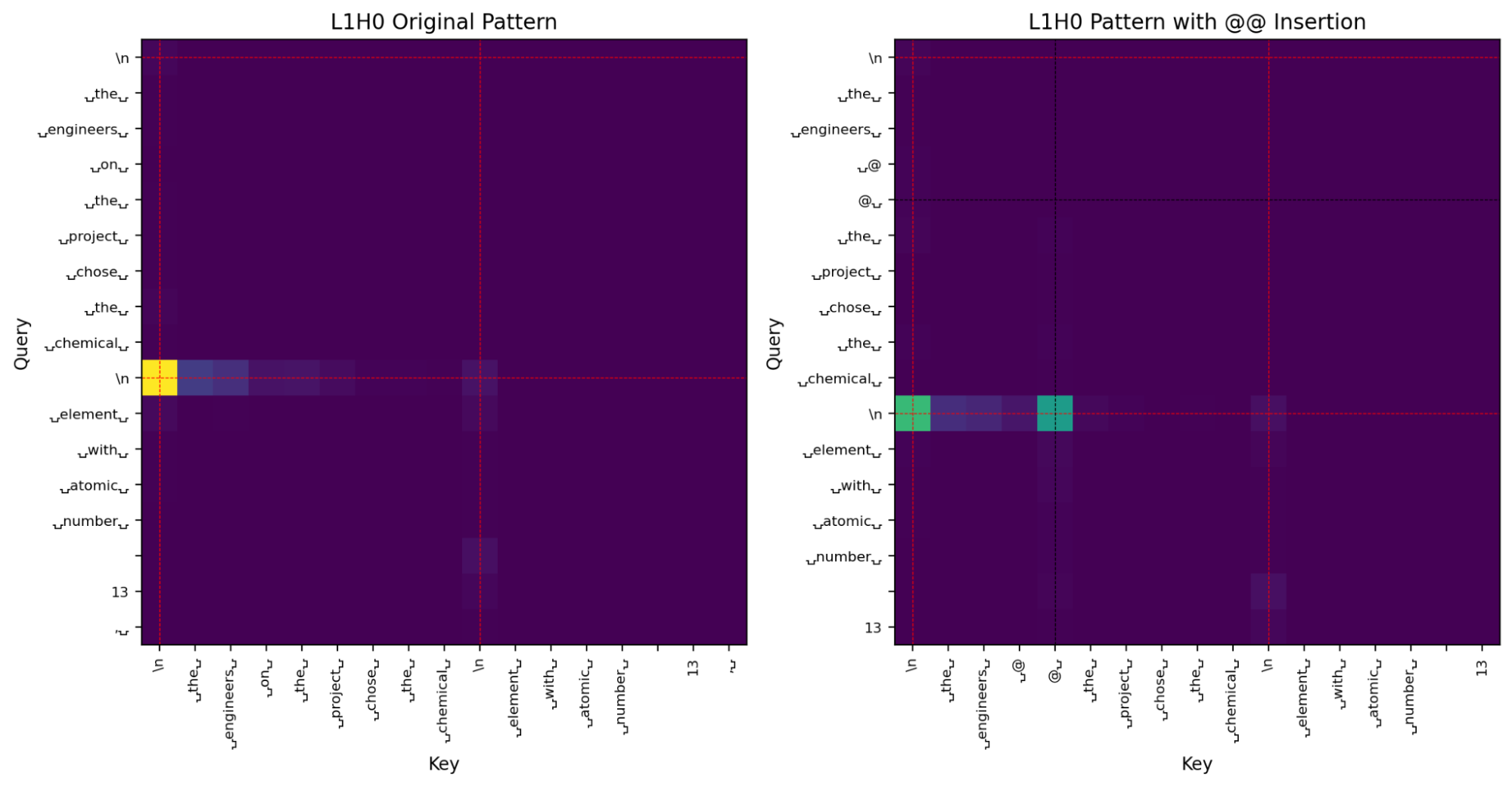}
    \label{fig:gdoc_32}
    \caption{Insertion of \promptinline{@@} ‘distracts’ an attention head which normally attends from \textbackslash{}n back to the previous \textbackslash{}n. (left) Original attention pattern (truncated). (right) Attention pattern (truncated) with \promptinline{@@} insertion. Now it also attends back to the \promptinline{@@}.}
\end{figure}

How specific is this result: does any pair of letters nonsensically inserted into the prompt fully disrupt the newline prediction? We analyzed the impact of inserting (at the same two positions) 180 different two-character sequences, half of which were a repeated character. We found that while most inserted sequences moderately impact the probability of predicting a newline, newline usually remains the top prediction. There was also no clear difference between sequences consisting of the same or different characters. However, a few sequences substantially disrupted newline prediction, most of which appeared to be related to code or delimiters: \promptinline{''}  \promptinline{>>}  \promptinline{\}\}}  \promptinline{;|}  \promptinline{||}  \promptinline{',}  \promptinline{@@}.

We further analyzed the extent to which there was a relationship between ‘distraction’ of the important attention heads and the impact on the newline prediction. Indeed we found that many of the sequences with potent modulation of newline probability---and especially code-related character pairs---also exhibited substantial modulation of attention patterns.

\begin{figure}[!htp]
    \centering
    \includegraphics[width=0.6\textwidth,height=0.7\textheight,keepaspectratio]{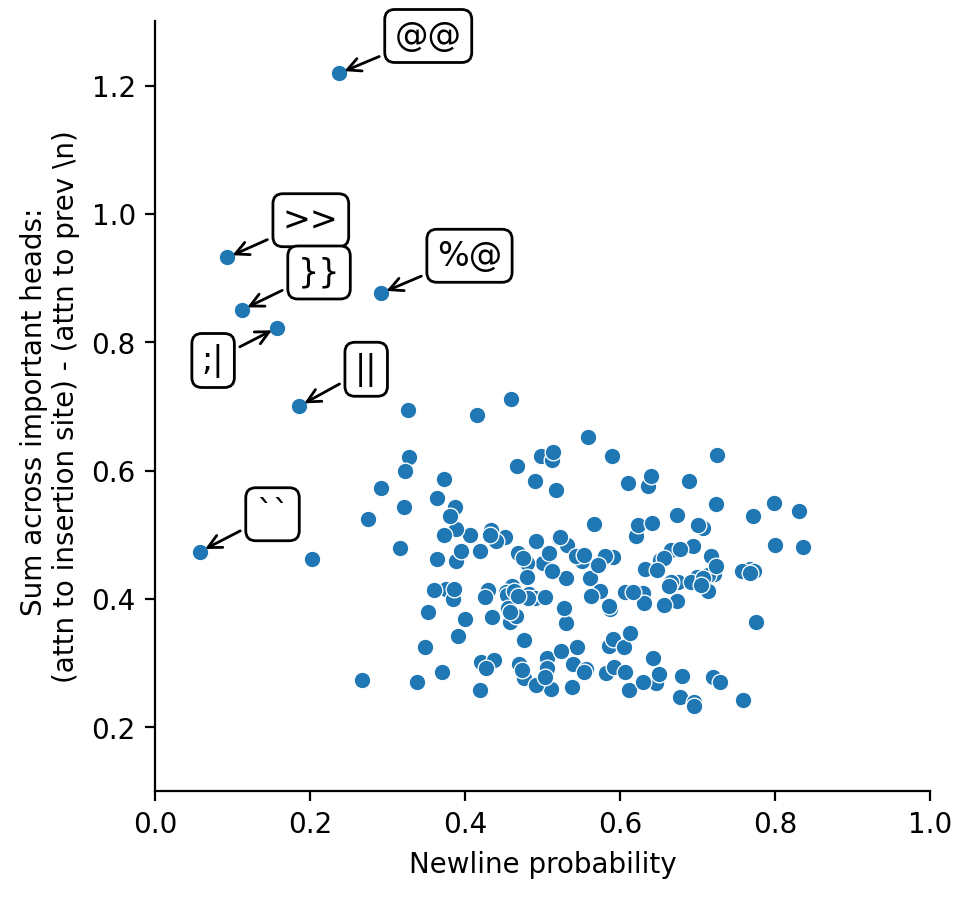}
    \label{fig:gdoc_33}
    \caption{Insertion of most pairs of characters only moderately impacts the probability of predicting a newline. A subset of pairs, most of which appear related to code or delimiters, substantially disrupt newline prediction. The impact on newline prediction (originally 0.79) is correlated with how much inserted tokens ‘distracts’ character counting attention heads.}
\end{figure}

While in the aluminum prompt the task is implicit, this illusion generalizes to settings where the comparison task is made explicit. These direct comparisons are perhaps more analogous to the Ponzo, Sander, and Müller-Lyer illusions, where the perception and comparison is more direct.

\begin{figure}[!htp]
    \centering
    \includegraphics[width=0.9\textwidth,height=0.7\textheight,keepaspectratio]{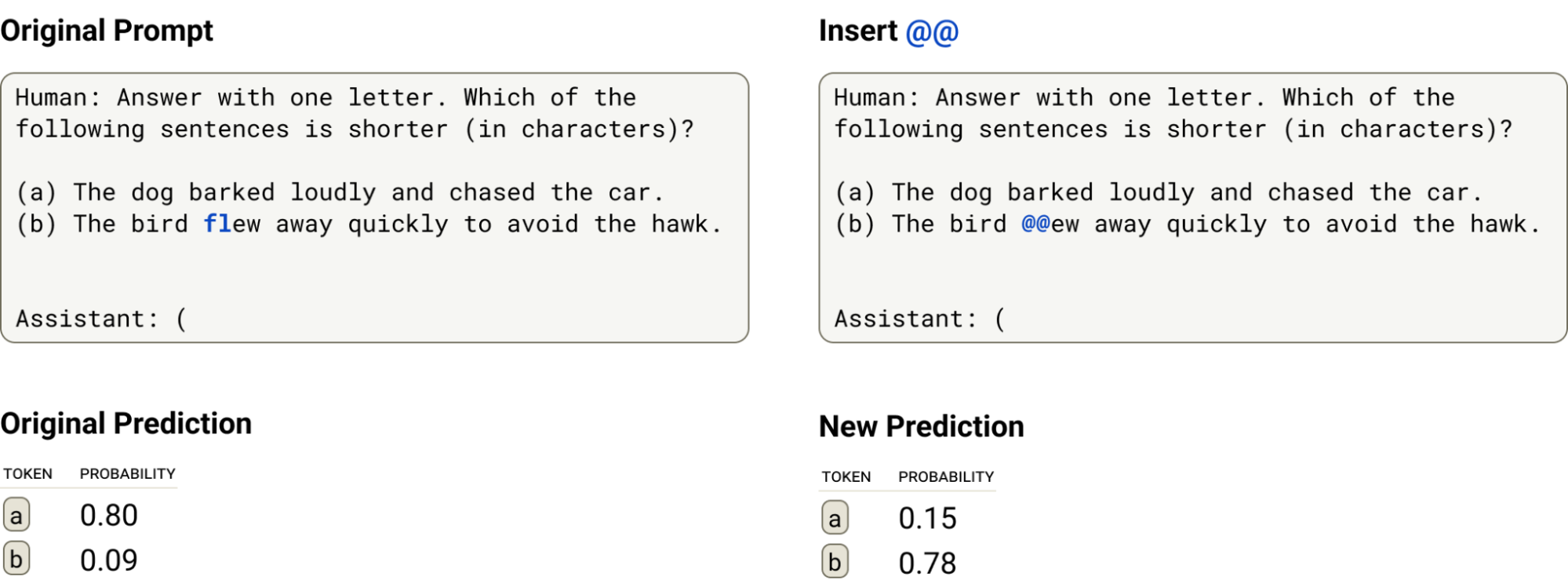}
    \label{fig:gdoc_34}
    \caption{Prediction for explicit length comparison task with normal prompt (left) and with visual illusion insert (right).}
\end{figure}

These effects are robust to multiple choice orderings. Moreover, if the length of the text following the \promptinline{@@} exceeds that of the alternative choice, the alternative choice is selected as being shorter.

While we are not claiming any direct analogy between illusions of human visual perception and this alteration of line character count estimates, the parallels are suggestive. In both cases we can see the broader phenomena of contextual cues, and the application of learned priors about those cues, modulating estimates of object properties of entities. In the human case, priors such as three-dimensional perspective can influence perception of object size, or color constancy can influence estimates of luminance (such as in the checker shadow illusion). Here, one possible interpretation of our results is that mis-application of a learned prior, including the role of cues such as \promptinline{@@} in git diffs, can also modulate estimates of properties such as line length.

\section{Related Work}\label{sec:related-work}

\textbf{Objective}.\textbf{  }This work is at the intersection of LLM “biology” (making empirical observations about what is going on inside models; e.g. \cite{lindsey2025biology,rogers2020primer}) and low level reverse engineering of neural networks  (attempting to fully characterize an algorithm or mechanism; e.g. \cite{olah2020zoom,wang2022interpretability,nanda2023progress,li2025language}). Methodologically, our work makes heavy use of attribution graphs \cite{ameisen2025circuit,ge2024automatically,dunefsky2024transcoders} with QK attributions \cite{kamath2025tracing} built on top of crosscoders \cite{lindsey2024crosscoders}.

\textbf{Linebreaking}. Michaud\textit{ et al. }\cite{michaud2023the} identified linebreaking in fixed-width text as one of the top 400 “quanta” of model behavior in the smallest model (70m parameters) in the Pythia suite.

\textbf{Position. } Prior interpretability work on positional mechanism has largely focused on \textit{token} position (e.g., \cite{yedidia2023gpt2,yedidia2023positional,voita2023neurons,chughtai2024understanding,gurnee2024universal}). These works have shown that there exist MLP neurons \cite{voita2023neurons,gurnee2024universal}, SAE features \cite{chughtai2024understanding}, and learned position embeddings \cite{yedidia2023gpt2} with periodic structure encoding absolute token position. Our work illustrates how a model might also want to construct non-token based position schemes that are more natural for many downstream prediction tasks.

Others have also studied, even going back to LSTMs, the existence of mechanisms in language models for controlling the length of output responses \cite{shi2016neural,moon2025length}, as well as performed more theoretical analyses of the space of counting algorithms \cite{suzgun2019lstm,chang2024language}.

\textbf{Geometry and Feature Manifolds. }  Beyond position, there has been extensive work in understanding the geometric representation of numbers, especially in toy models (e.g., \cite{nanda2023progress,zhong2023clock,morwani2023feature}) and in the context of arithmetic in LLMs (e.g., \cite{stolfo2023mechanistic,zhou2024pre,nikankin2024arithmetic,kantamneni2025trig,hu2025understanding}). Collectively, these works have shown that both real LLMs and toy transformers learn periodic representations \cite{zhou2024pre,kantamneni2025trig,hu2025understanding}, with numbers arranged in a helix to enable certain matrix multiplication based addition algorithms \cite{nanda2023progress,kantamneni2025trig}, and that these representations are provably optimal in certain settings \cite{morwani2023feature}. In our context, we similarly observe helical representations \cite{kantamneni2025trig}, numeric dilation \cite{alquboj2025number}, and distributed algorithms across components that collectively implement a correct computation \cite{hanna2023does,hu2025understanding}.

Multidimensional features with clear geometric structure have been found in more natural contexts \cite{gould2023successor,engels2025not,modell2025origins}, like in the representation and computation of certain ordinal relationships (e.g., months of the year). In vision models, curve detector neurons \cite{cammarata2020curve} and features \cite{gorton2024missing} have been especially well studied and closely resemble the kind of discretization we observe with the families of character counting features. Many other topics have received interpretability analysis of the underlying geometry, such as grammatical relations \cite{hewitt2019structural,reif2019visualizing}, multilingual representations \cite{chang2022geometry}, truth \cite{marks2023geometry}, binding \cite{feng2023language}, refusal \cite{wollschlager2025geometry}, features \cite{hindupur2025projecting,li2025geometry}, and hierarchy \cite{park2024geometry}, though more conceptual research is needed \cite{wattenberg2024relational}.

Perhaps most relevant is recent work from Modell et al. \cite{modell2025origins}, who provide a more formal notion of a feature manifold, and propose that cosine similarity encodes the intrinsic geometry of features. When testing their theory, they observe highly structured and interpretable data manifolds that have ripples and dilation, similar to our counting manifolds. These observations raise a methodological challenge in how to best capture data with different structure (see e.g. \cite{hindupur2025projecting,michaud2025understanding,huang2025decomposing}), but also the exciting hypothesis that many naturally continuous variables (e.g., \cite{heinzerling2024monotonic,gurnee2024language}) exist in more organized manifolds.

\textbf{Biological Analogues}\textbf{.  }The geometric and algorithmic patterns we observe have suggestive parallels to perception in biological neural systems. Our character count features are analogous to place cells on a 1-D track \cite{Moser2008PlaceCG} and our boundary detecting features are analogous to boundary cells \cite{solstad2008representation}. These features exhibit dilation---representing increasingly large character counts activating over increasingly large ranges---mirroring the dilation of number representations in biological brains \cite{dehaene2003neural,piazza2004tuning}. Moreover, the organization of the features on a low dimensional manifold is an instance of a common motif in biological cognition (e.g., \cite{perich2025neural}). While the analogies are not perfect, we suspect that there is still fruitful conceptual overlap from increased collaboration between neuroscience and interpretability \cite{vilas2024position,he2024multilevel,leshinskaya2025cognitively}.

\section{Discussion}\label{sec:discussion}

In this paper, we studied the steps involved in a large model performing a naturalistic behavior. The linebreaking task, frequently encountered in training, requires the model to represent and compute a number of scalar quantities involving position in character count units that are not explicit in its input or output\footnote{Tokens do not come annotated with character counts, and there are no vertical bars on the page showing the line width.}, then integrate those values with the outputs of complex semantic circuits (that predict the next proper word) to predict the next token. We found sparse features corresponding to each important step of the computation, and for those steps involving scalar quantities, we were able to find a geometric description that significantly simplified the interpretation of the algorithm used by the model. We now reflect on what we learned from that process:

\textbf{Naturalistic Behavior and Sensory Processing}. Deep mechanistic case studies benefit from choosing behaviors that the model performs consistently well, as these are more likely to have crisper mechanisms. This means prioritizing tasks that are natural in pretraining over tasks that seem natural to human investigators, and ideally, that are easily supervisable. As in biological neuroscience, perceptual tasks are often both natural and easy to supervise for interpretability (e.g., it is easy to modify the input in a programmatic way). Although we sometimes describe the early layers of language models as responsible for “detokenizing” the input \cite{elhage2022solu,gurnee2023finding,ferrando2024information,lad2024remarkable}, it is perhaps more evocative to think of this as perception. The beginning of the model is really responsible for \textit{seeing} the input, and much of the early circuitry is in service of sensing or perceiving the text similar to how early layers in vision models implement low level perception \cite{olah2020zoom,lepori2024beyond}.

\textbf{The Utility of Geometry.  }Many of the representations and computations we studied had elegant geometric interpretations. For example, the counting manifolds are the result of an optimal tradeoff between capacity and resolution, with deep connections to space-filling curves and Fourier features. The boundary head twist was especially beautiful, and after discovering one such head, we were able to correctly predict that there would need to be additional heads to provide curvature in the output. The distributed character counting algorithm was more complex, but we were still able to clarify our view by studying linear actions on these manifolds. For other computations, like the final breaking decision, the linear separation was clearly a part of the story but there must be some additional complexity we were not able to see yet to handle multitoken outputs. For the more semantic operations, we purely relied on the feature view. Of course, describing any behavior in full is immensely complicated, and there is a long list of possible subtleties we did not study: how the model accounts for uncertainty in its counting, its mechanism for estimating the line width given multiple prior lines of text, how it adapts to documents with variable line width, how it handles multiple plausible output tokens of different lengths or multitoken words, or various special cases (e.g., a \LaTeX{} \textbackslash{}footnote\{\} or a markdown link). For the inspired, we share transcoder attribution graphs for a fixed-width line break prompt on \href{https://www.neuronpedia.org/gemma-2-2b/graph?slug=fourscoreandseve-1757368139332\&pruningThreshold=0.8\&densityThreshold=0.99\&pinnedIds=14_19999_37\&clerps=\%5B\%5B\%2214_200290090_37\%22\%2C\%22nearing+end+of+the+line\%22\%5D\%5D}{Gemma 2 2B} and \href{https://www.neuronpedia.org/qwen3-4b/graph?slug=fourscoreandseve-1757451285996\&pruningThreshold=0.8\&densityThreshold=0.99\&clerps=\%5B\%5B\%2230_117634760_39\%22\%2C\%22nearing+end+of+line\%22\%5D\%5D\&pinnedIds=30_15307_39}{Qwen 3 4B}, using the new neuronpedia interactive interface.

\textbf{Unsupervised Discovery  }It likely would not have been possible to develop this clarity if it were not for the unsupervised sparse features. In fact, when we started this project, we attempted to just probe and patch our way to understanding, but this turned out poorly. Specifically, we did not understand what we were looking for (e.g. we didn’t know to distinguish line width vs. character count), where to look for it (e.g., we didn’t expect line width to only be represented on the newline), or how to look for it (we started by training 1-D linear regression probes). However, after identifying some relevant features but before spending substantial effort systematically characterizing their activity profiles, we were also confused by what they were representing. We saw dozens of features that were vaguely about newlines and linebroken text, but their differences were not obvious from flipping through the activating examples. Only after we tested these features on synthetic datasets did their role in the graph and the underlying computation become clear. We suspect better automatic labels \cite{bills2023language,paulo2024automatically,gur2025enhancing} enhanced with agentic workflows \cite{shaham2024multimodal,bricken2025automating} would accelerate this work, especially in less verifiable domains.

\textbf{Feature-Manifold Duality.  }The discrete feature and geometric feature-manifold perspectives offer dual lenses on the same underlying object. For example, in this work the model's representation of character count can be completely described (modulo reconstruction error) by the activities of the features we identified, where the action of the boundary heads is described by virtual weights that expand out the feature interactions via attention head matrices. The same character count representation can be described by a 1-dimensional feature manifold -- a curve in the residual stream parametrized by the character count variable -- where linear action of the boundary heads is described by continuous “twisting” of the manifold. In general, geometric structures learned by the model will likely admit both global parametrizations and local discrete approximations.

\textbf{The Complexity Tax.} Despite this duality, the descriptions produced by the two perspectives differ in their simplicity. The discrete features shatter the model into many pieces, producing a complex understanding of the computation. This seems like a general lesson. It seems like discrete features and attribution graphs may provide a true description of model computation, which can be found in an automated way using dictionary learning. Getting any true, understandable description of the computation is a very non-trivial victory! However, if we stop there, and don't understand additional structure which is present, we pay a \textit{complexity tax}, where we understand things in a needlessly complicated way. In the line breaking problem, constructing the manifold paid down this tax, but one could imagine other ways of reducing the interpretation burden.

\textbf{A Call for Methodology. }Armed with our feature understanding, we were able to directly search for the relevant geometric structures. This was an existence proof more than a general recipe, and we need methods that can automatically surface simpler structures to pay down the complexity tax. In our setting, this meant studying feature manifolds, and it would be nice to see unsupervised approaches to detecting them. In other cases we will need yet other tools to reduce the interpretation burden, like finding hierarchical representations \cite{costa2025flat} or macroscopic structure \cite{olah2023interpretability} in the global weights \cite{ameisen2025circuit}.

\textbf{A Call for Biology.  }The model must perform other elegant computations. We can find these by starting with a specific task the model performs well, study this from multiple perspectives, develop methodology to answer the remaining questions, and relentlessly attempt to simplify our explanations. Because the investigation is grounded in specific examples of a behavior, it provides a fast feedback loop, can shed light on weaknesses of existing methods and inspire new ones, and can sharpen our conceptual language for understanding neural networks. We would be excited to see more deep case studies that adopt this approach.

\bibliographystyle{plainnat}
\bibliography{bibliography}

\newpage
\appendix

\section{Citation Information}\label{sec:citation-info}

For attribution in academic contexts, please cite with this BibTeX citation

\begin{lstlisting}
@article{gurnee2025when,
  author={Gurnee, Wes and Ameisen, Emmanuel and Kauvar, Isaac and Tarng ,Julius and Pearce, Adam and Olah, Chris and Batson, Joshua},
  title={When Models Manipulate Manifolds: The Geometry of a Counting Task},
  journal={Transformer Circuits Thread},
  year={2025},
  url={https://transformer-circuits.pub/2025/linebreaks/index.html}
}
\end{lstlisting}

\section{Acknowledgments}\label{sec:acknowledgments}

We would like to thank the following people who reviewed an early version of the manuscript and provided helpful feedback that we used to improve the final version: Owen Lewis, Tom McGrath, Eric Michaud, Alexander Modell, Patrick Rubin-Delanchy, Nicholas Sofroniew, and Martin Wattenberg. We are also thankful to all the members of the interpretability team for their helpful discussion and feedback, especially Doug Finkbeiner for discussions of rippling and ringing, Jack Lindsey on framing, Tom Henighan for feedback on clarity, Brian Chen for improving the design of the figures and line edits of the text, and the team who built the attribution graph \cite{ameisen2025circuit} and QK attribution infrastructure \cite{kamath2025tracing}.

\section{Haiku Task Performance}\label{sec:appendix-task-performance}

Haiku is able to adapt to the line length for every value of $k$, predicting newlines at the correct positions with high probability by the third line. Of course, some error is to be expected even with a perfect estimate of line length, as the model may incorrectly predict the next semantic token. Below is the mean log-prob and accuracy for newline prediction of Haiku on 200 prose sequences that were synthetically wrapped to have lines of character length $k$, for $k = 20, 40, \ldots, 140$.

\begin{figure}[!htp]
    \centering
    \includegraphics[width=0.9\textwidth,height=0.7\textheight,keepaspectratio]{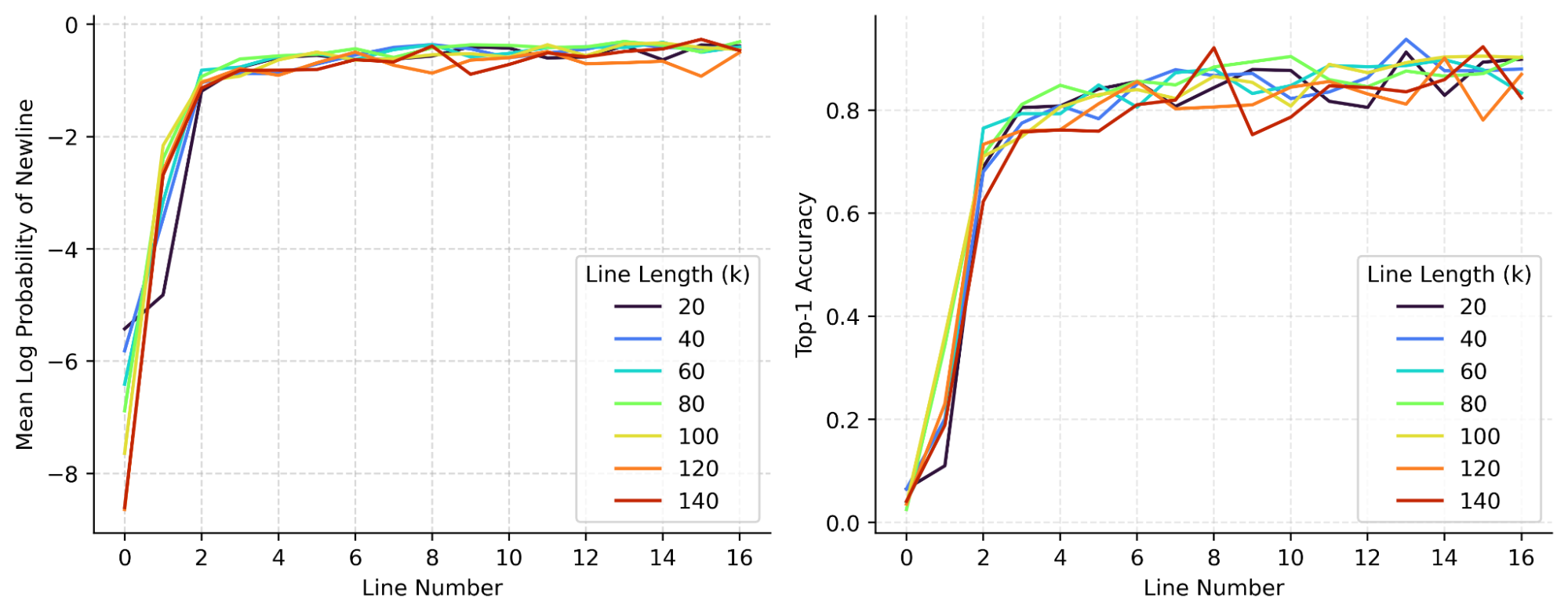}
    \label{fig:gdoc_35}
    \caption{Task performance of Haiku (as measured in mean log prob of newline and top-1 accuracy) on the linebreaking task as a function of number of lines in the context.}
\end{figure}

\section{Feature Splitting and Universality}\label{sec:appendix-feature-splitting}

It is natural to ask if the character counting features are fundamental, or simply one discretization of the space among many. We found that dictionaries of different sizes learn features with very similar receptive fields, so this featurization -- including the slowly dilating widths -- is in some sense \textit{canonical}. We hypothesize that this canonical structure emerges from boundary constraint: positions near zero (start of line) create a natural anchoring point for feature development.

\begin{figure}[!htp]
    \centering
    \includegraphics[width=0.9\textwidth,height=0.7\textheight,keepaspectratio]{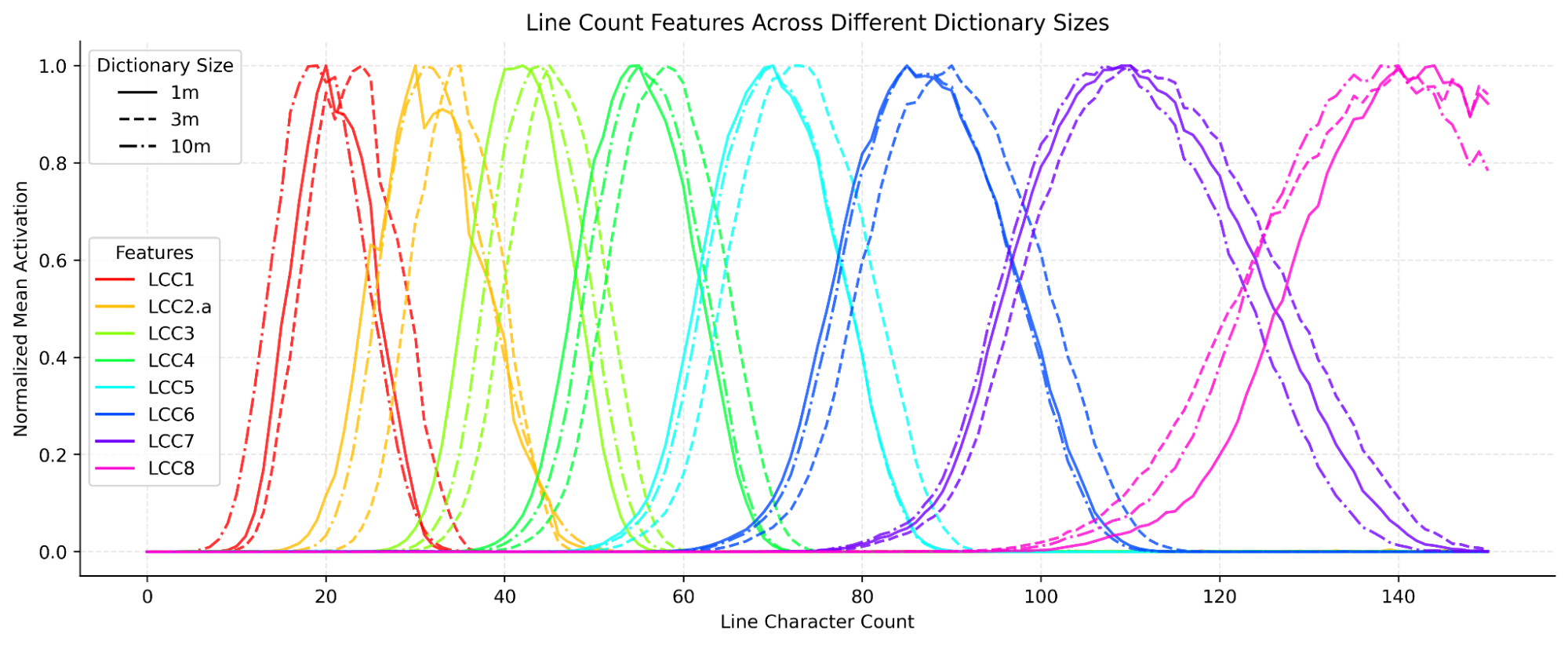}
    \label{fig:gdoc_36}
    \caption{Feature response curves from three differently sized dictionaries, suggesting that this particular sparse parameterization is canonical.}
\end{figure}

The geometry of the decoder directions is also fairly consistent between the dictionaries, showing characteristic ringing.

\begin{figure}[!htp]
    \centering
    \includegraphics[width=0.9\textwidth,height=0.7\textheight,keepaspectratio]{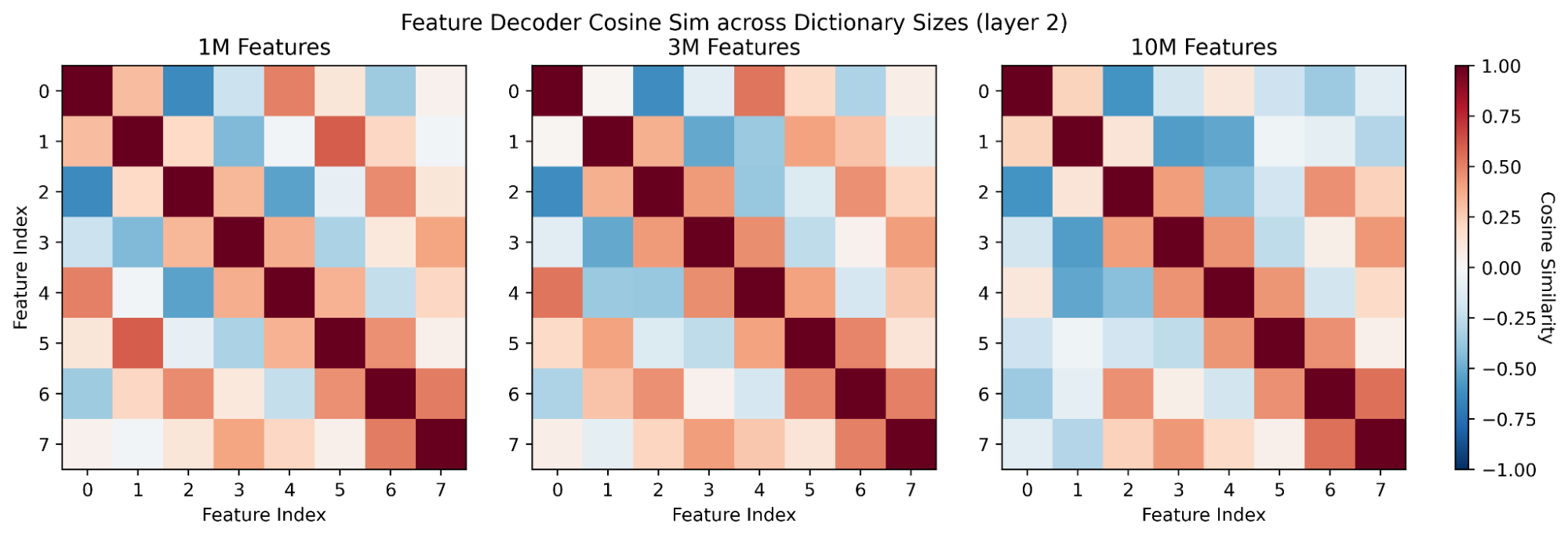}
    \label{fig:gdoc_37}
    \caption{Pairwise cosine similarity of feature decoders across different dictionary sizes.}
\end{figure}

However, we do see some evidence of feature splitting. For example, below are three character count feature which activate on the same interval ($\sim$20--45 characters in the line), but differentially activate for lines of different widths: LCC2.a activates on all line widths, LCC2.b preferentially activates on long line widths, and LCC2.c preferentially activates when close to the line width boundary.

\begin{figure}[!htp]
    \centering
    \includegraphics[width=0.9\textwidth,height=0.7\textheight,keepaspectratio]{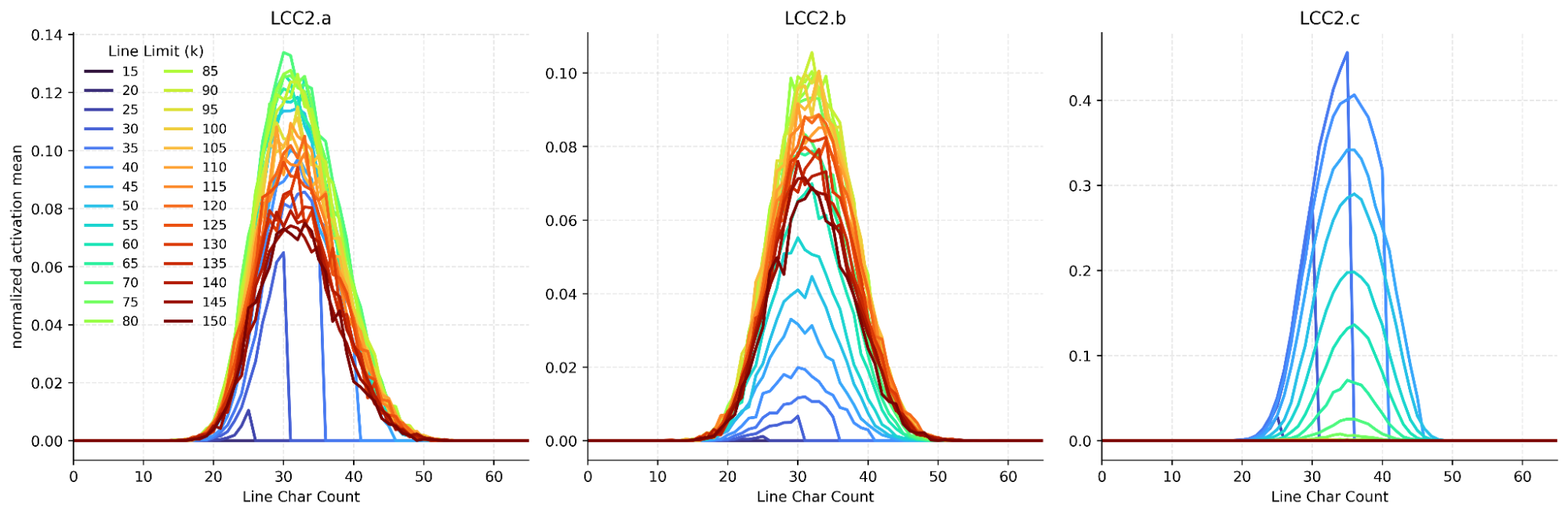}
    \label{fig:gdoc_38}
    \caption{Response curves for three different features that all activate from the to 20-50 character for all possible line widths.}
\end{figure}

Recent work has raised the possibility that feature dictionaries could behave pathologically where there exist feature manifolds \cite{michaud2025understanding}, because a dictionary could allocate an increasing number of features in a finer tiling of the space. However, our observation that cross-coders of varying size tile this feature manifold in a canonical way suggest that this behavior does not occur in this setting.

\section{Line Width Features}\label{sec:appendix-width-features}

Line width features tile the space similar to character count features.

\begin{figure}[!htp]
    \centering
    \includegraphics[width=0.9\textwidth,height=0.7\textheight,keepaspectratio]{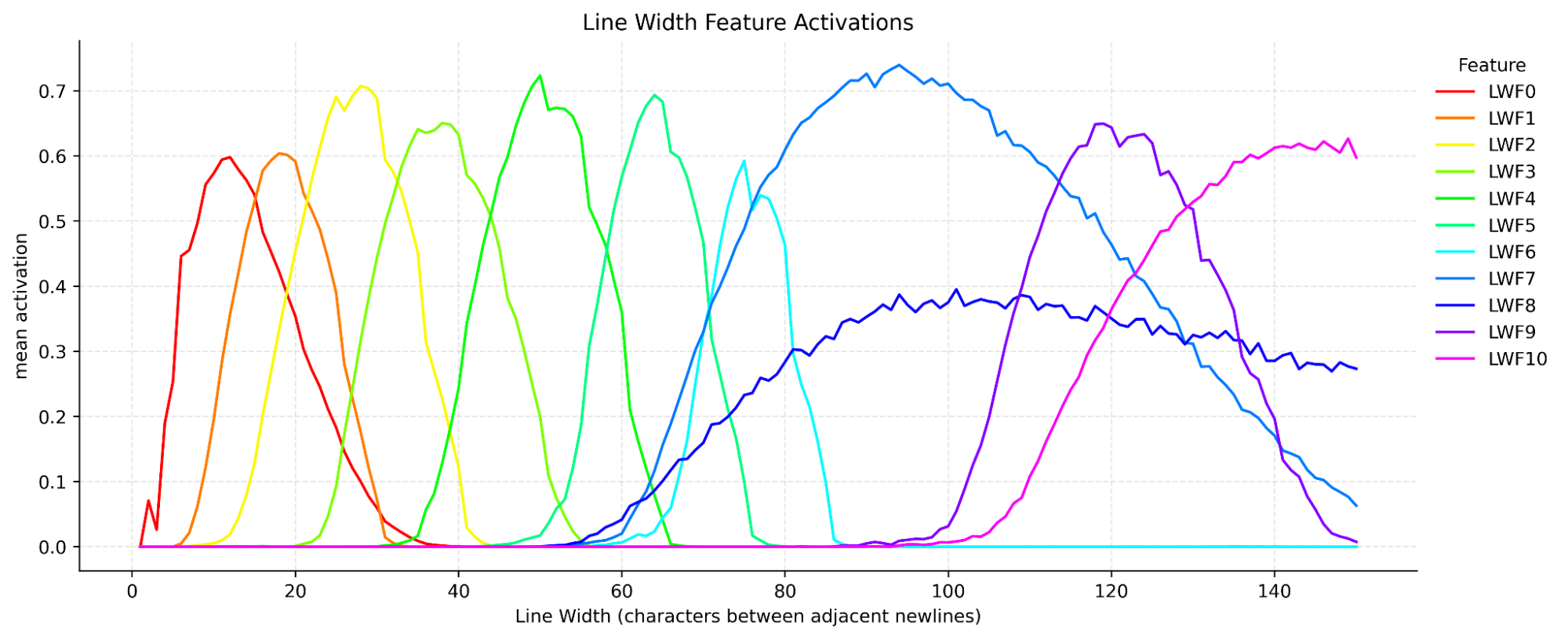}
    \label{fig:gdoc_39}
    \caption{Feature response curves for line width features.}
\end{figure}

\section{Dynamical System Model}\label{sec:appendix-dynamical}

We simulate $N = 100$ points on the unit $(n-1)$-sphere in $\mathbb{R}^n$ ($n \in \{3,\ldots,8\}$) with pairwise forces: $\mathbf{F}_{ij} = \begin{cases} \frac{1 - (d_{ij} - 1)/2}{r_{ij}} \hat{\mathbf{r}}_{ij} & \text{when }d_{ij} \leq w \\ -\frac{\min(5, 1/r_{ij})}{r_{ij}} \hat{\mathbf{r}}_{ij} & \text{when }d_{ij} > w \end{cases}$, where $r_{ij} = \|\mathbf{x}_j - \mathbf{x}_i\|$, $\hat{\mathbf{r}}_{ij} = (\mathbf{x}_j - \mathbf{x}_i)/r_{ij}$, $w$ is the attractive zone width parameter, and $d_{ij} = \min(|j-i|, |j-i+N|, |j-i-N|)$ is the index distance (for the circular topology; for the interval it is just $d_{ij} = |j-i|$). Evolution follows $\dot{\mathbf{v}}_i = \sum_{j \neq i} \mathbf{F}_{ij} - 0.05\mathbf{v}_i$ and $\dot{\mathbf{x}}_i = \mathbf{v}_i$ with sphere constraint $\mathbf{x}_i \leftarrow \mathbf{x}_i/\|\mathbf{x}_i\|$ enforced after each timestep ($\Delta t = 0.01$, damping $\alpha = 0.95$).

\section{Analytic Construction of Ringing and Fourier Modes}\label{sec:appendix-gibbs}

We explore a deeper connection between the ringing observed in the character count feature manifold, and a connection to Fourier analysis in an analytical construction.

Suppose that we wish to have a discretized circle's worth of unit vectors, each similar to its neighbors but orthogonal to those further away. Then the cosine similarity matrix $X$ of these will be the circulant matrix of a narrow-peaked function $f$ (left, below). The columns of the square root of $X$ are $n$ vectors $v_1,\ldots,v_n \in \mathbb{R}^n$, where $n$ is the number of discrete points on the circle, whose inner products reproduce the similarity matrix.\footnote{The entire continuous circle embeds into the infinite-dimensional Hilbert space $L^2\mathbb{S^1}$ via this construction.} Now suppose we would like to find vectors in a lower-dimensional space whose similarity matrix approximates $X$, like the model does for character counts. The best $k$-dimensional approximation of this similarity, in an $L^2$ sense, is given by taking the eigendecomposition of $X$ and truncating it to the top $k$ eigenvectors; the square root of the result will provide $n$ vectors in $k$-dimensions with the corresponding similarity pattern. If $\pi_k$ is the projector onto the span of the top $k$ eigenvectors, then the images $\pi_k v_i$, which live in a $k$-dimensional subspace, are precisely those vectors. We can see below that the resulting low-rank matrix has ringing (\href{https://colab.research.google.com/drive/13L51UzyNQ6SnjNZRyhWsx_QuyQ3LoosW\#scrollTo=c1q8ozayknvl}{colab notebook}). Finally, because the \href{https://en.wikipedia.org/wiki/Circulant_matrix}{discrete Fourier transform diagonalizes circulant matrices}, the Fourier coefficients of $f$ are in fact the principal values of $X$; the low-rank approximation consists of truncating the small fourier coefficients of $f$; and the resulting rows of $X$ exhibit ringing.

\begin{figure}[!htp]
    \centering
    \includegraphics[width=0.9\textwidth,height=0.7\textheight,keepaspectratio]{figures/fig_040_gdoc.png}
    \label{fig:gdoc_40}
    \caption{Pairwise cosine similarity of the ideal similarity matrix and its rank-5 approximation, reproduces ringing behavior.}
\end{figure}

One essential feature of the representation of line character counts is that the “boundary head” twists the representation, enabling each count to pair with a count slightly larger, indicating that the boundary is close. That is, there is a linear map QK which slides the character count curve along itself. Such an action is not admitted by generic high-curvature embeddings of the circle or the interval like the ones in the physical model we constructed. But it is present in both the manifold we observe in Haiku and, as we now show, in the Fourier construction. First, note permutating the coordinates of $\mathbb{R}^n$ by taking $e_i \mapsto e_{i+1}$ has the effect of mapping $v_i \mapsto v_{i+1}$. That is, the (linear) action of this permutation $\rho$ on $\mathbb{R}^n$ acts by a rotation of the embedded circle with respect to its intrinsic geometry. Because conjugation by $\rho$ fixes the circulant matrix $X$, it therefore respects its eigendecomposition, and thus commutes with the projection $\pi_k$ onto the vector space spanned by its top $k$ eigenvectors. The restriction of $\rho$ to that subspace, $\overline{\rho}:=\pi_k \circ \rho \circ \pi_k$, acts by rotation on the lower-dimensional vectors $\overline{\rho}: \pi_k v_i \mapsto \pi_k v_{i+1}$. Thus we have found $n$ vectors in a $k$-dimensional space, whose similarity is as close as possible to $X$ (and has ringing), with a linear action of $\mathbb{R}^k$ that rotates the vectors along a rippled embedded circle.

We evaluate whether a Fourier decomposition of the character count curve is optimal, and find that it is quite close given that it does not account for dilation. Fourier components explain at most 10\% less variance than an equivalent number of PCA components, which are optimal for capturing variance.

\begin{figure}[!htp]
    \centering
    \includegraphics[width=0.9\textwidth,height=0.7\textheight,keepaspectratio]{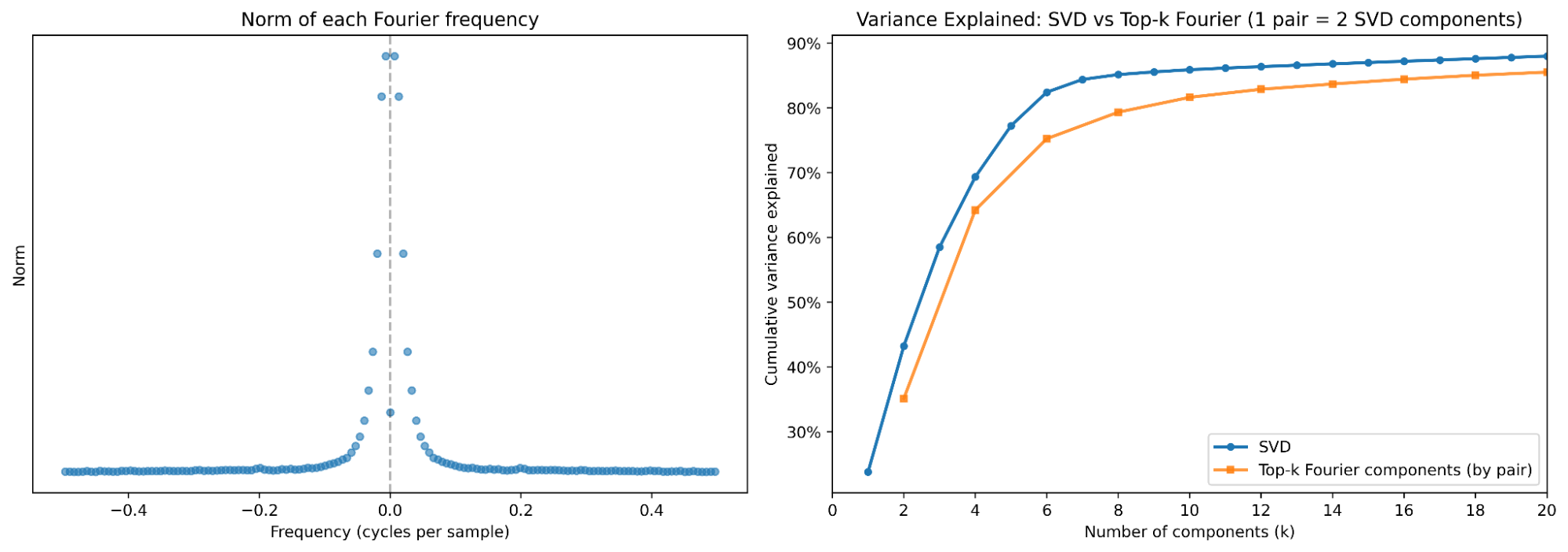}
    \label{fig:gdoc_41}
    \caption{Fourier results on character counting curve.}
\end{figure}

Finally, we note that as one moves through layers, the representation becomes more peaked. This sharpening of the receptive field is useful to the model to better estimate character counts, and corresponds to higher curvature in the embedding and, as predicted by the model above, more pronounced ringing. Below we show cross-sections (at character count 30, 60, 90, 120) of the cosine similarity matrix of probes trained after layers 0, 1, 2, and 3. With each subsequent layer, the graphs get more tightly peaked and secondary rings go higher.

\begin{figure}[!htp]
    \centering
    \includegraphics[width=0.9\textwidth,height=0.7\textheight,keepaspectratio]{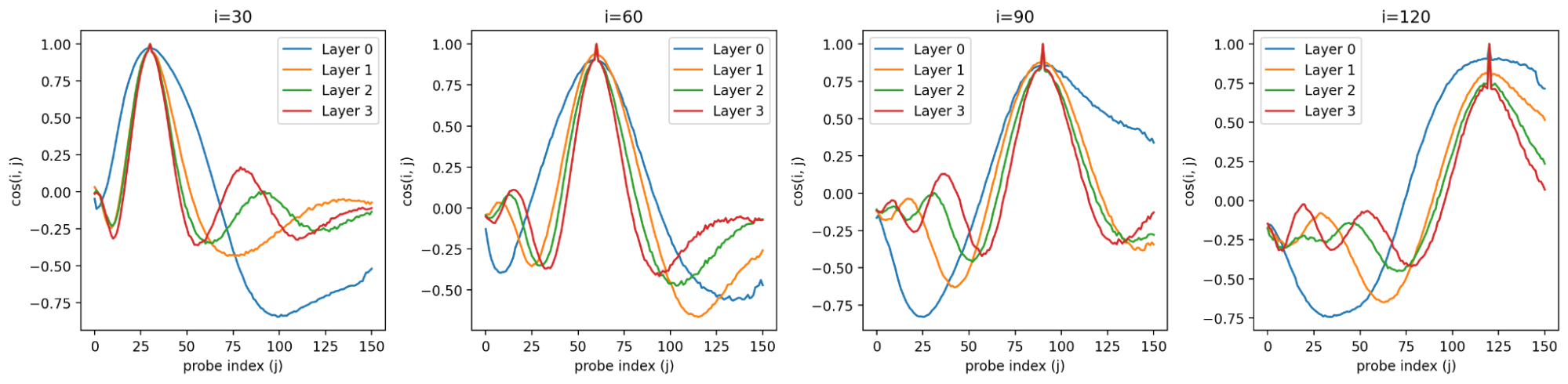}
    \label{fig:gdoc_42}
    \caption{Cross sections of character count pairwise similarity.}
\end{figure}

\section{Geometry of Twisting}\label{sec:appendix-twisting}

Different heads access and manipulate the space in different ways. Below, we show the cosine similarity of both probe sets through QK for three heads: one which keeps them aligned, one which shifts character count to align better with later line widths, and one which does the opposite.

\begin{figure}[!htp]
    \centering
    \includegraphics[width=0.9\textwidth,height=0.7\textheight,keepaspectratio]{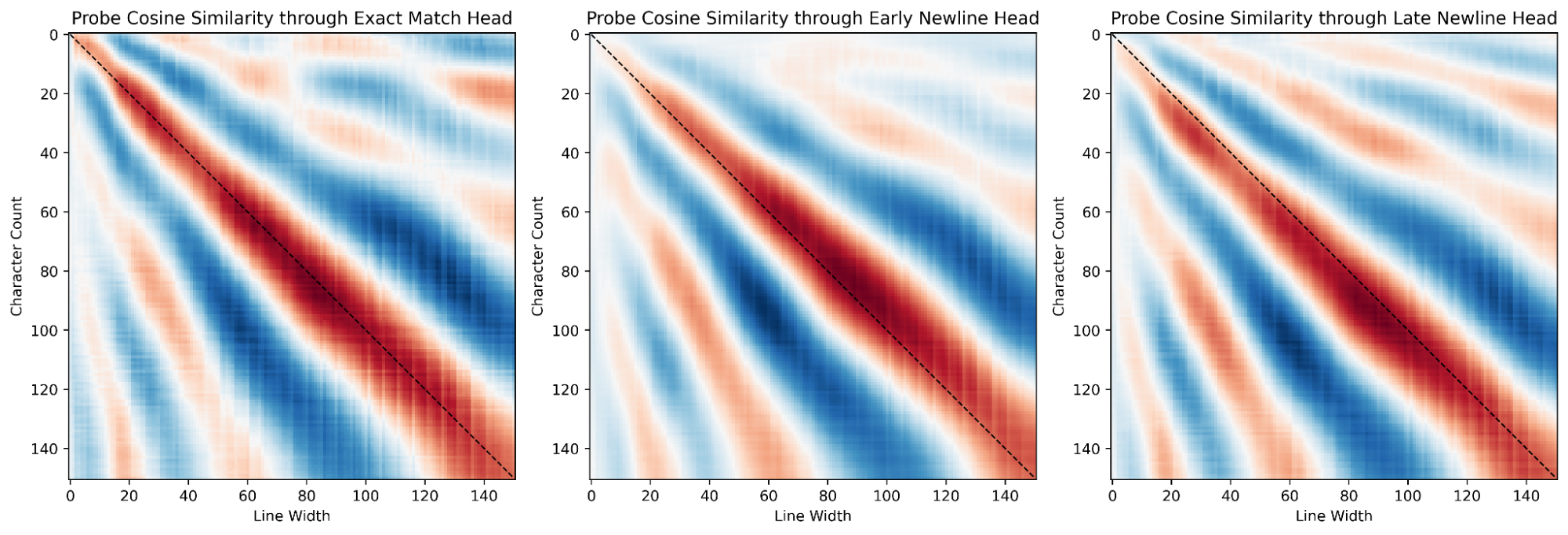}
    \label{fig:gdoc_43}
    \caption{Pairwise cosine similarity of probes through different boundary detection heads.}
\end{figure}

We can also look at this transformation by visualizing the Singular Value Decomposition of each set of probes in a joint basis after passing them through QK. Once more, the alignment, left offset, and right offset can be read directly from the components.

\begin{figure}[!htp]
    \centering
    \includegraphics[width=0.9\textwidth,height=0.7\textheight,keepaspectratio]{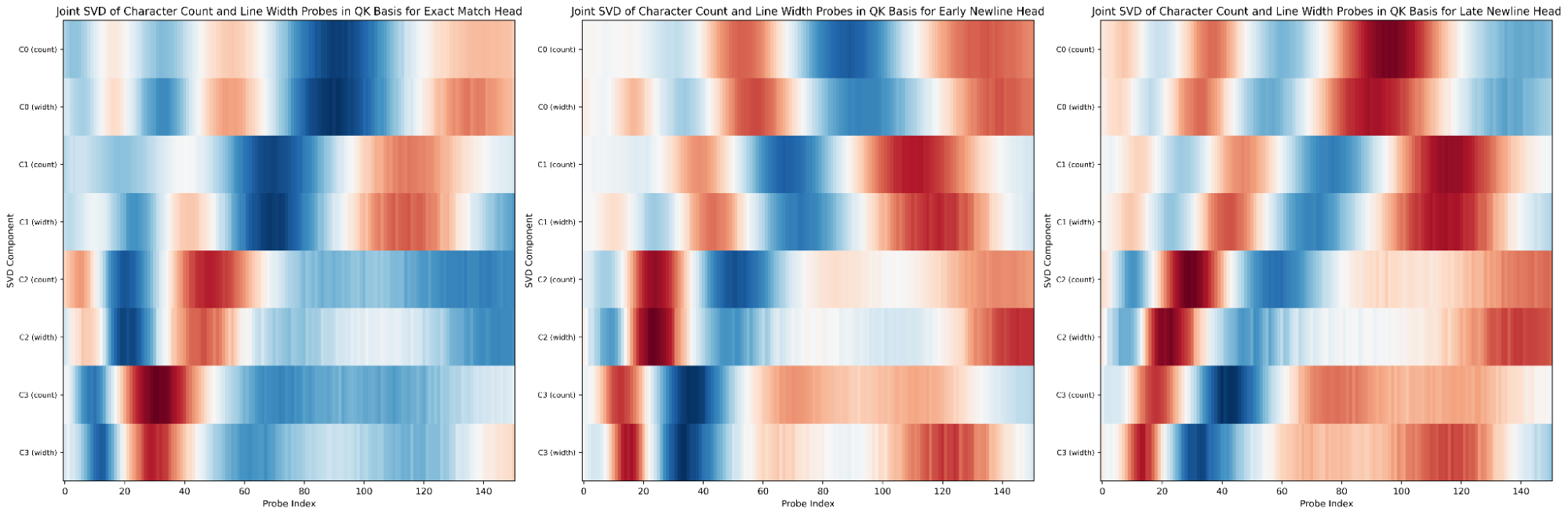}
    \label{fig:gdoc_44}
    \caption{Loading of different PCA components as a function of count for three different boundary detection heads.}
\end{figure}

We can directly plot the first 3 components of the joint probe space after passing them through each QK. Doing so shows that one head keeps the representations aligned, while the others twist them either clockwise or counterclockwise.

\begin{figure}[!htp]
    \centering
    \includegraphics[width=0.9\textwidth,height=0.7\textheight,keepaspectratio]{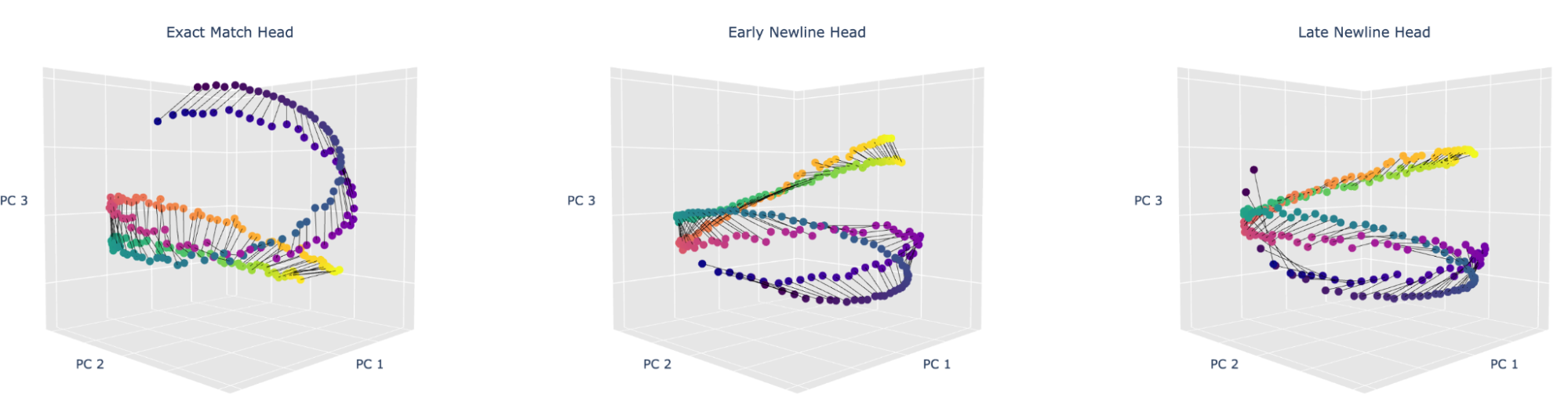}
    \label{fig:gdoc_45}
\end{figure}

\section{Break Predictor Features}\label{sec:appendix-break-predictors}

Boundary detector features (at about $\sim$1/3 model depth) do not take into account the length of the next token.

\begin{figure}[!htp]
    \centering
    \includegraphics[width=0.9\textwidth,height=0.7\textheight,keepaspectratio]{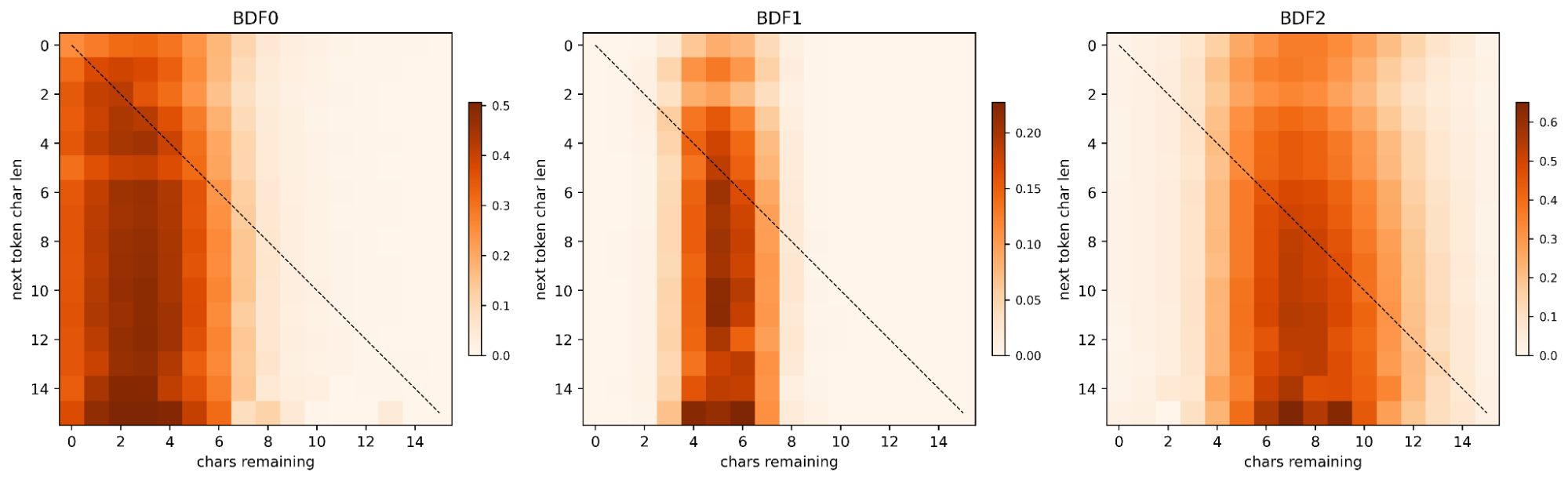}
    \label{fig:gdoc_46}
\end{figure}

Later in the model, there exist features which incorporate both the number of characters remaining \textit{and} the length of the most likely next token. These features only activate when the most likely next token is longer than the number of characters remaining (i.e. below the red diagonal below), as is the case in our aluminum prompt.

\begin{figure}[!htp]
    \centering
    \includegraphics[width=0.9\textwidth,height=0.7\textheight,keepaspectratio]{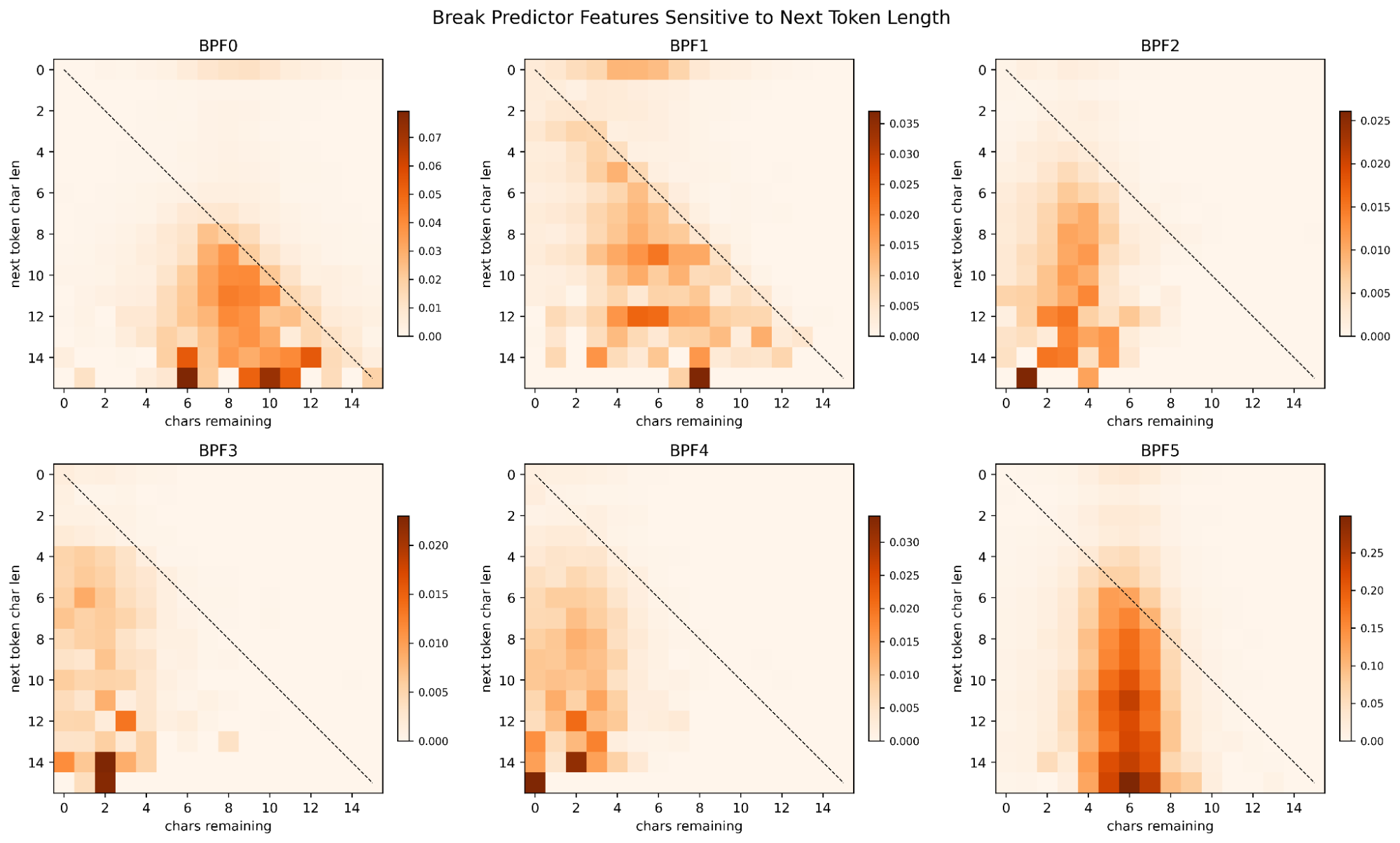}
    \label{fig:gdoc_47}
\end{figure}

We also found features for the converse: features which suppress the newline because the predicted next token is \textit{shorter} than the number of characters remaining in the line.

\begin{figure}[!htp]
    \centering
    \includegraphics[width=0.9\textwidth,height=0.7\textheight,keepaspectratio]{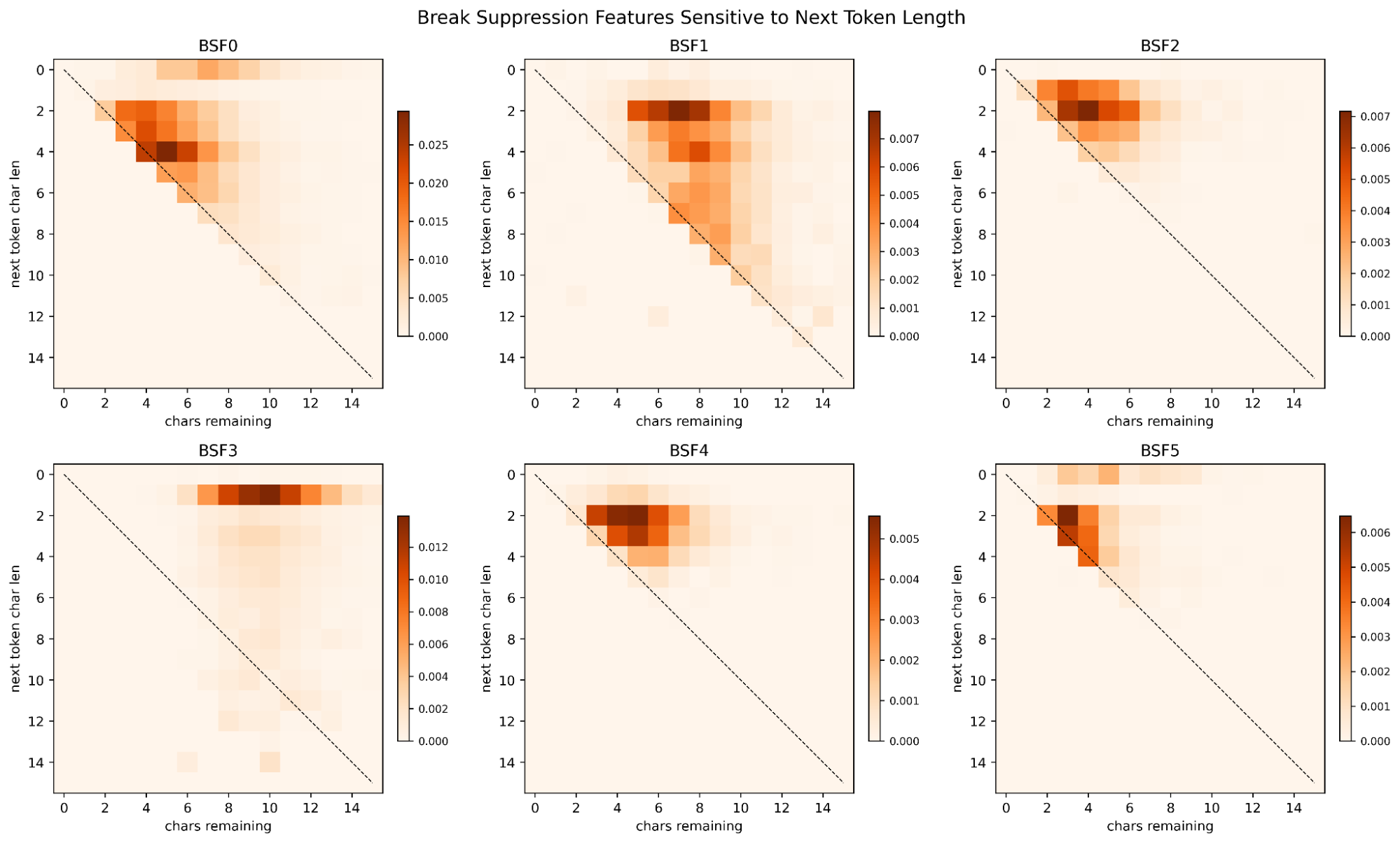}
    \label{fig:gdoc_48}
\end{figure}

Both break prediction and suppression features sometimes also have interpretable logit effects on the output of all tokens, not just the newline. For instance, the features below respectively excite and suppress the newline as their top effect, but also systematically suppress tokens with more characters. This is because if the model is wrong about the value of the next token (and whether it's a newline), the token must at least be short enough to fit on the line.

\begin{figure}[!htp]
    \centering
    \includegraphics[width=0.9\textwidth,height=0.7\textheight,keepaspectratio]{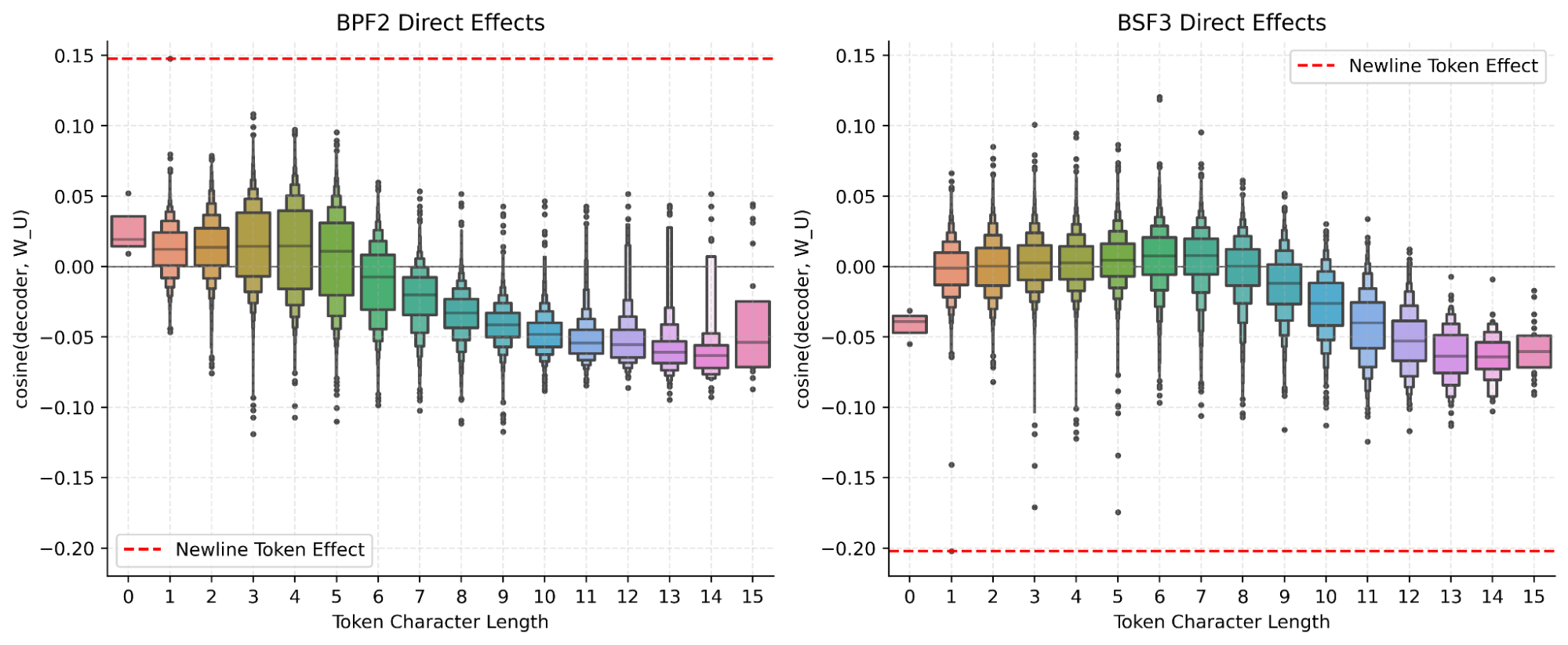}
    \label{fig:gdoc_49}
\end{figure}

\section{Representing Token Lengths}\label{sec:appendix-token-lengths}

We find layer 0 features that activate as a function of the character count of individual tokens.

\begin{figure}[!htp]
    \centering
    \includegraphics[width=0.9\textwidth,height=0.7\textheight,keepaspectratio]{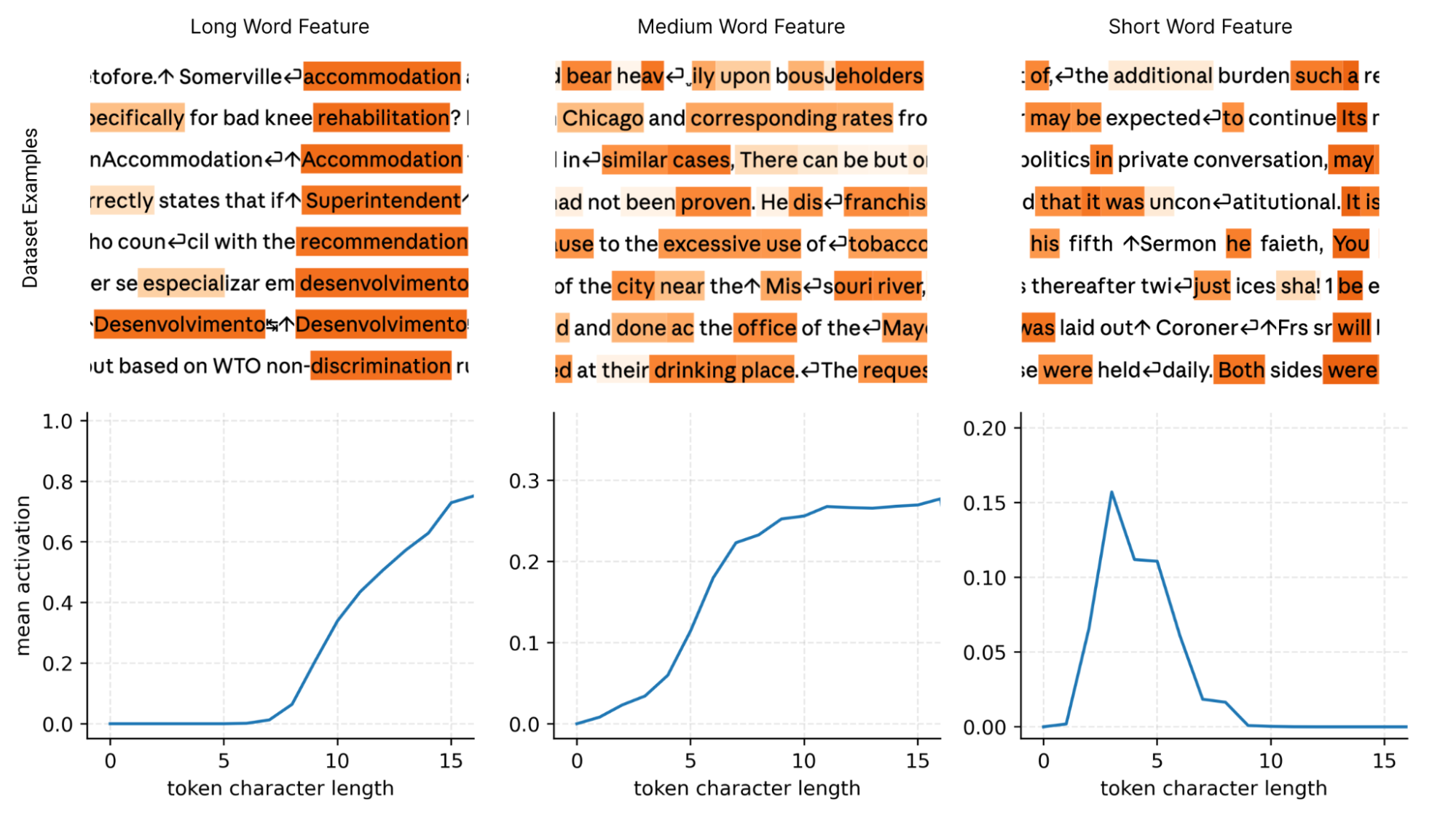}
    \label{fig:gdoc_50}
\end{figure}

These features are overlapping (e.g. there are tokens for which the long word and medium word features are both active) and non-exhaustive (none of them fire on some common tokens, where we suspect the representation of character length is partially absorbed \cite{chanin2024absorption} into features which just activate for that token).

\section{The Mechanics of Head Specialization}\label{sec:appendix-mechanics}

Heads collaborate to generate the count manifold, but \textit{how} does each head aggregate counts?

As a toy model, consider the following construction for character counting with a single attention head:

\begin{itemize}
    \item The head uses the previous newline token as a “sink” where it defaults all of its attention to (i.e., attention 1)
    \item Each token since the newline gets $\alpha$ attention, such that after $j$ tokens the newline has $1 - \alpha j$ attention. Note this limits the construction to only work on lines of up to $1 / \alpha$ tokens, but the model could use multiple heads at different offsets to count for longer sequences. The model could also dedicate attention to tokens proportionally to token length
    \item The output of the head on the newline is 0, and the output of each non-newline token is a vector with the same direction but magnitude proportional to the character count of the token.
\end{itemize}

This produces a ray with total length proportional to the character count of the line.

\begin{figure}[!htp]
    \centering
    \includegraphics[width=0.9\textwidth,height=0.7\textheight,keepaspectratio]{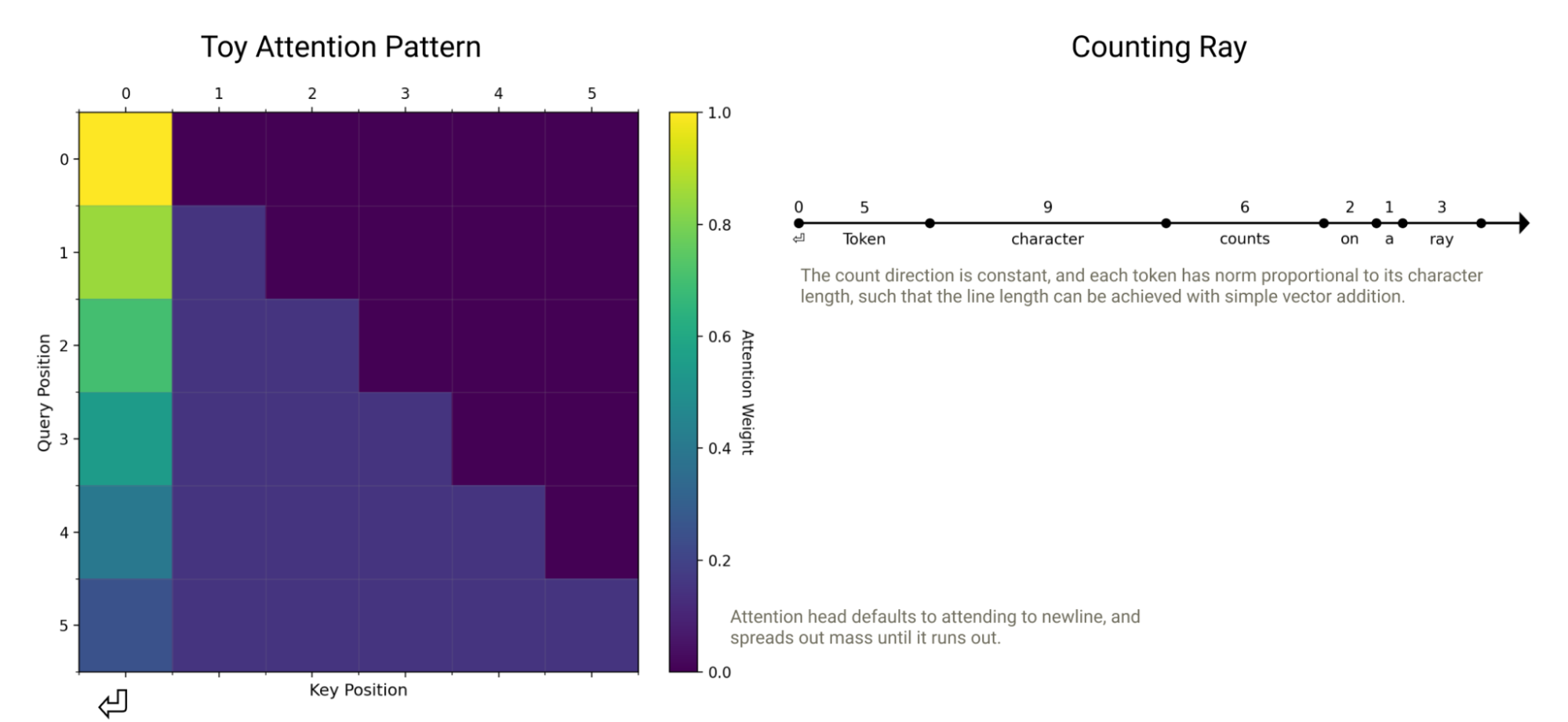}
    \label{fig:gdoc_51}
\end{figure}

In practice, we observe that individual attention heads do indeed use the newline as an attention sink, but at different offsets. As an example, we visualize the attention patterns of 4 important Layer 0 heads on several prompts with different line widths (starting from the first newline in the sequence).

\begin{figure}[!htp]
    \centering
    \includegraphics[width=0.9\textwidth,height=0.7\textheight,keepaspectratio]{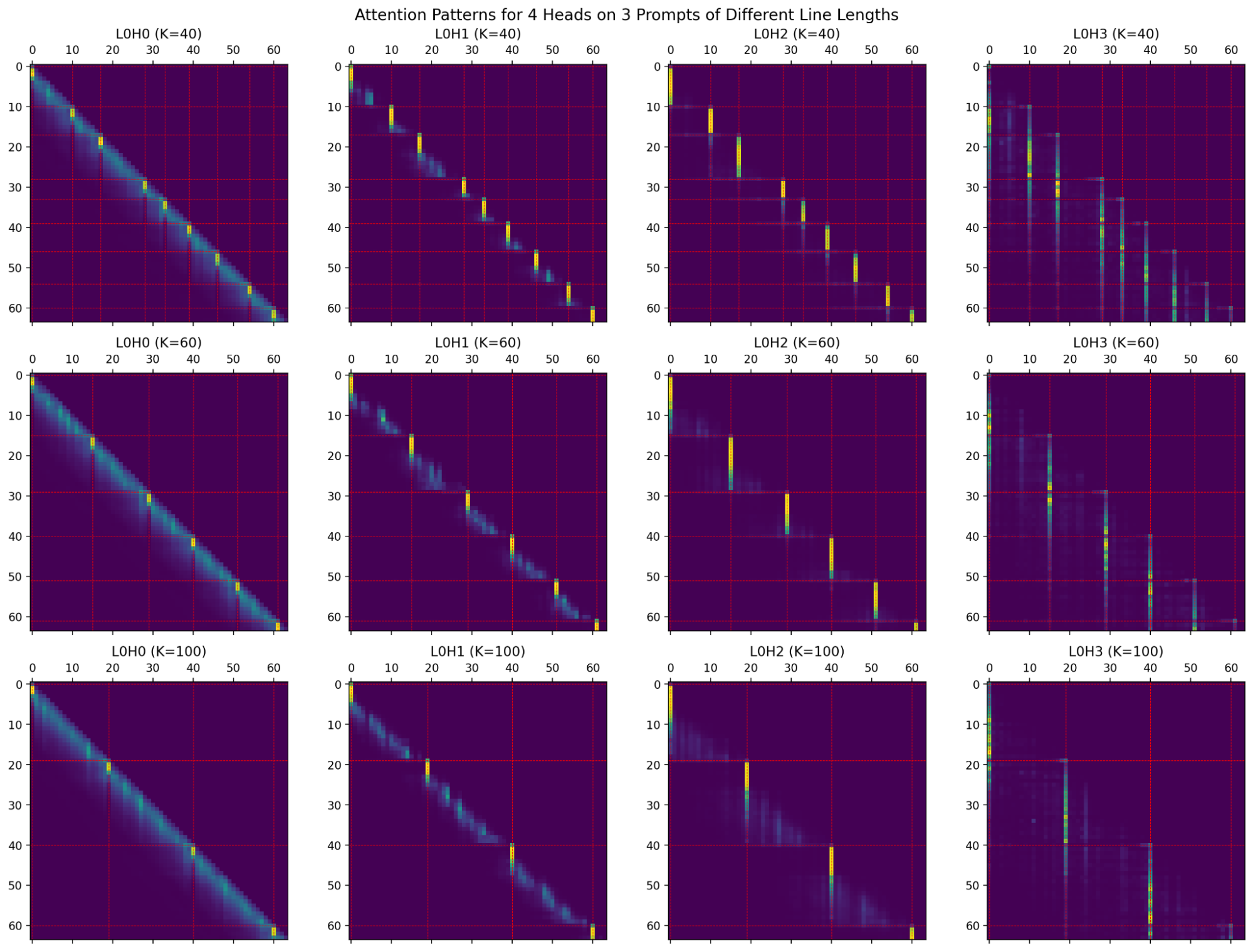}
    \label{fig:gdoc_52}
    \caption{Attention patterns of four important Layer 0 heads (columns) for 3 different prompts (rows) prompt showing head specialization. Patterns start from the first newline with red dashes indicating linebreaks.}
\end{figure}

To characterize the mechanism more precisely, we compute the average attention as a function of the number of tokens since the previous newline and also as a function of the character length of individual tokens.

\begin{figure}[!htp]
    \centering
    \includegraphics[width=0.9\textwidth,height=0.7\textheight,keepaspectratio]{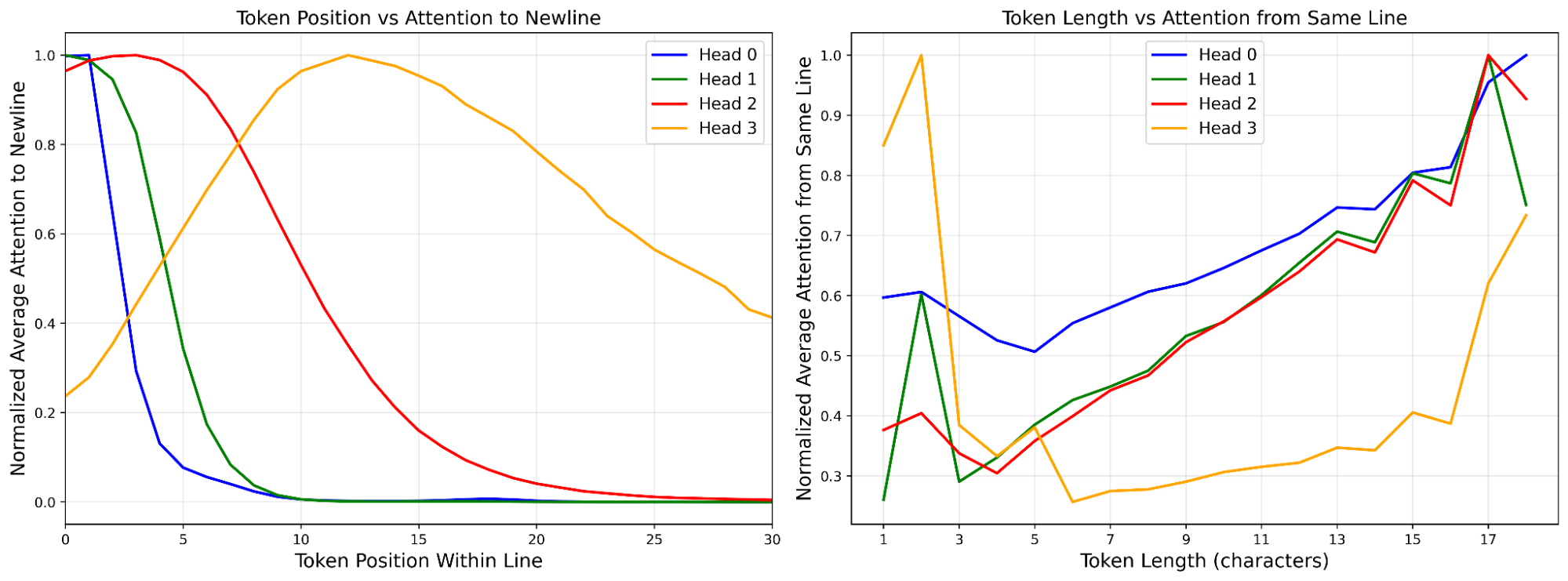}
    \label{fig:gdoc_53}
    \caption{Normalized attention as a function of tokens since newline (left) and token character length (right) for four heads. Each head specialized in a different offset similar to boundary heads.}
\end{figure}

Similar to boundary detection, individual attention heads specialize in particular offsets to tile the space. Moreover, we observe that most of these attention heads have a bias towards attending to longer tokens.

In addition to QK, a head can change its output based on the OV circuit \cite{nelhage2021mathematical}. We study this by analyzing the pairwise interaction of probes as mediated by the OV matrix. Specifically, for the averaged token embedding vectors for each token length $E_t$\footnote{That is, for each token character length $i$, we compute the average embedding vector in $W_E$. We also prepend this with the newline embedding vector to make the plot below.}, our line length probes $P_c$, and the weight matrices for the attention output of each head $W_{OV}$, we compute $P_c^T W_{OV} E_t$.

\begin{figure}[!htp]
    \centering
    \includegraphics[width=0.9\textwidth,height=0.7\textheight,keepaspectratio]{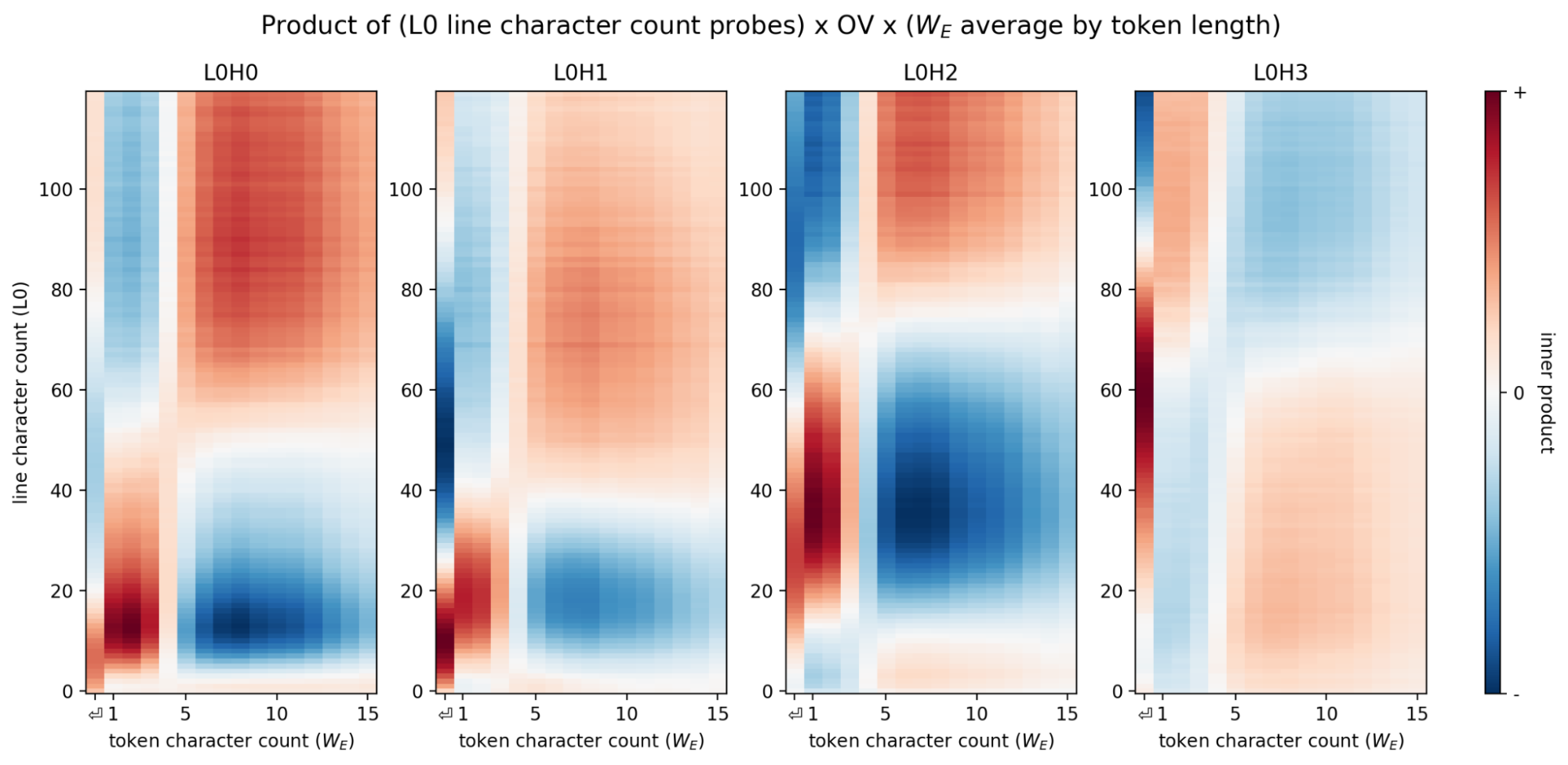}
    \label{fig:gdoc_54}
    \caption{Inner product of line character count and token character count through OV for 4 important layer 0 heads. The OV responses reflect the difference in attention pattern biases.}
\end{figure}

The output of each head can be thought of as having two components: (1) a character offset from the newline driven by the attention pattern and (2) an adjustment based on the actual character length of the tokens. Note that the average character count of a token is approximately 4.5 (and the median is 4), so we can interpret these effects shifting from a mean response (i.e. the transition point is always around count 4).

To walk through a head in action, consider the perspective of L0H1, which attends to the newline for the first $\sim$4 tokens and then spreads out attention over the previous $\sim$4--8 tokens:

\begin{itemize}
    \item While attending to the newline, L0H1 writes to the 5--20 character count (5--20CC) directions and suppresses the 30--80CC directions. This makes sense as an approximation because attending to the newline implies that the line is currently at most 3 tokens long (newline has no width) and on average, any 3 token span is $\sim$15 characters (and is unlikely to be >30).
    \item While \textit{not} attending to the newline at all, L0H1 defaults to predicting CC40, since not attending to the newline implies that there are $\sim$8 tokens in the line with $\sim$5 characters each on average (including spaces). Then, there is an additional correction applied depending on how long the tokens are:
\end{itemize}

\begin{itemize}
    \item (1) If tokens being attended to are \textit{short} (<4 chars), then upweight 10--35CC and downweight >40CC.
    \item (2) If tokens being attended to are \textit{long} (\ensuremath{\geq}5 chars) then do the opposite.
\end{itemize}

\begin{itemize}
    \item In cases with some attention on newlines and nonnewlines, linearly interpolate the above predictions.
\end{itemize}

Other heads perform a similar operation, except with different offsets depending on their newline sink behavior. Layer 1 heads also perform a similar operation, though they also can leverage the character count estimate of the Layer 0 heads (see \hyperref[sec:appendix-l1-attn]{Layer 1 Head OVs}).

\section{Layer 1 Head OVs}\label{sec:appendix-l1-attn}

Similar to the OVs of the Layer 0 attention heads, Layer 1 heads write to the character count features in accordance to how long the tokens they attend to are.

\begin{figure}[!htp]
    \centering
    \includegraphics[width=0.9\textwidth,height=0.7\textheight,keepaspectratio]{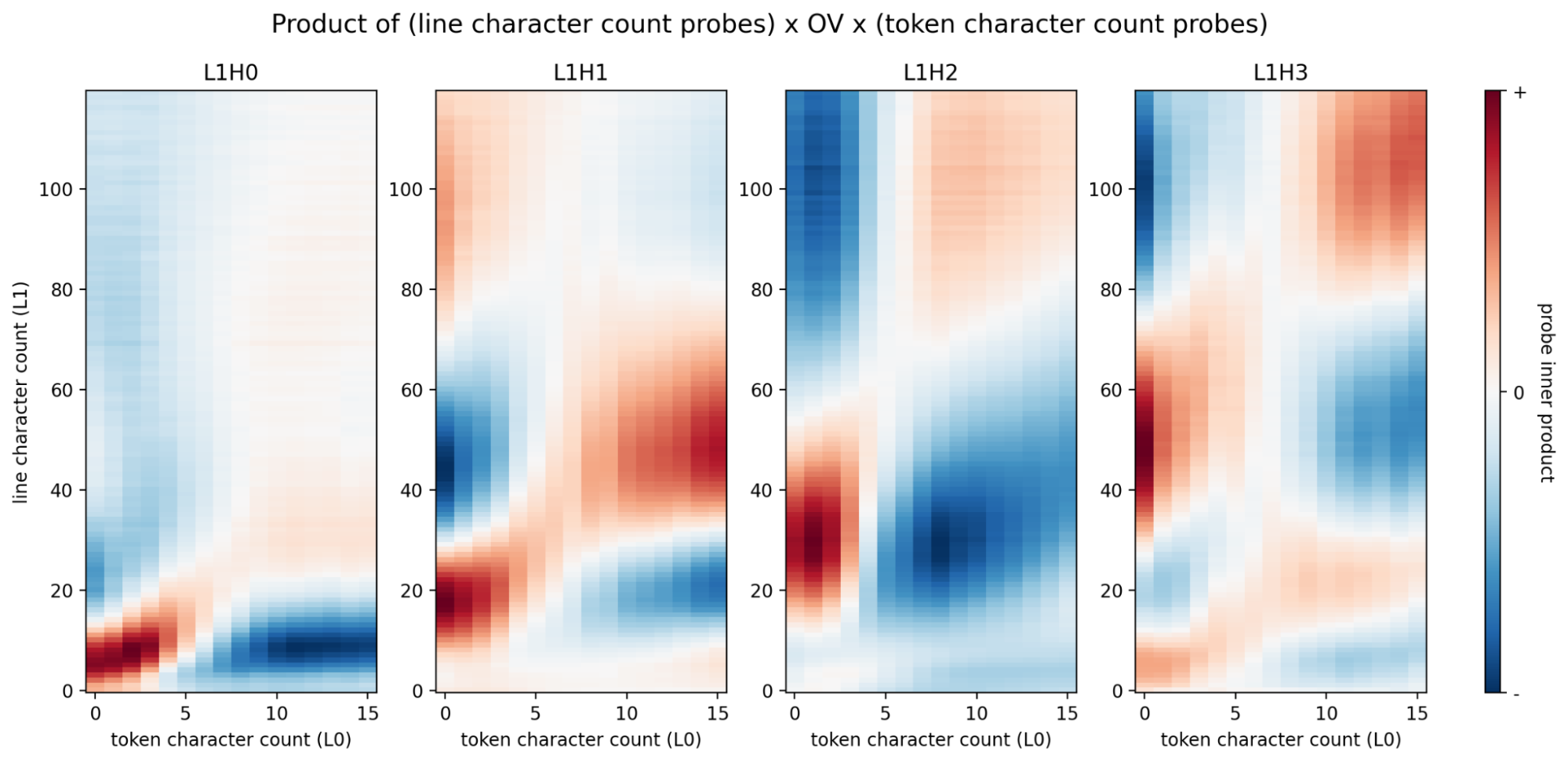}
    \label{fig:gdoc_55}
\end{figure}

However, in addition to the token character length, Layer 1 heads also use the initial line length estimate constructed in Layer 0 to create a more refined estimate of the character count.

\begin{figure}[!htp]
    \centering
    \includegraphics[width=0.9\textwidth,height=0.7\textheight,keepaspectratio]{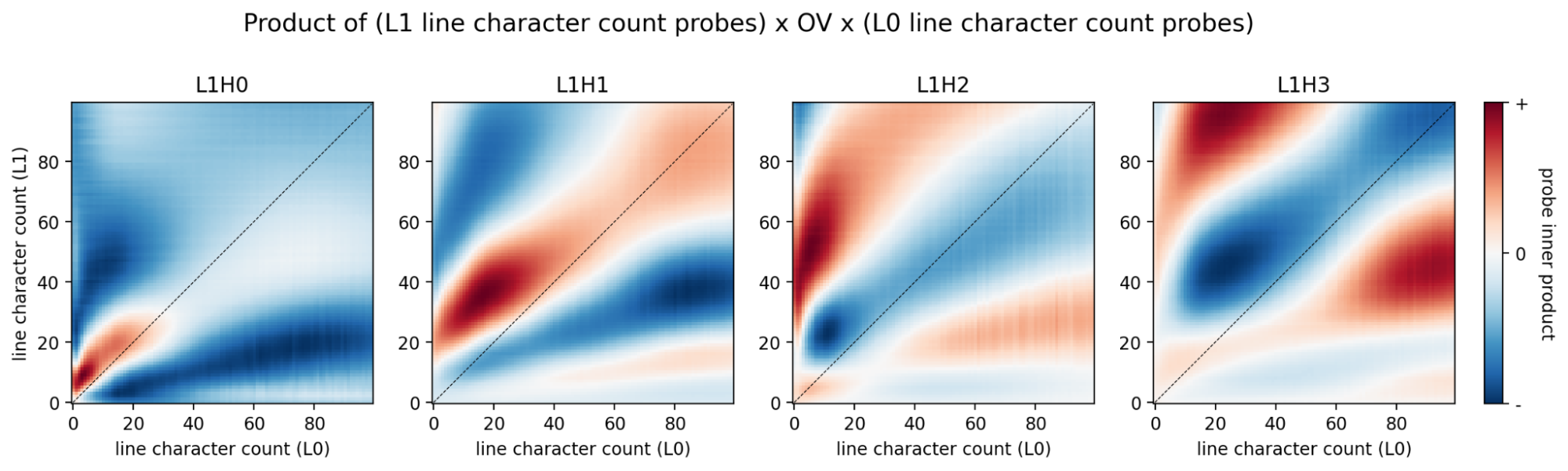}
    \label{fig:gdoc_56}
\end{figure}

These repeated computations appear responsible for implementing the sharpening of representations.

\begin{figure}[!htp]
    \centering
    \includegraphics[width=0.9\textwidth,height=0.7\textheight,keepaspectratio]{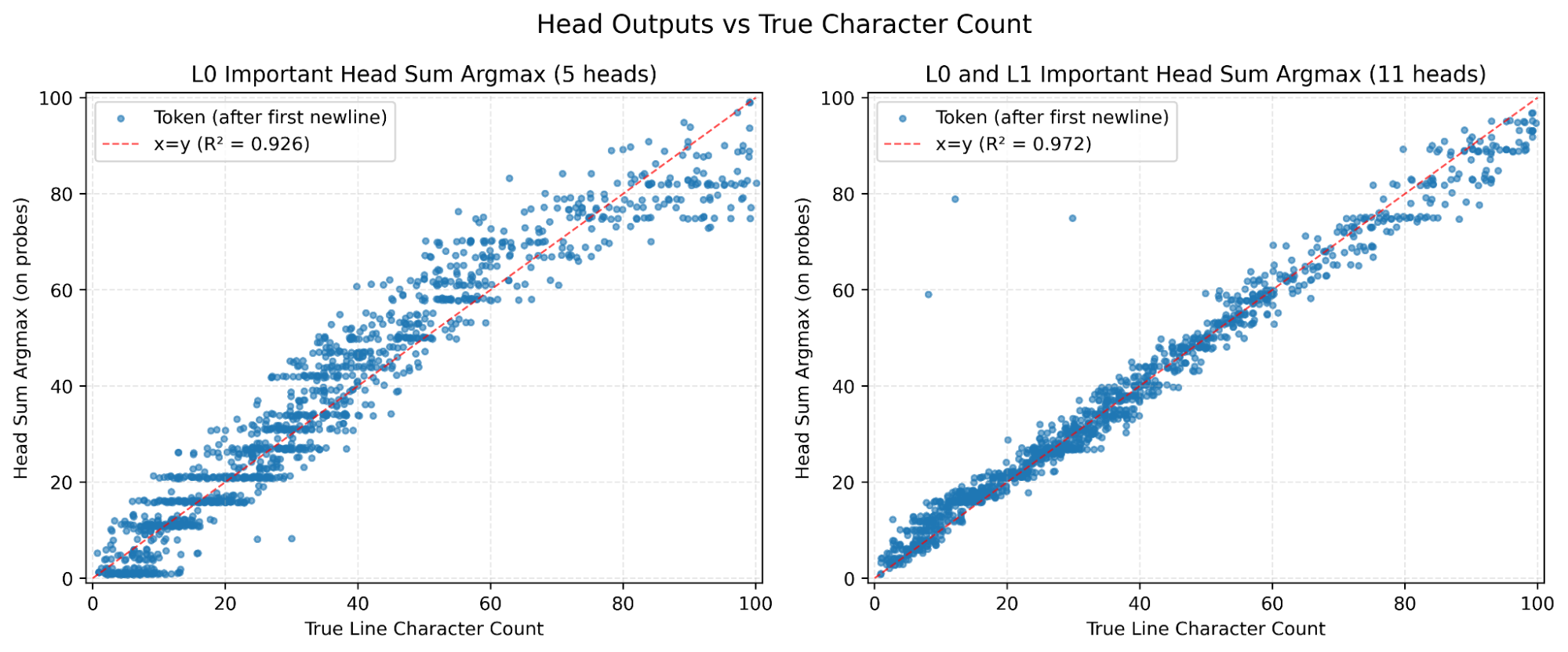}
    \label{fig:gdoc_57}
\end{figure}

\section{Full Layer 0 Attention Results}\label{sec:appendix-full-l0-attn}

Below, we show head sums for 3 different prompts with different line widths.

\begin{figure}[!htp]
    \centering
    \includegraphics[width=0.9\textwidth,height=0.7\textheight,keepaspectratio]{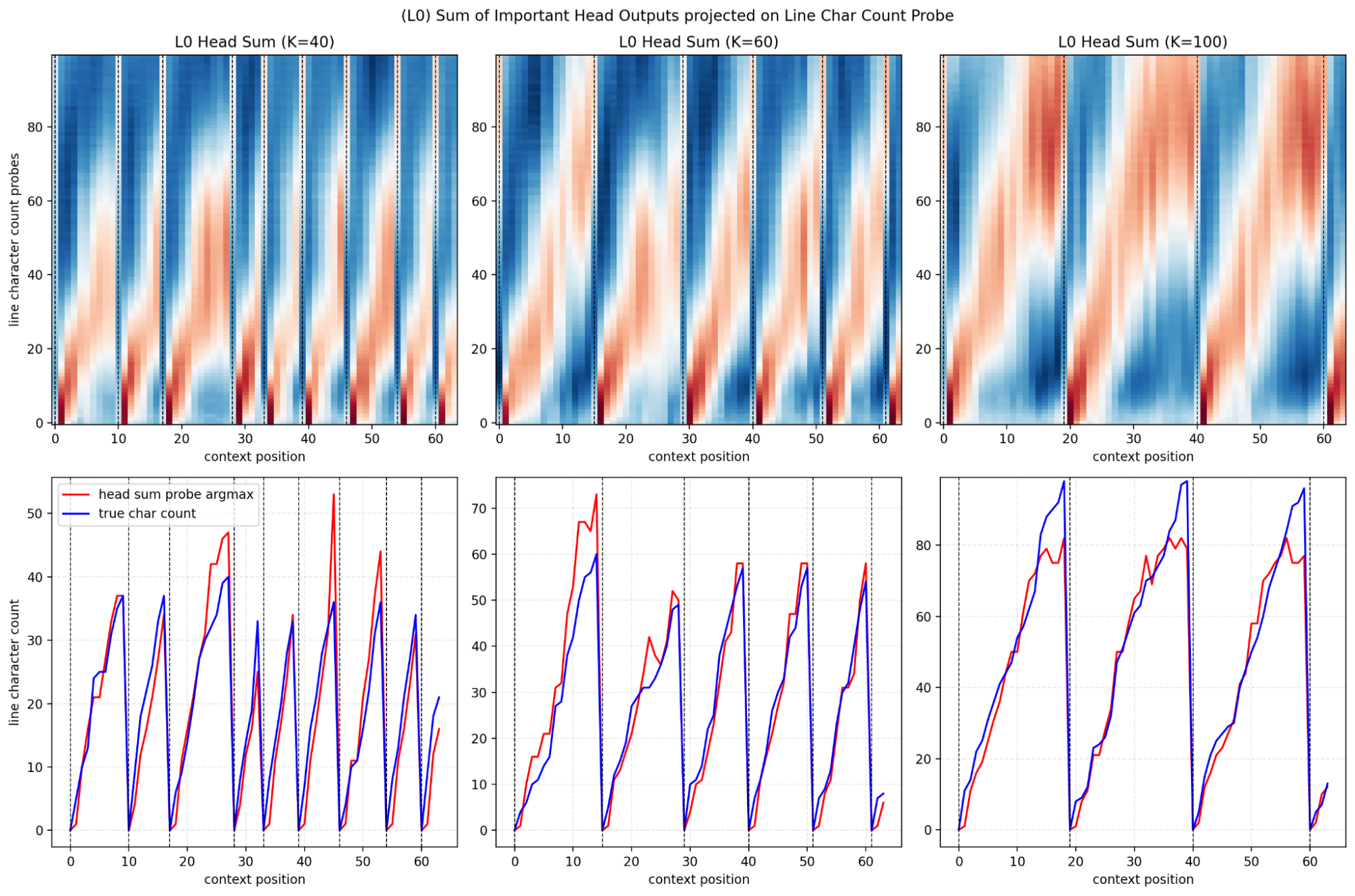}
    \label{fig:gdoc_58}
\end{figure}

As before, we can look at their decomposition.

\begin{figure}[!htp]
    \centering
    \includegraphics[width=0.9\textwidth,height=0.7\textheight,keepaspectratio]{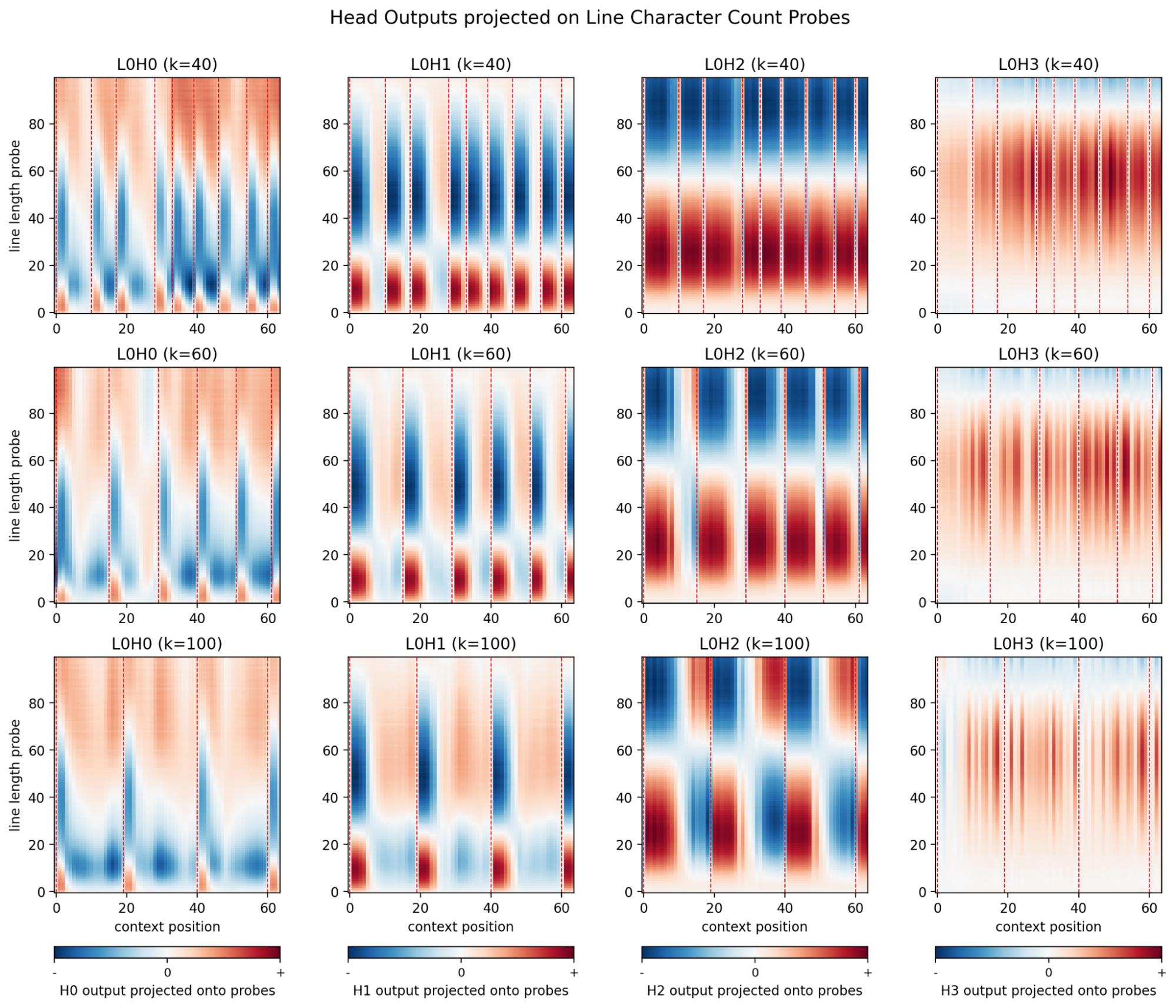}
    \label{fig:gdoc_59}
\end{figure}

\section{More Sensory and Counting Representations}\label{sec:appendix-sensory}

While in this work we carefully studied the perception of line lengths and fixed width text, there are many tasks which language models must perform which benefit from a positional, visual, or spatial representation of the underlying text. In the course of our investigation, we came across several other feature families and representations for these behaviors and report several below.

\subsection{What Follows an Early Linebreak?}

In addition to line width for tracking the absolute character length of a full line of text, there also exist features that are sensitive to lines which have ended early (i.e, lines where the character count is substantially shorter than the line width $k$). While these features are less useful for linebreaking, they enable the model to better predict the token following a linebreak. Specifically, if a line ends $c$ characters before the line limit $k$, the next word should be at least $c$ characters, otherwise it would have been able to fit in the previous line.

\begin{figure}[!htp]
    \centering
    \includegraphics[width=0.9\textwidth,height=0.7\textheight,keepaspectratio]{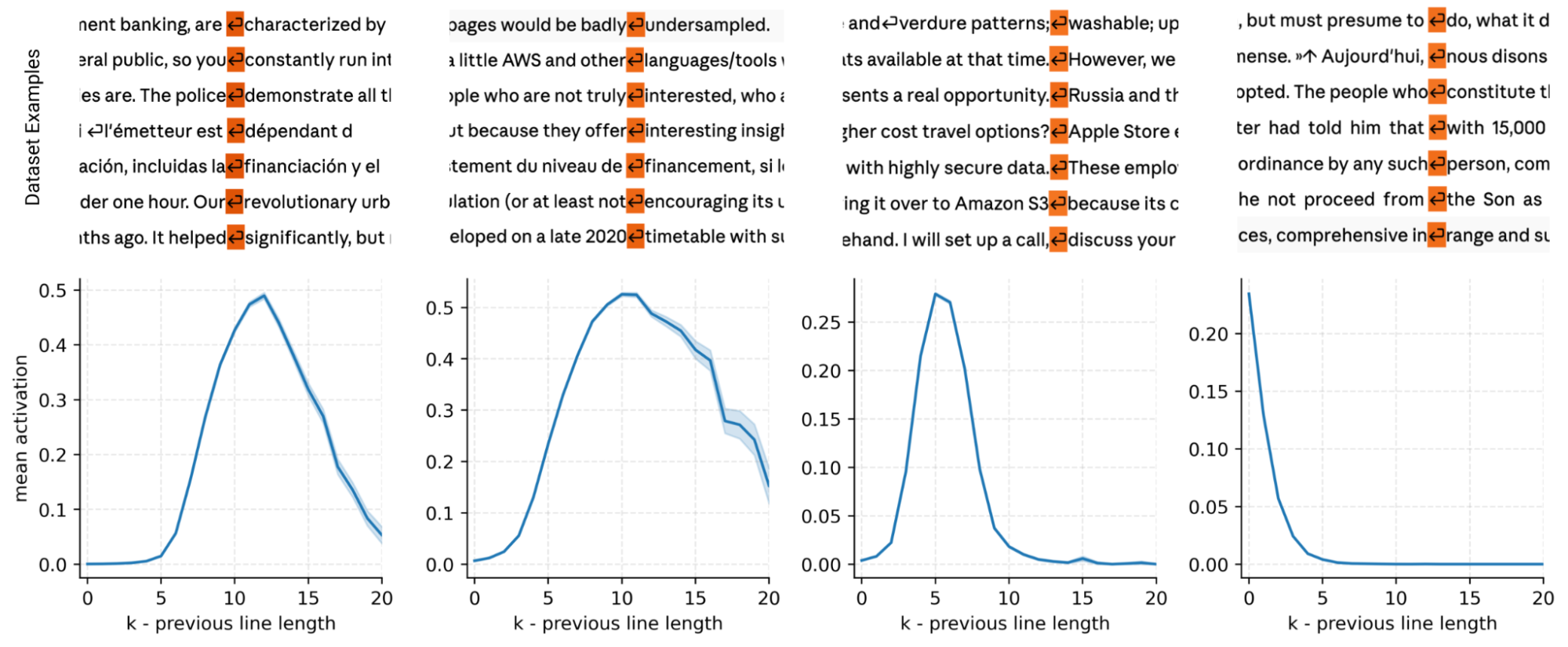}
    \label{fig:gdoc_60}
    \caption{A feature family for how many characters were remaining in a line after it was broken.}
\end{figure}

It is worth emphasizing that the role of these features, like others in this work, is not obvious from a typical workflow of quickly looking at dataset examples. It might be tempting to ignore these as ``newline'' features, but careful analysis yields quite clear behavior.

\subsection{Markdown Table Representations}

In addition to prose, language models must parse other kinds of more structured data like tables. Accurate prediction of a table’s content requires careful integration of row and column information (e.g. is this a column of text or numbers?). To facilitate this, we use a synthetic dataset of 20 markdown tables to find feature families which activate on separator tokens, specialized to particular rows or columns. Visualizing feature activations on each of these 20 tables (arranged by location in the table) showed clear patterns.

\begin{figure}[!htp]
    \centering
    \includegraphics[width=0.9\textwidth,height=0.7\textheight,keepaspectratio]{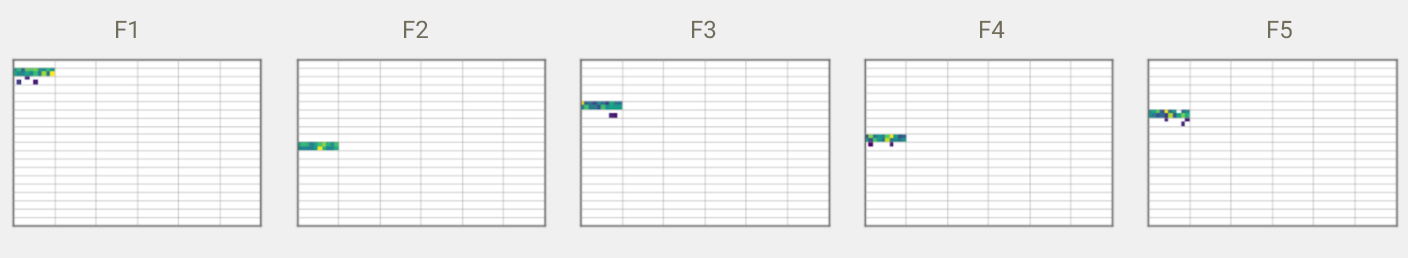}
    \label{fig:gdoc_61}
    \caption{A feature family for the row index in markdown tables. Activations are shown for 20 tables on the ''|'' token in the first column of the nth row.}
\end{figure}

\begin{figure}[!htp]
    \centering
    \includegraphics[width=0.9\textwidth,height=0.7\textheight,keepaspectratio]{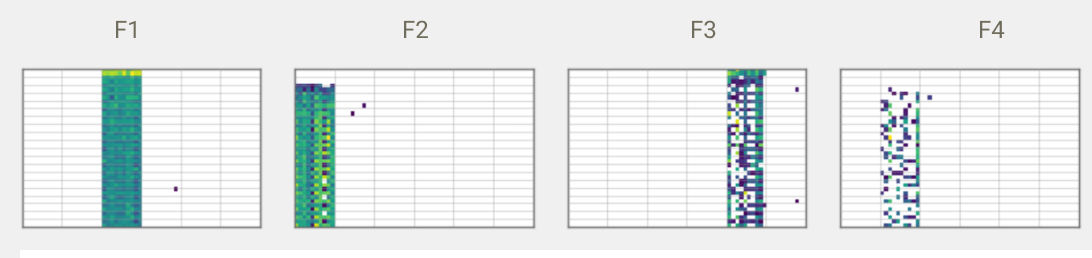}
    \label{fig:gdoc_62}
    \caption{A feature family for the column index in markdown tables. Activations are shown for 20 tables on the ''|'' tokens in the nth column.}
\end{figure}

On a synthetic dataset of larger tables, we also observe counting representations for the column and row index that resemble the character counting representations. Specifically, we see ringing in the pairwise probe cosine similarities and the characteristic “baseball seam” in the PCA basis.

\begin{figure}[!htp]
    \centering
    \includegraphics[width=0.9\textwidth,height=0.7\textheight,keepaspectratio]{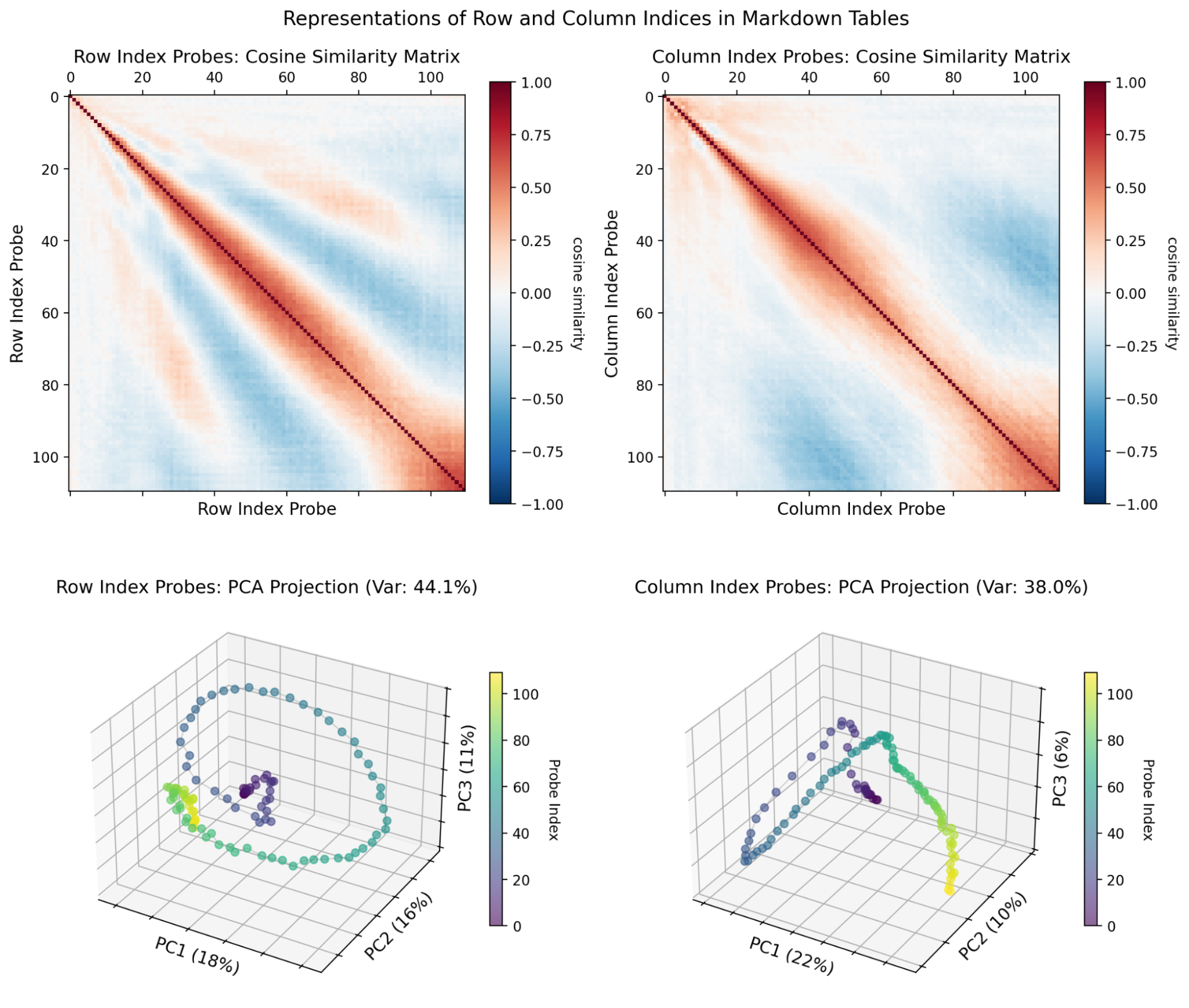}
    \label{fig:gdoc_63}
    \caption{Representations of markdown table row indices (left) and column indices (right). (Top) pairwise inner products of probes trained to predict the index; (bottom) probes projected into a 3D PCA basis.}
\end{figure}

\section{Rejected Titles}\label{sec:appendix-title}

\begin{itemize}
    \item A General Language Assistant as a Laboratory for A-line-ment
    \item A-line-ment Science: The Geometry of Textual Perception
    \item The Geometry of Textual Perception: How Models Stay Aligned
    \item The Geometry of Counting: How Transformers Perceive and Manipulate Spatial Structure in Text
    \item The Mechanistic Basis of Alignment
    \item Reading between the lines: The Perception of Linebreaks
    \item Linebreaking: More than you wanted to know
    \item The Line Must Be Drawn Here! Character Counting in Neural Networks
    \item Newline, Who Dis? Attention Heads and Their Distractible Nature
    \item The End of the Line: How Language Models Count Characters
    \item How I Learned to Stop Worrying and Love the Carriage Return
    \item Breaking down the Linebreak Mechanism: The Geometry of Text Perception
    \item We Found a GPS Inside a GPT
\end{itemize}

\end{document}